# Combining Bayesian Inference and Reinforcement Learning for Agent Decision Making: A Review

Chengmin Zhou, Ville Kyrki, Pasi Fränti, and Laura Ruotsalainen

*Abstract*--**Bayesian inference has many advantages in decision making of agents (e.g. robotics/simulative agent) over a regular data-driven black-box neural network: Data-efficiency, generalization, interpretability, and safety where these advantages benefit directly/indirectly from the uncertainty quantification of Bayesian inference. However, there are few comprehensive reviews to summarize the progress of Bayesian inference on reinforcement learning (RL) for decision making to give researchers a systematic understanding. This paper focuses on combining Bayesian inference with RL that nowadays is an important approach in agent decision making. To be exact, this paper discusses the following five topics: 1) Bayesian methods that have potential for agent decision making. First *basic Bayesian methods and models* (Bayesian rule, Bayesian learning, and Bayesian conjugate models) are discussed followed by variational inference, Bayesian optimization, Bayesian deep learning, Bayesian active learning, Bayesian generative models, Bayesian meta-learning, and lifelong Bayesian learning. 2) *Classical combinations of Bayesian methods with model-based RL* (with approximation methods), *model-free RL*, and *inverse RL*. 3) *Latest combinations of potential Bayesian methods with RL*. 4) *Analytical comparisons of methods that combine Bayesian methods with RL* with respect to data-efficiency, generalization, interpretability, and safety. 5) *In-depth discussions in six complex problem variants of RL*, including unknown reward, partial-observability, multi-agent, multi-task, non-linear non-Gaussian, and hierarchical RL problems and *the summary of how Bayesian methods work in the data collection, data processing and policy learning stages of RL* to pave the way for better agent decision-making strategies.**

## I. INTRODUCTION

Rapid development of generative models [1][2][3] like ChatGPT and DeepSeek is expected to push artificial intelligence (AI) to more general purposes. Generative models suffer in mathematical applications such as consistently delivering high-quality proofs in math tests [4] due to the quality and quantity of training datasets, but this problem currently can be alleviated by combining RL and statistical methods like Bayesian inference (e.g., model-based reward models [3]) which enable coherent and consistent decision making with additional supervision. There are more specific challenges like sim-to-real transfer from the perspective of robotics. For example, data collection and exploration from the real-world is costly, time consuming, and unsafe. Transferring policies learned from simulations to the real-world is difficult. It is hard for RL to handle these challenges, but many promising solutions can be obtained if RL is combined with Bayesian inference. For example, Bayesian inference can make RL policy learned from simulations fast adapt to real-world tasks. Bayesian inference makes RL policy safer by constructing safe datasets and weights. In short, a further investigation of combining Bayesian inference with RL is the right direction to enable agent decision-making strategies to give deep insights across domains and original solutions for complex scenarios.

Bayesian inference and (model-based and model-free) RL are inspired by neural mechanisms [5][6][7]. Recent research reveals that the neural basis of behavior is consistent with normative probabilistic theories, and neural representations of individual components could underlie Bayesian inference [8][9][10][11]. Creatures like humans may start to learn by mechanisms resembling model-free RL and then continue with ones similar to the model-based RL, and they might be co-existing in the brain [12]. Based on recent studies, behavior patterns and dopamine activity that had previously been attributed to model-based RL might arise from model-free RL [12][13][14]. Linking this reasoning to RL, model-based RL might be guided in part by value estimates and action tendencies learned through model-free RL [15]. The difference between model-based RL and model-free RL is clearly presented by their respective Bellman equations [16] where model-based RL still relies on value estimates that are widely seen to be learned by model-free RL.

Both Bayesian inference and RL are good candidate frameworks for decision making, therefore arousing researchers' interest in figuring out their connections. "RL as inference" [17][18][19][20] is a popular framework where RL can be formulated as *probabilistic inference* in the case of deterministic dynamics and *variational inference* (VI) in the case of stochastic dynamics under Bayesian umbrella. One flaw of RL as inference is that it fails on problems where accurate uncertainty quantification is crucial to performance [21], but this flaw is amendable via methods like K-learning to incorporate the uncertainty to value function [18]. Hence, we may assume RL as another representation of Bayesian inference, but actually RL is a subset of Bayesian inference because not all Bayesian inference problems can be formulated as RL within the MDP framework.

Bayesian inference and RL can be combined for better performance, instead of working alone. Bayesian inference (e.g., filters for prediction) suffers from learning high-dimensional problems, while RL suffers from generalization and data efficiency. RL can embrace more methods derived from Bayesian inference to improve its algorithmic structure to suit data with complex inner architecture and nature, instead of purely increasing the network capacity and size of datasets. This review builds on the above objective and investigates Bayesian methods that have potential for agent decision making (simplified as potential Bayesian methods). Given our knowledge, this is the first paper to systematically investigate the combinations of Bayesian inference and RL for agent decision making from a meta perspective. Existing reviews related to Bayesian inference and RL for agent decision-making focus on a narrow scope to address a specific method. For example, [22] focuses on classical Bayesian RL where RL learning is rewritten to be a Gaussian process or RL is integrated with Gaussian, but the resulting solutions are impractical for complex tasks. [23] is related to classical state estimation for robotics. [24] focuses on VI and related applications. [25] is about Bayesian optimization (BO) for simple low-dimensional problems. [26] presents the evolution of Bayesian neural network (BNN). [27] describes what is active learning and provides its simple applications. [28] focuses on model-based RL. [29] focuses on RL for multi-tasking solutions, while [30] presents what is meta RL. The above reviews have a limited view by just focusing on one of Bayesian methods or RL, and few works explicitly and analytically outline the existing and potential combinations of Bayesian inference and RL. This review attempts to fill this gap to present how RL-based agent decision-making could be framed under Bayesian context.

* Laura Ruotsalainen [1,2] (laura.ruotsalainen@helsinki.fi), Chengmin Zhou[1,2] (chengmin.zhou@helsinki.fi), Ville Kyrki [1,3] (ville.kyrki@aalto.fi), and Pasi Fränti [4](franti@cs.uef.fi). Affiliations: 1. Finnish Center for Artificial Intelligence (FCAI). 2. Department of Computer Science, University of Helsinki, 00560 Helsinki, Finland. 3. School of Electrical Engineering, Aalto University, FI-00076 AALTO, Finland. 4. School of Computing, University of Eastern Finland, FI-80100 Joensuu, Finland.

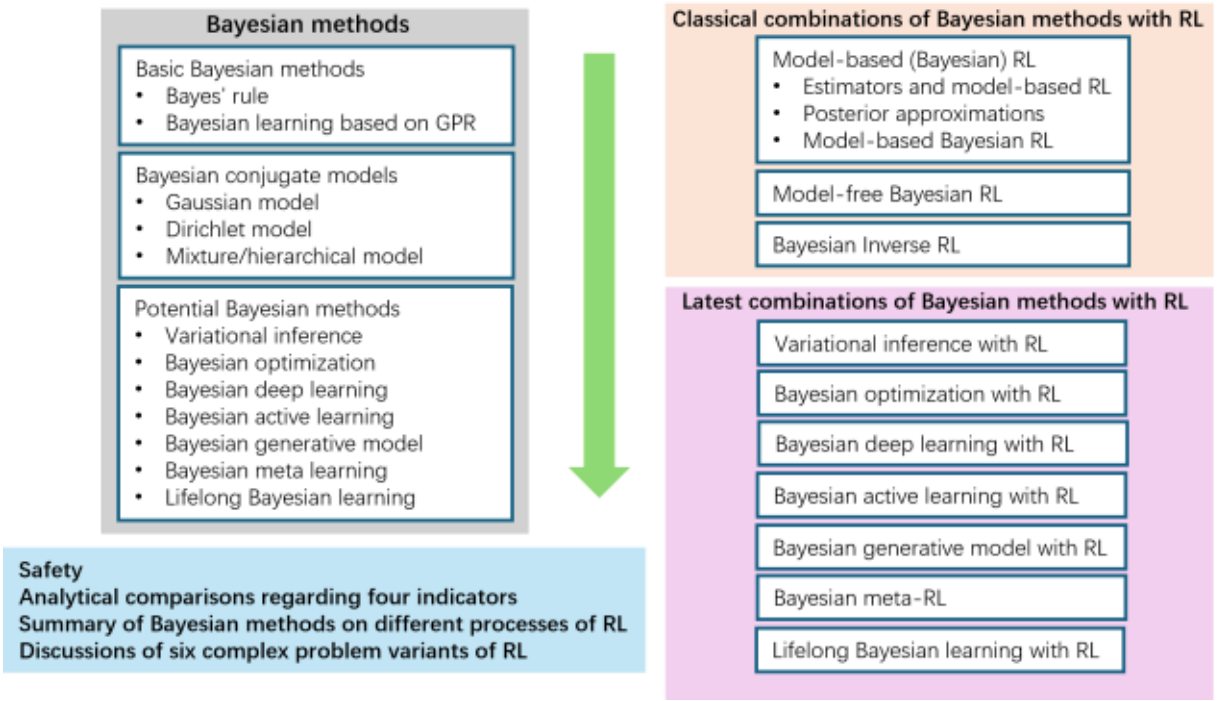

Figure 1. An overview of the paper structure.

To achieve the above goal, this review (Figure 1) starts with the introduction of Bayesian methods (**Section 2**). Bayesian methods investigated include *basic Bayesian methods and Bayesian conjugate models* and *popular Bayesian methods* consisting of VI, BO, Bayesian deep learning (DL), Bayesian active learning (AL), Bayesian generative models, Bayesian meta-learning and lifelong Bayesian learning. It is followed by recalling the *classical combination of Bayesian methods with RL* (**Section 3)**. Then, we arrive at the fundamental part of this review: *The latest combinations of potential Bayesian methods and RL* (**Section 4**) where we discuss safety aspects in an independent section (**Section 5**). Four indicators (data efficiency, generalization, interpretability, and safety) are applied to evaluate the contributions of Bayesian methods to RL for agent decision making (**Section 6**). The summary of Bayesian methods on different processes of RL are analyzed (**Section 6**). Finally, six complex problem variants of RL (unknow reward, partial observability, multi-agent, multi-task, nonlinear non-Gaussian, and HRL) for agents (e.g., robotics) are discussed analytically (**Section 7**).

## II. BAYESIAN METHODS

This section builds on *basic theories of probability* [23][31] (Supplement A1). Then, *basic Bayesian methods* are given to make readers well-prepared for complex Bayesian methods. Third, widely used *surrogate models* to capture the prevalent probability distribution from data are introduced. Finally, potential *Bayesian methods* are carefully summarized.

### *A. Basic Bayesian methods*

**Bayes' theorem (rule).** Given the joint density function $p(x,y)$ factorized by $p(x,y) = p(x|y)p(y) = p(y|x)p(x) = p(y,x)$, Bayes' theorem [32][33][34][35] is written by $p(x|y) = p(x) \cdot \frac{p(y|x)}{p(y)}$ where $y$ denotes the *observation/measurement* (evidence). $x$ denotes the *state* (hypothesis) that is normally parameterized. $p(x)$ denotes the *prior distribution function* that capsulates all priori information. The denominator $p(y)$ denotes the *probability of evidence* and is used for normalization that can be ignored generally. The *likelihood distribution function* $p(y|x)$ is also written as $\mathcal{L}(x;y)$. $p(x|y)$ denotes the *posterior distribution function*. Bayes' theorem presents a general learning process where the prior $p(x)$ is used to update the posterior $p(x|y)$ by applying the likelihood $p(y|x)$ (Supplement A2).

**Bayesian learning based on GPR.** Classical Bayesian learning builds on Gaussian processes (GP) and Gaussian process regression (GPR) [22][36][37]. GP is a stochastic process written by an unknown function $F(\cdot) \sim \mathcal{N}\left(\bar{f}(\cdot), k(\cdot,\cdot)\right)$ where $\bar{f}(\cdot)$ and $k(\cdot,\cdot)$ are the mean and covariance/*kernel* to represent the relationship of prior and posterior. GPR is a regression approach to infer GP Function $F$ (see Supplement A3). GPR provides credible intervals (covariance) for predictions (mean), providing uncertainty quantification. The limitations of GPR are two fold 1) Variables that can only take positive values cannot use Gaussian model. This problem trivializes GPs but it may be solved by the transformation via a warping function [38]; 2) GPR lacks robustness against outliers or inaccuracies in the noise distribution during uncertainty estimation, due to the absence of function evaluations that can correct the noise distribution in real-time [39]. Classical Bayesian learning is kernel-based/non-parametric and it can be extended to the parametric case.

***Kernel-based/non-parametric case.*** Bayesian learning in kernel-based/non-parametric case can be written as $Y = HF + N$ where $Y$, $H$, $F$, and $N$ are samples, transformation model, unknown function, and noise model, respectively. The learning process to figure out unknown function $F$ can be simplified as two stages [22]: 1) Preparations: *Prior* of $F$ and *transformation model* (linear) $H$ are selected. Time-series *samples* of $Y(x)$ are collected. 2) Inference: Bayes' rule is applied to infer posterior of $F$ conditioned on samples $D_T = \{(x_t, y_t)\}_{t=1}^{T}$, as GPR. See Supplement A4.

***Parametric case.*** GP function here is written as $F(\cdot) = \boldsymbol{\phi}(\cdot)^T\boldsymbol{W}$ where $\boldsymbol{\phi}(\cdot) = (\varphi_1, \varphi_2, \dots, \varphi_n)^T$ is the feature vector, and $\boldsymbol{W} = (W_1, W_2, \dots, W_n)^{\boldsymbol{T}}$ is the weight vector [22]. Hence, the linear model that describes the relationship between $F(x)$ and $Y(x)$ is written as $Y(x) = \boldsymbol{H}\,\boldsymbol{\phi}(\cdot)^T\boldsymbol{W} + N(x)$ . The randomness of GP function comes from $\boldsymbol{W}$. Assuming the prior $\boldsymbol{W} \sim \mathcal{N}(\overline{\boldsymbol{w}}, \boldsymbol{S_w})$ where $\overline{\boldsymbol{w}}$ and $\boldsymbol{S_w}$ denote the mean and variance. The posterior moments conditioned on $\mathcal{D}_T$ can be easily computed as the non-parametric case.

Fundamental limitations of GPR-based Bayesian learning include: 1) The difficulty in selecting kernel or weight size. The computing of posterior moments is heavily affected by kernel [40] that is hard to choose; $F(\cdot) = \boldsymbol{\phi}(\cdot)^T\boldsymbol{W}$ is assumed to be fixed-size that brings errors and the optimal size of $\boldsymbol{\phi}(\cdot)^T\boldsymbol{W}$ is unknown [41]. In practice, the selection of kernel is more difficult than selecting weight size. 2) Computational inefficiency with large datasets, either in kernel function or in parametric case, to directly compute the posterior. 3) Errors from linear Gaussian assumptions. Noise and prior are assumed to be captured by specific Gaussians, but their distributions are not captured by any fixed distribution. This introduces errors for computing the posterior. 4) Difficulty in handling non-linear transformation models. The transformation model is assumed to be linear, but is often nonlinear in real-world cases. This problem is expected to be alleviated by the use of the Gaussian process dynamical model (GPDM) and its variants [42] where the nonlinear transformation model is parameterized to be a linear combination of basis functions, therefore the problem turns to find the weight of basis functions. 5) Sample inefficiency in high-dimensional cases, either in kernel function case or in parametric case. The problems of computation and sample inefficiency are expected to be alleviated when the posterior is approximated by neural networks, resulting in a popular Bayesian method called

Bayesian Neural Network (BNN) [43].

### B. Bayesian conjugate models

**Gaussian model.** Probability density function (PDF) of Gaussian is $p(x|\mu,\sigma^2)=\frac{1}{\sigma\sqrt{2\pi}}\exp\left(-\frac{(x-\mu)^2}{2\sigma^2}\right)=\mathcal{N}(\mu,\sigma^2)$ where $\mu$ and $\sigma^2$ (written as $\Sigma$ in the multivariate case) denote mean and variance. Gaussian suits time continuous stochastic settings, but finds it hard to cope with skewed or multi-modal distributions (e.g., due to nonlinear dynamics) [42].

**Dirichlet model.** Dirichlet distribution is a multivariate generalization of Beta distribution [44]. PDF of Dirichlet is $p(\boldsymbol{\theta};\boldsymbol{\alpha})=\frac{1}{B(\boldsymbol{\alpha})}\prod_{i=1}^{K}\theta_i^{\alpha_i-1},\sum_{i=1}^{K}\theta_i=1$ where $B(\boldsymbol{\alpha})$ is the *multivariate Beta function* defined by $B(\boldsymbol{\alpha})=\frac{\prod_{i=1}^{K}\Gamma(\alpha_i)}{\Gamma(\sum_{i=1}^{K}\alpha_i)}$ where $\Gamma$ is the *gamma function* and $\Gamma(\mathrm{n})=(n-1)!$ . Dirichlet distribution describes the probability of $K$ outcomes (weight) $\boldsymbol{\theta}=\theta_1,\theta_2,\dots,\theta_k$, and better suits categorical/discrete settings (See Supplement A5).

**Mixture/hierarchical models.** Mixture models are often used to approximate non-Gaussian distributions (e.g., Gaussians transformed by a nonlinear dynamic model). In this case, data distribution becomes complex, making mixture models (finite/infinite) a better choice to provide computationally convenient representation of distribution. In finite mixture model, Gaussian is widely used as the component, resulting in *Gaussian mixture model* (GMM) $p(x)$ written as $p(x)=\sum_{i=1}^{n}\alpha_i p_i(x)$ where $\alpha_i$ is the weight of *i*-th mixture component, and $\alpha_i\in[0,1],\sum_{i=1}^{n}\alpha_i=1$. Multivariate Gaussian function is $p_i(x)=\frac{1}{(2\pi)^{d/2}|\Sigma_i|^{1/2}}\exp(-\frac{(x-\mu_i)^T\Sigma_i^{-1}(x-\mu_i)}{2})$ where the mean $\mu_i\in\mathbb{R}^d$. The covariance $\Sigma_i$ is a positive definite symmetric matrix. The computation of GMM is still expensive with large datasets, and *hierarchical clustering* helps reduce the computation, resulting in *hierarchical GMM* (HGMM) and variants in different datasets [45][46][47][48][49][50][51][52]. In infinite mixture models, Dirichlet is widely used as the component, resulting in *Dirichlet process mixture model* (DPMM) [53]. DPMM builds the representation for complex data with an infinite number of clusters. *Chinese restaurant process* (CRP) [54] is a popular representation of Dirichlet. It assumes an infinite number of tables (clusters) in a Chinese restaurant and aims at seating customers at tables. It is important to choose a proper parameter scaling the randomness, called temperature here, to decide the probability of generating a new cluster. Another representation of Dirichlet is the stick-breaking process (SBP) where a unit-length stick (sum of distributions) is broken into an infinite number of segments (individual distribution) given a positive scalar (concentration) [55] (see Supplement A6 for CRP and SBP).

Mixture/hierarchical models still face some challenges despite their advantages: 1) It's hard to select or learn a proper prior and a proper size of prior (number of mixture component *i*), given no/limited observations at the beginning. This problem might be alleviated by modeling the number of mixture components as an unknown parameter [56]. It is possible to learn a prior by Bayesian meta-learning where a hierarchical hyperprior is specified [57][58]. 2) In infinite settings, the trade-off of selecting the most suitable temperature parameter is hard to find. Other than widely-used Gaussian and Dirichlet, other distributions for mixture models include but are not limited to *Binomial (positive and negative)*, *multinomial*, *multivariate Gaussian or normal*, *Poisson*, *exponential* (generalized normal distribution), *log-normal* and *t-distributions* [59]. Overall, the comparisons of Bayesian models are shown in Supplement F1.

### C. Potential Bayesian methods

**Variational inference.** VI keeps being a popular tool for posterior approximation. This is achieved by 1) positing a *family* of approximation density $Q$ over hidden/latent variable $z$ where $z$ helps govern the data distribution, and 2) finding the *best member* of this family $q^*(z)$ that minimizes the asymmetric and nonnegative Kullback-Leibler (KL) divergence to the posterior $p(z|x)$ . That is $q^*(\boldsymbol{z})=\arg\min_{q(z)\in Q}KL(q(\boldsymbol{z})||p(\boldsymbol{z}|\boldsymbol{x}))$ where $\boldsymbol{z}=z_{1:n}$ and measurement $\boldsymbol{x}=x_{1:m}$ are the sets of variables. It is hard to compute $q^*(\boldsymbol{z})$ by directly minimizing $KL(q(\boldsymbol{z})||p(\boldsymbol{z}|\boldsymbol{x}))$, but it is possible by maximizing the evidence lower bound (ELBO) derived from $KL(q(\boldsymbol{z})||p(\boldsymbol{z}|\boldsymbol{x}))$ . That is $ELBO(q)=\mathbb{E}[\log p(\boldsymbol{x}|\boldsymbol{z})]-KL(q(\boldsymbol{z})\parallel p(\boldsymbol{z}))$ where the expectation is computed over $q(\boldsymbol{z})$ (see Supplement A7 for ELBO). The complexity of VI during the optimization relies on the complexity of the family. A classical family is the *mean-field variational family* defined by $q(\boldsymbol{z})=\prod_{j=1}^{m}q_j(z_j)$ where each latent variable $z_j$ is mutually independent and is governed by its own variational factor $q_j(z_j)$. Variational factors can be captured by Gaussian or categorical models. A practical VI that relies on the mean-field variational family is the coordinate ascent VI (CAVI) [60], but CAVI is computationally expensive when computing the expectation of joint distribution $\mathbb{E}_{-j}[\log p(z_j,\boldsymbol{z}_{-j},\boldsymbol{x})]$ . See Supplement A7 for CAVI. This problem can be alleviated via CAVI with Monte-Carlo sampling (MC-CAVI) to trade-off the computation and accuracy [61]. Other than MC sampling, VI can combine with natural gradients and stochastic optimization [62] to further improve its efficiency, resulting in efficient stochastic VI (SVI) [63], despite a poor interpretability.

**Bayesian optimization.** BO [64][65] focuses on finding converged posterior using as few data points as possible when the evaluation of unknown function is expensive especially in high-dimensional case. A practical example is the safety of robotics (See more in the Safety Section). BO is achieved by finding informative sample *locations/entries* $\boldsymbol{x}^*$ via $\boldsymbol{x}^*=\arg\max_{x\in\mathcal{X}}f(\boldsymbol{x})$ where the continuous objective function $f$ is modeled as an unknown random function, and $\mathcal{X}$ is the search space. The sample locations $\boldsymbol{x}^*$ are found indirectly by the *acquisition function* $a_n:\mathcal{X}\to\mathbb{R}$ that utilizes the posterior uncertainty of $f$ to guide the location search, therefore the next location $x_{n+1}$ is computed by maximizing $a_n$ . Quality of $x_{n+1}$ can be measured by the *cost function* (e.g., regret). The design/selection of the acquisition functions builds on the trade-off between the exploration of search space and the exploitation of promising avenue. Acquisition functions can be divided into four categories [25]: 1) *Improvement-based, 2) Optimistic, 3) Information-based*, and 4) *Portfolios of acquisition functions.* Improvement-based acquisition function builds on the *probability of improvement* (PI) [66] and *expected improvement* (EI) [67][68] that are used to measure the probability where a query point $x$ leads to an improvement upon the target $\tau$ (the best observation as target in practice). It works when the target is known, but suffers from aggressive exploitation that is efficient [69] but may cause suboptimal convergence. Optimistic acquisition function better balances exploration and exploitation by introducing a provable cumulative regret bound, resulting in an optimistic acquisition function in the face of uncertainty. GP-Upper Confidence Bound (GP-UCB) [69][70] is a typical example that uses mean $\mu$ and variance $\sigma$ to compute the next sample location. The optimal acquisition function is found with the decrease of uncertainty (variance), but it is hard to find a trade-off of uncertainty weight that balances the exploitation and exploration. Information-based acquisition function builds on the *entropy search* (ES) [71] that selects the point that may cause the largest reduction of entropy in posterior. This reduces the uncertainty, and avoids aggressive exploration in Thompson [72] and spectral samplings [73]. However, ES suffers from expensive computations in approximation of the posterior. The *predictive entropy search* (PES) [74] helps in reducing the computation via the symmetric property of mutual information. The max-value entropy search (MES) [75] moves further in reducing the computation via applying maximum location value to the acquisition function, resulting in a higher sample efficiency in high-dimensional problems. The portfolio or multi-acquisition function [76] builds on the objective on using multiple acquisition functions to suit all stages of tasks where each acquisition function provides a candidate location/entry. It works in complex tasks when preferred/optimal acquisition function varies during sequential optimization process. The *meta-criterion* (higher level of acquisition function) is designed to select entry point among candidates, resulting in adaptive acquisition functions. Popular examples are *GP-Hedge portfolio* [76] and *entropy search portfolio* (ESP) [77]. GP-Hedge does not consider exploration information, resulting in slow convergence of posterior, while ESP considers exploration information that improves the convergence (see Supplement A8).

**Bayesian deep learning.** BNN [78] provides point estimates and uncertainty quantification, while classical neural networks just provide black-box point estimates. In BNN, unknown/latent parameters are the model weights $\boldsymbol{w}$. Given Bayes' theorem, the posterior $p(\boldsymbol{w}|\mathcal{D})$ over the model weights is written as $p(\boldsymbol{w}|\mathcal{D}) = \frac{p(\boldsymbol{w})p(\mathcal{D}|\boldsymbol{w})}{p(\mathcal{D})}$ where $\mathcal{D}$ are observations. The likelihood $p(\mathcal{D}|\boldsymbol{w})$ is decided by the network architecture and loss functions. The prior $p(\boldsymbol{w})$ is often a zero mean Gaussian with small variance. The evidence $p(\mathcal{D})$ can be seen as a constant with respect to unknown model weights, and it can be ignored. The prediction of any quantity is the expectation of posterior $\mathbb{E}_{p(\boldsymbol{w}|\mathcal{D})}[f] = \int f(\boldsymbol{w})\, p(\boldsymbol{w}|\mathcal{D}) d\boldsymbol{w}$ which can be seen as an average of $f$ weighted by posterior $p(\boldsymbol{w}|\mathcal{D})$. However, the computing of posterior is intractable, especially in cases with non-conjugate, nonlinear, and high-dimensional models. Posterior might be approximated by Laplace approximation [79][26], Markov Chain Monte Carlo (MCMC) [80][81], and VI [24]. Laplace approximation overfits with the increase of model complexity. MCMC can provide reliable and explainable approximations, but it suffers from finding a balance between accuracy and sample efficiency with large datasets. VI like CAVI [24][60][82] sacrifices the strong *correlation* between network weights, resulting in unreliable and inaccurate approximations, while SVI [83][26][84] can provide fast and accurate approximation with the supports from reparameterization tricks [85] or dropout [86] to add the stochasticity to weights.

Practical methods in the case of deep BNN with SVI include the Bayes by backprop (BBB) [87] and MC dropout [88][89] where their priors are captured by a mixture of Gaussians. MC dropout suits large networks, sparse data, and limited time requirement for inference. It is less sensitive to prior choice and takes fewer samples and iterations for convergence, despite its latest flaw in providing consistent uncertainty estimations [90]. In contrast, BBB suits smooth data, sufficient data quantity, and sufficient time for inference. However, these two methods heavily rely on VI for posterior approximation. This causes the problems of under-estimated variance and unpredictable convergence that remain to be unsolved [26]. Above deep BNNs are based on MLP, and deep BNNs can be extended to the case of convolutional Gaussian process and the case of RNN, resulting in Bayesian convolutional neural network (BCNN) [91][92] and BRNN [93]. See Supplement A9.

**Bayesian active learning.** AL aims to select as few informative samples as possible for training, without sacrificing the training performance. This is achieved via the *acquisition function*. Given different application scenarios, AL can be divided into 1) *Membership query synthesis* where any unlabeled data can be queried by learner; 2) *Stream-based selective sampling* where independent judgement is used to decide whether each sample in the data stream should be queried; 3) *Pool-based AL* where samples in dataset are ranked and selected for training. The latter is achieved by selecting some samples first by the oracle like human to pretrain a model with supervised learning. Then, pretrained models are used to query and select more samples. The latest research focuses on pool-based AL, because it is suitable for most application scenarios and DL can be better integrated into it. Classical AL is incapable of handling high-dimensional data, but the combination of AL and DL known as deep AL (DAL) works. DAL inherits some query rules [94][95] from classical AL, like the uncertainty-based query rules. This helps improve its query rules and develop new query rules. Popular query rules/ acquisition functions include: 1) *Diversity-based*, 2) *Uncertainty-based*, and 3) *Hybrid approaches* where the balance of uncertainty and diversity is considered [27].

In diversity-based acquisition function, diversity is measured by the *similarity or redundancy* among samples. It suits datasets with rich category content and large batch size. Practical example is the density-based query rules [96]. Density-based query rules aim to construct a core subset that represents the distribution of feature space of the entire dataset. This is achieved by building an expensive *distance matrix* on the unlabeled dataset. In uncertainty-based acquisition function, the uncertainty is normally measured by *information entropy*. It suits data with small batch size and less category contents and is easy to combine with DL. Practical example is the batch mode DAL (BMDAL) [97][98] where a batch of samples is queried. Here the uncertainty is the joint *mutual information* between samples and model parameters. Joint mutual information not only considers the uncertainty of samples, but also considers the correlations among samples, unlike classical AL that ignore the correlations among samples where samples are selected one by one. Another example is the deep Bayesian AL (DBAL) [99]. BNN/BCNN with AL is a promising direction to provide better uncertainty of unlabeled samples. Once the posterior $p(\theta|X,Y)$ is obtained, for a given new sample point (entry) $x^*$, it is easy to predict its value $\hat{y}$ via $p(\hat{y}|x^*,X,Y) = \int p(\hat{y}|x,\theta)p(\theta|X,Y)d\theta$ that is the expectation $\mathbb{E}_{\theta\sim p(\theta|X,Y)}[f(x;\theta)]$ where $f(x;\theta)$ is neural network with weight $\theta$. The challenge is the pattern collapse phenomenon [100] due to VI, resulting in the overconfident prediction. DBAL is expected to be improved by combining with RL to generate tunable acquisition function [101][102][103], or by combing with meta leaning to strike a balance of sample importance between historical selected samples and newly selected samples, resulting in unbiased prediction [104], or by combining with both RL and meta-learning to make query policy more generalizable across heterogeneous datasets [105].

**Bayesian generative models.** Deep generative models include the auto-regressive model (AR) [106], generative adversarial network (GAN) [107], variational autoencoders (VAE) [108], and diffusion model [109]. This section focuses on VAE and diffusion models because they are often/directly used in agent decision making, while AR and GAN are likely to be used in language processing and image generation although GAN has potential to aid classical path planning algorithms (e.g. Rapidly-exploring Random Tree) for decision making [110].

VAE consists of the encoder and decoder. The encoder aims at projecting the high-dimensional observations to low-dimensional latent states. The decoder aims to reconstruct the observations based on the latent states. The encoder is useful to construct high-dimensional data representation via the *recognition model* $q(\boldsymbol{z}|\boldsymbol{x})$, while the decoder reconstructs the observations via the *generative model* $p(\boldsymbol{x}|\boldsymbol{z})$ that shows how good the generative model understands the observations. VAE is designed to jointly learn the recognition model and generative model. Assuming latent/true data $z$ is generated from prior $p(z)$ and observation $x$ is generated from conditional distribution $p(x|z)$, the ideal encoder is the posterior $p(z|x)$ that is not computable, but it can be approximated by variational posterior $q(z|x)$ via VI. This is achieved by minimizing $KL(q(\boldsymbol{z}|\boldsymbol{x})||p(\boldsymbol{z}|\boldsymbol{x}))$ that means $\mathbb{E}[\log q(\boldsymbol{z}|\boldsymbol{x})] - \mathbb{E}[\log p(\boldsymbol{z},\boldsymbol{x})] + \log p(\boldsymbol{x}) \geq 0$. Therefore, $\log p(\boldsymbol{x}) \geq \mathcal{L} = -\mathbb{E}[\log q(\boldsymbol{z}|\boldsymbol{x})] + \mathbb{E}[\log p(\boldsymbol{z},\boldsymbol{x})] = -KL(q(\boldsymbol{z}|\boldsymbol{x})||p(\boldsymbol{z})) + \mathbb{E}[\log p(\boldsymbol{x}|\boldsymbol{z})]$ where $\mathcal{L}$ is known as *variational lower bound*. Assuming the parameters of generative model $p$ and variational posterior $q$ are $\theta$ and $\phi$ respectively, the optimal parameters are possible to be computed by the maximum likelihood/maximum a posterior (MAP) [111] via maximizing the likelihood $p(\boldsymbol{x})$ or marginal likelihood $\log p(\boldsymbol{x})$. However, optimal parameters are not computable via maximum likelihood/MAP with large dataset, but they can be computed by maximizing the variational lower bound. Let $p(\boldsymbol{z})$ be the multivariant Gaussian. Let $p(\boldsymbol{x}|\boldsymbol{z})$ be multivariant Gaussian or Bernoulli, as that of BNN. Let $q(z|x)$ be a multivariant Gaussian with a diagonal covariance structure. After MC estimates and reparameterization trick to sample and add stochasticity to parameters, a practical variational lower bound [108] is obtained. VAE provides high-quality representations and better interpretability due to encoder and decoder, but it suffers from unexplainable VI.

Diffusion models aim to train a denoising network to predict the intractable original data distribution $p(x_0)$ by three processes: 1) *Forward*, 2) *Reverse*, and 3) *Sampling* processes. Forward process perturbs intractable distribution $p(x_0)$ to tractable terminal distribution $p(x_T)$ (e.g., Gaussian normal) via forward transition $p(x_t|x_{t-1})$ where $T$ is the number of timesteps. After the forward process, reverse process starts to remove the noise recursively from terminal distribution where the reverse transition $p_\theta(x_{t-1}|x_t)$ is approximated by $\mathcal{N}(x_{t-1}; \mu_\theta(x_t,t), \Sigma_\theta(x_t,t))$. $\mu_\theta$ and $\Sigma_\theta$ are the mean and variance respectively predicted by a denoising network $\theta$. The forward and reverse processes are written by $p(x_t|x_0) := \prod_{t=1}^{T} p(x_t|x_{t-1})$ and $p_\theta(x_0) := p(x_T)\prod_{t=1}^{T} p_\theta(x_{t-1}|x_t)$ where noise $\epsilon_t$ is added between two timesteps and the denoising network is trained by minimizing the divergence between $p(x_0)$ and $p_\theta(x_0)$. Reverse optimization process is intractable, therefore VAE is applied to minimize the variational lower bound. After a converged denoising network $\theta^*$ is obtained, intractable distribution $p(x_0)$ can be predicted by $p_{\theta^*}(x_0)$ via sampling process. Practical variants of diffusion models are the denoising diffusion probabilistic model (DDPM) with noise schedule in discrete setting [112] and the continuous variant with stochastic differential equation (SDE) [113]. Diffusion models have five challenges: 1) *Data representation*, 2) *Noise schedule* (how much noise should be added), 3) *Selection of terminal distribution*, 4) *Parameterization of denoising network*, and 5) *Guidance* of denoising/sampling process. Real-world data is often high-dimensional and multi-modal, and it is important to represent data to shorten the forward and reverse processes and to find a trade-off between accuracy and computation via latent representation with encoder-decoder frameworks (e.g. U-Net/Transformer [114][115] as approximation of denoising network). A suitable noise schedule results in balanced exploration and exploration, but it is hard to find. A varying/adaptive terminal distribution will shorten the forward and reverse processes, but it requires prior knowledge. A good parameterization of denoising network accelerates the convergence, but it requires prior knowledge for the predictions of unknown $x_0$, noise $\epsilon_t$ or the gradient of two timesteps (score) [109]. A flexible incorporation of denoising/sampling guidance contributes to the generation speed and quality of predicted data ${x_0}^*$ for unseen scenarios where the vanilla, classifier and classifier-free guidance are mostly used. Vanilla guidance directly applies conditions $c$ to guide the denoising and it is ineffective. Classifier guidance trains an extra classifier to modify the unconditional denoising direction and it is unstable and expensive, while the classifier-free guidance is a mixture of unconditional model and vanilla guidance to find an acceptable trade-off between computation and accuracy. Moreover, it is also possible to select an intermediate timestep with extra network (truncation), a subset of timesteps (selection strategies) and knowledge distillation to fasten the denoising [109].

**Bayesian meta learning.** Meta learning is known as *learning how to learn*. It is data-efficient and learned parameters/models are generalizable to other unseen tasks. Bayesian meta learning (BML) is meta learning with VI (e.g., stein variational gradient descent (SVGD) [58]) that makes the computation of parameters easy. Meta learning generally consists of *black-box (recurrent) approaches* and *optimization-based approaches*. BML is based on optimization-based approaches. Black-box approaches can combine task inference methods based on Bayesian inference for better performance (see Section IV-F).

Optimization-based approach formulates gradient-based meta learning as the hierarchical Bayes. Let global latent variable of BML be the meta model $\theta$. Let the distribution of each subset or task (local latent variable/local model) be $\phi_i$ where $\phi_i$ depends on $\theta$ ($\phi_i$ is initialized from $\theta$). Let dataset of each task be $\mathcal{D}_i$. The objective is to maximize the marginal log-likelihood $\log\prod_{i=1}^{M} p(\mathcal{D}_i)$ to compute the optimal parameters, but it is intractable. It is possible to maximize the variational lower bound of marginal log-likelihood $\mathcal{L}(\psi,\lambda_1,\ldots,\lambda_M)$ to compute optimal parameters via $\log\prod_{i=1}^{M} p(\mathcal{D}_i) \geq \mathcal{L}(\psi,\lambda_1,\ldots,\lambda_M)$ where $\psi$ and $\lambda_i$ are the variational posterior approximations of $\theta$ and $\phi_i$, respectively. $M$ is the number of tasks or episodes (minibatch). This process faces the challenge of two-layer optimizations, and it might be solved by the *inner* loop and *outer* loop to compute local variables and global variable, respectively. Two-layer optimization is holding $\theta$ constant (meta model initialization) and computing $\lambda_i$ on the fly with amortized VI [116], and then $\theta$ is used to compute each $\lambda_i$ from $\mathcal{D}_i$ where the convergence is sensitive to meta model initialization (see Supplement A10 for amortized VI). Optimization-based approaches suit low-dimensional cases, e.g., parametric/simple neural network cases in model-agnostic meta learning (MAML) [117], domain-adaptive meta-learning (DAML) [118], and Bayesian MAML [58].

Black-box approaches directly use deep black-box neural networks [119][120], such as RNNs [121] to compute all local latent variables $\phi_i$ first and then $\phi_i$ are used to compute losses for updating $\theta$. Specifically, the objective of black-box approaches is defined by $\max_{\theta}\sum_{T_i}\sum_{(x,y)\sim\mathcal{D}_i^{test}}\log f_{\phi_i}(y|x)$ where $(x,y)$ is the input/output pair. $T_i$ is a task index. This objective is intractable. However, $\phi_i$ are the local latent variables that are generated directly by *one* deep neural network $f_\theta$ via $\phi_i = f_\theta(\mathcal{D}_i{}^{train})$. Global latent variable $\theta$ is maintained in $f_\theta$, and $\phi_i$ is updated via backpropagation once new dataset is fed to network $\theta$ where the optimality of $\phi_i$ is not guaranteed. $\mathcal{D}_i$ splits into training dataset $\mathcal{D}_i{}^{train}$ and test dataset $\mathcal{D}_i{}^{test}$. *Another* neural network $f_{\phi_i}$ receives $\phi_i$, and log-likelihood $L(\phi_i,\mathcal{D}_i{}^{test}) = \sum_{(x,y)\sim\mathcal{D}_i^{test}}\log f_{\phi_i}(y|x)$is used as the loss between $\phi_i$ and $\mathcal{D}_i{}^{test}$ to update global latent variable $\theta$. Therefore, practical objective of black-box approaches can be rewritten as $\max_{\theta}\sum_{T_i}L(f_\theta(\mathcal{D}_i{}^{train}),\mathcal{D}_i{}^{test})$. Overall, black-box approach suits high-dimensional problems, but it is data inefficient when learning from scratch because of the suboptimal update of $\phi_i$ and deep networks. The performance of $\phi_i$ relies on the convergence of $\theta$ because $\phi_i$ is adapted from $\theta$ via $\phi_i = f_\theta(\mathcal{D}_i{}^{train})$, and the update of $\theta$ relies on the performance of $\phi_i$ to compute the loss via $L(\phi_i,\mathcal{D}_i{}^{test})$. Therefore, converged $\theta$ and $\phi_i$ are more similar (tight relationship). This results in a good specialization of $\theta$ in test tasks with similar distributions but has poor generalization in tasks with large distribution differences and tasks that are out of distribution (OOD). Optimization-based approaches suit low-dimensional cases where local variables and the global variable have a "loose" relationship because they are updated in separated processes (inner/outer loops). This results in good generalization among tasks with different distributions, but the convergence is sensitive to meta model initialization.

**Lifelong Bayesian learning.** Unlike classical lifelong learning that focuses on designing adaptive weights of neural network [122][123][124], lifelong Bayesian learning builds on the infinite mixture model as global model to enable an incremental learning for infinite number of local polices. This is achieved by maintaining a two-layer policy hierarchy. The bottom layer is model-based policy like Gaussian/model-free policy, while the top layer is the infinite mixture model like DMPP [125][126][55] to cluster data of different distributions.

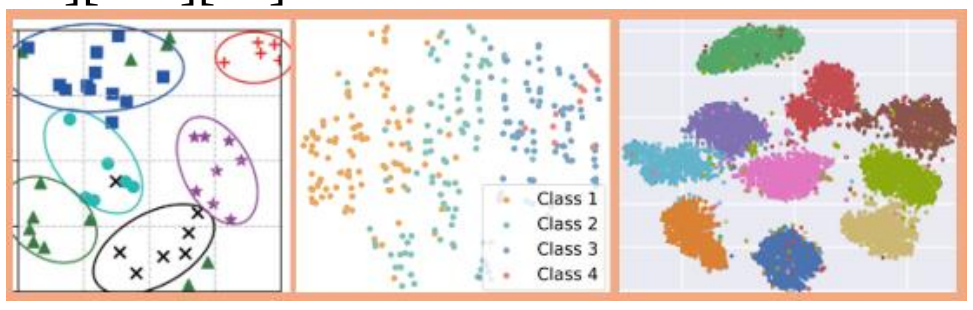

Fig. 2. The clustering examples of DPMM given EM [126], MCMC [127] and VI [55] where the dots denote the tasks or data from different distributions.

However, DMPP cannot be applied for practical purposes due to intractable posterior computation. The posterior can be approximated via online Bayesian inference (e.g., expectation and maximization (EM)) [125][126], MCMC (e.g., Metropolis-Hasting/Gibbs [127]), and VI [55]. Therefore, the convergence of lifelong Bayesian learning relies on the performance of approximations. EM converges slowly, resulting in poor data quality of bottom-layer policies. MCMC suffers from the trade-off of computation and accuracy. VI converges fast but results in an unreliable/unexplainable optimization process. The clustering examples of DPMM given EM, MCMC and VI are shown in Fig. 2. Overall, we compare basic and potential Bayesian methods in Supplement F2.

## III. CLASSICAL COMBINATIONS

This section addresses classical combinations of Bayesian methods with model-based and model-free RL. In model-based part (Section 3.1), we first summarize the *model-based optimal control/model-based RL* given estimators and *typical posterior approximation methods*. Then, we give summary for *model-based Bayesian RL*. The *model-free Bayesian RL* and *Bayesian inverse RL* are addressed in Sections 3.2 and 3.3 respectively.

### *A. Bayesian methods for model-based RL*

**Estimators and model-based RL**. Model-based optimal control builds on state estimators. With the estimators (dynamics), the optimal control problem can be formulated to be model-based RL problem given the feedback (reward/cost function) from the environment (Fig. 3).

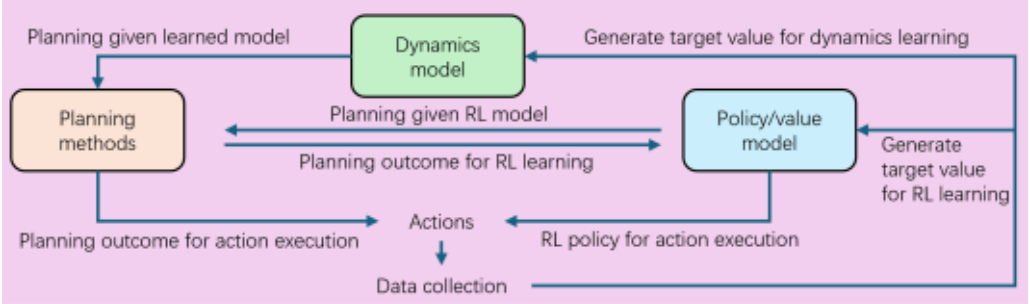

Fig. 3. Model-based RL. Classical model-based RL just includes the learning of dynamics model and RL model, while the latest model-based RL include the planning additionally.

The *linear* and *nonlinear* cases are considered when computing estimators where GPR is used for linear-Gaussian estimations and Bayes filters for nonlinear estimations. Given the property of dataset, there are two situations to access dataset: 1) Access a batch of data with or without the time-sequential dependency (*batch discrete-time* case). 2) Recursively access time-sequential data (*recursive discrete-time* case). The estimator/posterior is easy to compute by directly applying GPR in linear batch discrete-time case [23]. However, the future data is used to infer the past state here, but future data is inaccessible when inferring the states online. GPR is also appliable for linear recursive discrete-time case [23]. A practical example is Kalman filter (KF) [128][129]. KF suits low-dimensional problems with cheap computation, but it builds on linear transformation model and suffers from accumulated errors of each step. This results in inaccurate and unstable estimations (see Supplement B1 for GPR). In nonlinear recursive case where the transformation model (motion and observation models of robots) is nonlinear, the recursive discrete-time estimation is easy to achieve by Bayes filter [130][131][132][133][134]. The system is assumed to have *Markov property*: A stochastic process has the Markov property, if the conditional PDFs of future states depend only upon current state, and not on any other past states. Bayes filter [135][136][137][138] seeks computing the posterior belief $p(x_k|\check{x}_0, v_{1:k}, y_{0:k})$ where $x_k$, $y_{0:k}$, and $v_{1:k}$ are motion, observation, and control input respectively. Posterior belief can be factored by Bayes rule and Markov property and solved by approximations. Specifically, Bayes filter is intractable since it has infinite-dimension *space of PDF* which requires infinite memory to represent the posterior belief and the computing of integral requires infinite computing resources. To overcome this problem, 1) the space of PDF can be *constrained* by surrogate models like Gaussian; 2) motion/observation models can be simplified via *linearization*; 3) the integral can be *approximated via sampling* to facilitate the computation. The linearization results in Extended KF (EKF) [139] and Iterative EKF (IEKF) [23]. The approximation via deterministic sampling results in the Sigmapoint KF (SPKF) and iterated SPKF (ISPKF) [23]. The approximation via *the importance sampling method* based on MC sampling [140][141] results in the particle filter (PF) and its variants [142][143][144]. It is possible to apply MCMC for better approximation of PF. To summarize, regarding approximations of Bayes filter in nonlinear recursive case, existing linearization in EKF brings errors that result in inaccurate estimations. Deterministic sampling in SPKF suffers from finding a suitable user-definable parameter that scales how far away Sigmapoints are from the mean, affecting the overall accuracy. MC/MCMC in PF suffers from finding the trade-off of accuracy and computation. VI is efficient in approximations but it suits high-dimensional problems, therefore it is not a suitable choice here. In nonlinear batch discrete-time case, Bayes filter is based on Markov property and cannot be extended to the batch discrete-time case. However, Bayesian inference (rules) can be used to the batch discrete-time case for nonlinear estimation. It shares the same steps as that in linear-Gaussian estimation after the *linearization* of motion and observation models. Bayesian inference can be extended to the continuous-time estimation [145] where the mechanism is almost the same as that in the discrete-time estimation because Bayesian inference in discrete-time case can be seen as a special case of that in continuous-time case. More generally, the posterior can be directly computed via classical statistical methods like the maximum a posterior (MAP) using Gauss-Newton [23], Maximum Likelihood via Gauss-Newton [111][23], and sliding-window filters (SWFs) [146] for linear/nonlinear cases (see Supplement B2 for Bayes filter, MAP, maximum likelihood, and SWFs).

With state estimators (dynamics model), if the feedback (reward/cost function) from the environment is known, it is possible to achieve optimal control that can be formulated to be model-based RL [147][148] as Fig. 4, while the optimal control under constraints and nonlinear dynamics still remains difficult. Classical model-based RL includes the learning of dynamics model and RL model where these two models are trained simultaneously. This brings the challenge of two-layer optimization where RL is the global model and dynamics is the local model, and results in slow convergence. The latest model-based RL incorporates function approximations and planning methods [147]. Function approximations with neural networks enable model-based RL to learn in high-dimensional case. Planning methods consider the information from the learned dynamics model and RL model (policy/value) to generate planning outcomes. The planning outcome can directly improve RL learning, while it also improves RL and dynamics training indirectly by guiding the generation of data. This results in high-quality target value for better convergence.

**Posterior approximations.** This section summarizes approximations and their connections. In the above sections, we simply mentioned linearization (see Supplement B3 for linearization) and sampling methods, while sufficient descriptions were given to VI in Section 2.3. The emphasis of this section is on sampling and connections of sampling, VI and EM. These approximations are not just for approximating the posterior of Bayes filter, but also for more complex posteriors in the following sections.

*Sampling.* MC sampling follows a particular pattern: 1) Define the domain of inputs and generate inputs randomly from a probability distribution over the domain; 2) Perform a deterministic computation like counting on the inputs; 3) Aggregate the results for final approximation. When passing a PDF through a nonlinearity, many samples are first generated from the input density. Second, each sample is transformed through non-linearity. Third, the output density is built from the transformed samples. Overall, MC sampling suits any PDF, any non-linear function without the need to know the mathematical form, but it faces two challenges: 1) More samples result in higher accuracy but lower computation speed (trade-off of computation and accuracy); 2) The means of input and output are the same in linear case, while the means may change in nonlinear case. MC sampling grounds Sigmapoint KF (SPKF) where the samples are generated deterministically (deterministic sampling). SPKF can be improved by iterated SPKF (ISPKF) [149] by computing input Sigmapoints around an operating point. However, the user-definable parameter $\kappa$ in SPKF, which scales how far away the Sigmapoints are from the mean, is hard to define, resulting in poor accuracy. SPKF and PF are classical examples where MC sampling is used to solve the nonlinear Bayesian filter. The samples in SPKF are generated deterministically because the Sigmapoint transformation is deterministic, while the samples in PF are generated by sampling the importance sampling (resampling process). PF is more accurate if there are as many samples as possible, but it is computationally expensive (See Supplement B4 for MC). MCMC builds on the rejection algorithms where samples will be accepted if the acceptance condition satisfied [150]. MCMC sampling algorithm can be seen as the combination of the rejection sampling algorithm, Markov chain, and MC sampling. MH algorithm [80] and *Gibbs sampler* [81] are two representatives of MCMC sampling. MH suffers from finding a balanced covariance of self-designed proposal distribution like Gaussian for sampling. Gibbs sampler suits case of $n$-dimensional $\boldsymbol{x}$ where $\boldsymbol{x} = (x_1, x_2, \dots, x_n)$ and PDF $p(x)$ in MH algorithm is replaced by *full conditions* expressed by $p(x) = p\left(x_j \middle| x_1, \dots, x_{j-1}, x_{j+1}, \dots, x_n\right) = p\left(x_j \middle| x_{-j}\right), j = 1, \dots, n$ where $x_{-j} = (x_1, \dots, x_{j-1}, x_{j+1}, \dots, x_n)$ is the supplement of $x_j$ given $n$-dimensional $\boldsymbol{x}$ [151], and $p\left(x_j \middle| x_{-j}\right), j = 1, \dots, n$ include $n$ *joint* distributions. If the full conditions are available and belong to the family of standard distributions like Gaussian, the new samples can be drawn *directly.* The high-dimensional joint distributions (conditions) $p\left(x_j \middle| x_{-j}\right), j = 1, \dots, n$ can be factorized into a directed graph, resulting in the directed acyclic graphs [152] which makes Gibbs sampling practical for the approximation of Bayesian inference (see Supplement B5 for rejection algorithm, Markov chain, MH, and Gibbs).

*Connections of sampling, VI and EM.* VI is an alternative to MCMC. Their key difference is that MCMC approximates the posterior models via *sampling* with theoretical *guarantee* to approximation accuracy, while VI via the *optimization* without guarantee. VI is *faster* than MCMC to handle the large dataset and complex posterior models like Gaussian mixture model and conditionally conjugate model, but VI *underestimates* the variance of the posterior [24]. EM [153] algorithm is a classical iterative algorithm in maximum likelihood and MAP to estimate the local optimum of the parameters [154]. It is a practical but expensive algorithm built on Bayesian rules to compute the posterior. EM can be more efficient once it combines with the MCMC. EM can also be extended to the variational Bayes for cases with *large dataset* and *complex posterior models*. Given the variables $\{x, z\} \in \mathcal{X}$ where $x$ and $z$ denote the *observable* and *hidden (latent)* parts of the variables respectively, the local optimum of the likelihood $p(x|\theta)$ can be found by iterating the E step and M step respectively, to compute the expectation of complete log likelihood $\log p(x, z|\theta)$ via E: $Q(\theta) = \int [\log p(z, x|\theta)] p(z|x, \theta^{old})\, dz$ and M: $\theta^{new} = \arg\max_{\theta} Q(\theta)$ where $\theta$ is captured by e.g., Gaussian, and $\theta^{old}$ the parameter value at previous time step. However, the computing of expectation in E step is intractable, and popular way to solve this is to approximate $p(z|x, \theta^{old})$ by sampling methods like MC and MCMC, resulting in the stochastic EM [155], MC-EM [156], and MCMC-EM [157]. For example, after MCMC (e.g., HM), EM is simplified to $E: \hat{Q}(\theta) = \frac{1}{N_i} \sum_{j=1}^{N_i} \log p(z^{(j)}, x|\theta)$ and $M: \theta^i = \arg\max_{\theta} \hat{Q}(\theta)$ where $N_i$ denotes the number of samples in $i$-batch, and $\hat{Q}(\theta)$ the estimated value of complete log likelihood. It is interesting to notice that 1) $ELBO(q) = \mathbb{E}[\log p(\mathbf{z},\mathbf{x})] - \mathbb{E}[\log q(\mathbf{z})]$ where $\mathbb{E}[\log p(\mathbf{z},\mathbf{x})]$ is the objective of EM in E step; 2) Given $\log p(\mathbf{x}) = KL(q(\mathbf{z})||p(\mathbf{z}|\mathbf{x})) + ELBO(q)$, $ELBO(q) = \log p(\mathbf{x})$ when $q(\mathbf{z}) = p(\mathbf{z}|\mathbf{x})$. Given the above two facts that reveal the connection of EM and VI, VI can be seen as the extension of EM when maximizing the likelihood over the latent variables, but they have the difference: EM maximizes the expected complete likelihood $\mathbb{E}[\log p(\mathbf{z},\mathbf{x})]$, while VI tries to find a balance between $\mathbb{E}[\log p(\mathbf{z},\mathbf{x})]$ and $-\mathbb{E}[\log q(\mathbf{z})]$ in the maximization, or VI maximizes the log marginal likelihood $\log p(\mathbf{x})$ when $q(\mathbf{z}) = p(\mathbf{z}|\mathbf{x})$.

**Model-based Bayesian RL.** Dynamics-based model-based RL works well in simple/low-dimensional problems, but poor in medium/large-scale problems unless the function approximations with (Bayesian) neural networks are applied [147]. Model-based RL can be further improved by enjoying the advantages of the *large transition matrix*, resulting in model-based Bayesian RL. Model-based Bayesian RL [22][158][147] builds on RL, Markov decision process (MDP), and partial observable MDP (POMDP) [22] (See Supplement B6). The basic idea of model-based Bayesian RL is that the agent uses the collected data to first build a model of the domain's dynamics and then uses this model to optimize its policy. A practical example is the Bayes-adaptive MDP (BAMDP). BAMDP [22][159][160] is defined by $< S', A, P', P_0', R' >$ where $S'$ is the hyper-states (information states) $\mathcal{S} \times \Phi$. $P'(\cdot|s,\phi,a)$ is the transition distribution between hyper-states. Reward $R'(s,\phi,a) = R(s,a)$. Transition function $\Phi$ is the posterior parameter. $\phi$ a realization of transition function $\Phi$ and $\Phi = \{\phi_{s,a}, \forall s,a \in S \times A\}$. BAMDP assumes the structure form of posterior to be Dirichlet to simplify expensive computation of transition function in POMDP (See Supplement B7), but BAMDP still faces the problem of two-layer optimization as that of model-based RL.

The advantages of BAMDP are two folds: 1) It better handles the complex small-scale real-world problems than the dynamics of model-based RL; 2) The policy of BAMDP is expressed over the information state which includes model uncertainty. This makes BAMDP more explainable than other RL algorithms like the model-free RL. However, the computation of integrals in the value iteration is still expensive. Another critical problem of the BAMDP is the *exploration and exploitation dilemma*. That is, the balance of short-term information (instant reward) and long-term information which results in higher future rewards. Therefore, the *value approximation methods* and *exploration bonus methods* are required for posterior approximation and policy learning. Here the value approximation methods investigated includes the *offline value approximation*, *online near-myopic value approximation*, and *online tree search approximation*. The *finite-state controllers* and *Bayesian exploration exploitation tradeoff in learning* (BEETLE) [22] are two representatives of the offline value approximation that requires much computation resources. The online near-myopic value approximation just *accesses fewer hyper-states to find the policy*. Therefore, it requires fewer computing resources, but causes suboptimal convergence of value estimation. *Bayesian dynamic programming* and *the value of information heuristic (1-step estimation)* [22] are two representatives of the online near-myopic value approximation methods. Online tree search approximation methods are typically based on the classical *forward search tree* or the *MC tree search* (MCTS) [161], resulting in the Bayesian adaptive MC planning (BAMCP) [162][163]. Compared with the online near-myopic value approximation, BAMCP better improves the convergence with reasonable computation. Regarding exploration Bonus methods, Bayesian exploration bonus (BEB) [164] integrates the exploration bonus to the value function. This makes BEB approach the optimal Bayesian solution $V^*(s,\phi)$ faster. Exploration bonus can be a variance-based reward bonus (VBRB) [165] where exploration bonus is decided by posterior variance that bounds the exploration complexity to approach optimal Bayesian solution $V^*(s,\phi)$ faster. See Supplement B8.

### B. Bayesian methods for model-free RL

Classical combination of Bayesian methods with model-free RL results in model-free Bayesian RL that directly learns the policy from collected data without learning transition/dynamics model. The advantage of model-free Bayesian RL over model-based (Bayesian) RL is its efficiency in cases where the solution space (e.g., policy space of robot's behavior) exhibits more regularity (stability) than the underlying dynamics/transition. Regularity means the mapping function $f: a \leftarrow f(s)$ is stable and unique. In cases with large samples and less noise, model-free RL exhibits more regularity, while in small samples with noise, model-based (Bayesian) RL exhibits more regularity. Model-free RL suits the large-scale real-world problems than the model-based (Bayesian) RL, but it must manage the exploration-exploitation trade-off. Classical model-free RL like the temporal difference learning [166], policy gradient (PG) [167] and actor-critic (AC) [168] can be integrated to the Bayesian framework, resulting in the *value function Bayesian RL*, *Bayesian PG (BPG)*, and *Bayesian AC*.

A typical representative of value function Bayesian RL is GP temporal difference learning (GPTD) [22][169][170][171] that builds on GPR because its value function can be rewritten as GP via $\boldsymbol{R_{T-1}} = \boldsymbol{HV_T} + \boldsymbol{N_{T-1}}$ where *R, V, N,* and *H* denote the reward, value, noise, and transformation model respectively. The posterior moments can be computed and approximated using GPR and MC sampling. BPG requires expensive computation of the integral. This problem can be solved by the Bayesian quadrature (BQ) which is a Bayesian approach to evaluate the integral via the observed samples. BQ [172] solves the computing of integral by simplifying the integral $\zeta$ to $\zeta = \nabla\eta(\theta) = \int f(x)g(x)\,dx$ where *unknown* $f(x)$ can be solved via GPR to infer its posterior. $g(x)$ is a *known* function. Given BQ, the gradient of BPG $\nabla\eta(\theta)$ [22][172] can be easily computed with lesser computation resources, but large variance is introduced due to trajectory-based policy gradient. Bayesian AC [173][174] is the combination of GPTD and BPG. Given GPTD, it is easy to obtain the posterior moments of critic (value function), while policy (actor) gradient is computed via BQ as BPG. Bayesian AC makes the best of Markov property of trajectory by step-based expected return (despite the bias introduced by one-step prediction in Q value), instead of the trajectory-based expected return in BPG. This reduces the variance of policy update; therefore, Bayesian AC is data efficient than BPG. Actor-critic architecture contributes to convergence; therefore, Bayesian AC is data-efficient than GPTD. See Supplement B9.

### C. Bayesian methods for inverse RL

IRL [175][176] aims to find reward function $r(s,a)$ given the expert demonstrations $Z = \{\zeta_1, \zeta_2, ..\}$ where $\zeta_i = \{(s_t, a_t)\}_{t=0}^T$ generated by optimal policy $\pi^*$. Reward function can be formulated by $r(s,a;\boldsymbol{f})$ parameterized by feature vector $\boldsymbol{f} := [f_1, f_2, .., f_N]^T$. The challenge of IRL under MDP is how to predict the reward function $\boldsymbol{f}$ with dataset $\mathcal{D} := (\boldsymbol{y}, \boldsymbol{\mathcal{X}})$ where $\boldsymbol{\mathcal{X}}$ is the observed input and $\boldsymbol{y}$ the noisy reward feature defined by $\boldsymbol{y} := [f_1 + e_1, f_2 + e_2, .., f_N + e_N]^T$. Assuming noise $e_i$ is independent and identically distributed like Gaussian white noise $e_i \sim \mathcal{N}(0, \sigma^2)$. The reward feature can be treated as the realization of GP $f(x) \sim GP(\mu(x), k(x, x'))$. Hence, the objective of Bayesian IRL [177] is to acquire the estimated posterior of reward function $\hat{f}(x)$ (e.g., learning human behavior for robotic imitation). This is achieved via the expectation of prior $f(x)$ given the dataset $\mathcal{D}$ by $\hat{f}(x) = \mathbb{E}(f(x)|\mathcal{D})$ where $f(x)$ and $e_i$ are the Gaussians. To select better kernels, the kernel of $f(x)$ can be parameterized and stabilized by maximizing its log likelihood and $l_1$-regularizer (See Supplement B10). In complex non-linear non-Gaussian case, the approximation methods like the MCMC play an important role to approximate the posterior [178][179]. When the parameter space is high-dimensional, BO is applied to simplify the parameter space via projecting the parameters to a single point in a latent space. This ensures that the nearby points in latent space correspond to reward functions yielding similar likelihoods. The comparisons of Bayesian methods on classical RL are in Supplement F3.

## IV. LATEST NOVEL COMBINATIONS

In this section, we investigated how to integrate seven potential Bayesian methods into RL to further improve RL for more complex scenarios.

### *A. Combining variational inference with RL*

There are less cases where VI can directly derive RL policy, but VI is an efficient tool to *1) assist deriving RL policy* and *2) assist approximating reward function*. Specifically, VI can assist other Bayesian methods like BNN [43] and VAE [19][20][180] to derive RL policy via maximizing ELBO with proper variational distributions. For example, [19][20] model the system as probabilistic graphical model (PGM) and compute the optimal parameters of PGM by maximizing the likelihood where the reward (optimality) is a part of data. It is hard to compute the likelihood directly, but ELBO can be derived. The maximizing of ELBO results in the learning of generative models and RL policy. VI can be used to predict the extrinsic reward of RL agent [181]. Agent's reward model can be defined by $y = g(f(x))$ where *feature model* $f(\cdot)$ is related to deriving time-step importance of inputs and is modeled as GP to extract features that capture the correlations between time steps and those across different episodes. $g(\cdot)$ is an explainable *reward prediction model* which infers the distribution of final rewards y and deliver time step importance. It can be modeled as categorical distribution or GP. Direct inference of reward prediction model $y = g(f(x))$ is computationally expensive. This can be solved by VI which simplifies the maximizing of the log marginal likelihood to the maximizing of ELBO. Reward function can be intrinsic reward like Bayesian surprise. In this case, the maximization of ELBO on *log-likelihood of future states* [182] will result in the information gain $\mathcal{I}$ approximated by Bayesian surprise that can be taken as the intrinsic reward. Given the above two applications of VI (see Supplement C1), we can easily conclude that when the maximization of likelihood is expensive, we'd better turn to VI and maximize ELBO for faster convergence and cheap computation. This assists deriving RL policy and reward functions (extrinsic and intrinsic), but the optimization process is unreliable due to VI. Future research may focus on designing explainable VI to better contribute to efficient and explainable RL.

### *B. Combining Bayesian optimization with RL*

BO can *1) tune the hyper-parameter* and *2) reduce policy space of RL* to accelerate convergence. Specifically, [183] aims to find the optimal $d$-dimensional hyper-parameters $x^*$ (e.g., learning rate) in RL via $x^* = \arg\max_{x \in X} f(x, T_{max})$ where unknown function $f$ is captured by Gaussian. The objective is to make the training cost $\sum_{i=1}^{N} c(x_i, t_i)$ of evaluated settings $[x_i, t_i]$ as low as possible. This is achieved by evaluating training curves via BO to find the next optimal point $z_n$ that will result in $x^*$ where $z_n = [x_n, t_n] = \arg\max_{x \in X, t \in [T_{min}, T_{max}]} \alpha(x, t)/\mu_c(x, t)$ and $\alpha(x, t)$ is an expected improvement acquisition function. $\mu_c(x, t) = \beta^T [x, t]$ is the mean of cost where $\beta = (Z^T Z)^{-1} Z \boldsymbol{c}, Z = [x_i, t_i], \boldsymbol{c} = [c_i]$. The dataset for evaluation (training curves) is big with much redundant data, therefore AL [27][183] is first applied to compress dataset to provide informative dataset for BO. BO with GP-UCB is a popular but low-efficient way to reduce the parameter space and find the optimal parameters, as in [184]. [185] constrains the policy search space of RL to a sublevel-set $C_n$ of GP's uncertainty, such that $\theta_{n+1} = \arg\max_{\theta \in C_n} \alpha(\theta), C_n = \{\theta | \sigma_n(\theta) \leq \gamma \sigma_f\} \cap \Theta$ where $\theta_{n+1}$ is the next policy parameter. $C_n$ is the *confidence region* defined by a ball centered around $\theta_0$ with radius $r_0 = \|\theta - \theta_0\|$ determined by the condition $\sigma_0(\theta; r_0) = \gamma \sigma_f$. $\alpha(\theta)$ is the acquisition function. $\sigma_f$ is the signal standard deviation of GP's kernel. $\gamma \in (0,1]$ is a tunable parameter which governs the effective size of $C_n$. $\Theta$ is the parameter space. Confidence region $C_n$ can grow adaptively after each rollout of the system in a cautious manner via $\|J(\theta_i) - J(\theta_j)\| \leq L_J \|\theta_i - \theta_j\|, \theta_i, \theta_j \in C_n$ where $L_J$ is the Lipschitz continuity. Once the algorithm finds a local optimum, it explores the surrounding region until the global optimum of the policy. [186] applied BO with EI-based acquisition function on HRL to assist learning high-level subtasks. This is achieved by learning the value function at a *finite set* of states where state size is reduced via BO. Instead of using BO once, [187] uses two-stage BO for search space reduction. The first stage uses BO to create a reduced space (hyperplane), while the second stage takes another BO (knowledge gradient policy) with the PF to further reduce the search space and find the optimal parameters. Given the above two applications of BO, we can notice that in future research when the optimization problem is high-dimensional, it is possible to apply AL or BO to reduce the size/dimension of dataset to provide informative dataset. Then, BO is applied to efficiently search the optimal parameters like the hyper-parameter and policy parameter.

### *C. Combining Bayesian neural network with RL*

The advantage of BNN/BRNN with probabilistic backpropagation [93] over classical neural networks is the uncertainty estimates. BNN can be used in two different ways: *1) uncertainty estimates for RL policy exploration* and *2) replacing simple dynamics of model-based RL to suit high-dimensional cases*. Specifically, [188] applies BNN with MC dropout and deep ensemble as the *exploration policy* to compute the mean $\mathbb{E}_{q(Q|s_t, a_t)}(Q)$ and variance $\sigma^2_{q(Q|s_t, a_t)}(Q)$ of critic. Then, the state with highest variance in candidate states will be selected as the next state via $\arg\max_i \sigma^2_{q(Q|s_i, a_i)}(Q)$. This results in a policy that enable agents to avoid risky areas, but it is data inefficient. Instead of directly exploring the state/action space via the uncertainty [188], [189] formulates the exploration process as a *curiosity-driven exploration* where agents take actions that maximize the reduction of uncertainty about the environment dynamics. This is equivalent to maximizing the information gain (mutual information between the *next* state distribution at $s_{t+1}$ and the model parameter $\Theta$ ) defined by $I(s_{t+1}; \Theta | \xi_t, a_t) =$ $\mathbb{E}_{s_{t+1}\sim\mathcal{P}(\cdot|\xi_t,a_t)}\left[D_{KL}[p(\theta|\xi_t,a_t,s_{t+1}) \parallel p(\theta|\xi_t)]\right]$ where $\mathcal{P}$ is the dynamics. History $\xi_t = \{s_1, a_1, \dots, s_t\}$ . Then, the information gain $I(s_{t+1};\Theta|\xi_t,a_t)$ is added to the reward function, resulting in an intrinsic reward $r'(s_t,a_t,s_{t+1})$ that encourages the exploration. $p(\theta|\xi_t,a_t,s_{t+1})$ is computed via BNN where distribution is captured by full factorized Gaussian and posterior is approximated by VI. The uncertainty of network output (Q value) can be integrated into *the objective/loss function* [190] to improve the convergence of DQN [191]. This balances the policy exploration and exploitation, improving the convergence of DQN. Another example [192] incorporates uncertainty to actor-critic algorithm to *detect outlier data* (state-action pairs that are out-of-distribution (OOD)) and *down-weight* their distribution in loss function. This is achieved by inserting uncertainty (variance) into the actor and critic loss functions where dropout VI performs before each weighted layer, and also performs dropout at test time. This decreases the probability of maximizing Q function with OOD samples, and down-weigh related Bellman loss for Q function (See Supplement C2).

Instead of applying uncertainty for policy exploration, BNN can be used to *approximate the environment dynamics* [193], therefore assisting the learning of RL policy. BNN with weight $\boldsymbol{W}$ (corrupted by additive noise $z_n \sim \mathcal{N}(0,\Sigma)$ and random disturbance $z_n \sim \mathcal{N}(0,\gamma)$ ) can be used to approximate the transition $s_t = f(s_{t-1},a_{t-1},z_t;\boldsymbol{W})$ , therefore dynamics $p(s_t|s_{t-1},a_{t-1})$ can be factorized and approximated via the following predictive distribution $p(s_t|s_{t-1},a_{t-1}) \approx \int \mathcal{N}(s_t|f(s_{t-1},a_{t-1},z_t;\boldsymbol{W}),\Sigma)\, q(\boldsymbol{W})\mathcal{N}(z_t|0,\gamma)d\boldsymbol{W}dz_t$ where the posterior $p(\boldsymbol{W},\boldsymbol{z}|\mathcal{D})$ can be approximated by variational distribution $q(\boldsymbol{W},\boldsymbol{z})$ via minimizing the $\alpha$ -convergence $D_\alpha[q(\boldsymbol{W}) \parallel p(\boldsymbol{W}|\mathcal{D})]$. $q(\boldsymbol{W})\mathcal{N}(z_t|0,\gamma)$ is acquired after the factorization of $q(\boldsymbol{W},\boldsymbol{z})$ via Bayes' rule. At the same time, action $a$ is generated via a deterministic policy $a \leftarrow \pi(a;\boldsymbol{W_\pi})$ that is computed via minimizing an objective function $J(\boldsymbol{W_\pi}) = \mathbb{E}[\sum_{t=1}^{T} c(s_t)]$ where $c(s_t)$ is a cost function. Therefore, dynamics $\boldsymbol{W}$ and deterministic policy $\boldsymbol{W_\pi}$ are learned simultaneously in model-based RL. Given the above four applications of BNN, we can see that the uncertainty of BNN can be integrated into the exploration policy, reward function, and loss/objective function for policy exploration. Simple dynamics of model-based RL can be replaced by BNN where VI is required for approximation. Future research may focus on replacing black-box networks with BNN to provide better estimations, e.g. the back-box networks in meta-RL.

### *D. Combining Bayesian active learning with RL*

RL can aid AL by formulating the (batch) query process of AL as a RL process or MARL [194] to fine-tune the acquisition function via imitation learning [101], reinforced AL and deep reinforcement AL [102][103]. Unlike the combinations of other Bayesian methods with RL, there are few cases where AL can directly aid the convergence of RL. The contribution of AL to RL is currently limited to improving data quality. This indicates a promising direction to combine AL and RL in the future. AL features compressing data and finding informative data that has potential to indirectly contribute to the convergence of RL, especially the data-hungry model-free RL. However, informative data in real-world is always in sparse form that blocks a further application of AL on RL where time-sequential episodes with consistent data distribution are required. Currently, the application of AL is constrained in robotic explorations to generate exploration plans based on uncertainty (Fisher matrix) or ergodicity for the active and visual simultaneously localization and mapping (SLAM) [195], model predictive control (MPC), infotaxis and ergodic control [196] to aid the search and rescue, localization and mapping. To make AL fit the RL setting, it is possible to find special episodes that contain informative data or representations with salient features as much and efficient as possible for RL training. This unveils a good direction for AL to aid BO [27][183], inverse RL [197], imitation learning [198] or learning from demonstrations [199], and preference-based RL [200] for better sample efficiency and time efficiency given sufficient expert demonstrations, or for reducing the number of ranking queries to the expert in preference-based RL. This task is challenging because it is hard to assume an *oracle (priori model)* to fine-tune the acquisition function of AL. This oracle much have the ability to self-evolve and keep optimal with the change of the environment. In the regard of humans, the brain neurons have the properties of storing sensory, short-term and long-term information [201]. This coded priori information can be retrieved selectively (e.g., with attention mechanism) in real-time and contribute to instant construction of generalizable and task-specific priori model, an ability that artificial neurons don't have. Existing approaches turn to RL and meta learning to fine-tune the acquisition function [101][102][103][104][105]. The latest large language models (LLM) may be helpful in constructing a better priori model to fine-tune an acquisition function, but their implementation is expensive and struggles to give consistent and task-specific solutions, while lifelong Bayesian learning may be a cheaper solution.

### *E. Combing Bayesian generative models with RL*

In the case of integrating VI to RL [19][20], when actions and reward (optimality) are included in dataset for computing the maximum likelihood, actions and reward play an important role to connect with RL policy, resulting in a *joint learning* of generative models and RL policy simultaneously. These works based on VI build on PGM to derive four types of models: 1) *Generative models* $p(x_t|z_t)$ and $p(m_t|z_t)$. $p(x_t|z_t)$ decode the latent state to observed state, while $p(m_t|z_t)$ decodes latent state to sematic mask which can be seen as simplified observed state to provide interpretability showing how the system understands the environment semantically. 2) *Latent dynamics model* $p(z_{t+1}|z_t,a_t)$ which predicts the next latent state based on current latent state. 3) *Inference model* $p(z_{t+1}|x_{t+1},a_t)$ which infers the latent state based on observed state. 4) *Policy model* $\pi(a_t|z_t)$ like SAC [202] which selects optimal actions based on latent state. This is achieved by maximizing ELBO, instead of maximizing intractable likelihood of dataset.

In VAE case, RL policy can be learned *independently* by training independent VAE and independent RL policy like PPO first, and then matching the decoder of VAE $p(x_t|z_t)$ and RL policy $\pi(x_t|s,z_t)$ to make them consistent for overall convergence [203] where $z$ encodes fixed-size sample sequence, instead of one sample. This results in better convergence. VAE embeds action sequence and prepares high-level skill priors [203][204][205][206], therefore learned skill priors are generalizable to similar RL tasks. For example, this may be achieved by maximizing ELBO to train variational posterior $q(z|\boldsymbol{a_i})$ that is used to guide the learning of skill priors $p_{\boldsymbol{a}}(z|s_t)$ [204] via minimizing $\mathbb{E}_{(s,\boldsymbol{a_i})\sim\mathcal{D}}\left[D_{KL}[q(z|\boldsymbol{a_i}) \parallel p_{\boldsymbol{a}}(z|s_t)]\right]$ where $\boldsymbol{a_i}$ is action sequence with $i$ actions, and $z$ latent state. Then, learned skill priors $p_{\boldsymbol{a}}(z|s_t)$ are used to guide the learning of high-level RL policy via maximizing the objective $J = \mathbb{E}_\pi\left[\sum_{t=1}^{T} \tilde{r}(s_t,z_t) - \alpha\, D_{KL}[\pi(z_t|s_t) \parallel p_{\boldsymbol{a}}(z_t|s_t)]\right]$ where $\tilde{r}$ is the discounted reward, $\alpha$ a discount factor. $\pi(z_t|s_t)$ is high-level hierarchical policy that generates embedded skill $z_t$ conditioned on current observation $s_t$. After that, embedded skill $z_t$ will be decoded by low-level RL policy $\pi(a|z_t)$ to generate practical actions. Given the above comparisons of VI and VAE to derive RL policy, we can conclude that VAE has two advantages over VI: 1) Data-efficiency. In VAE case, RL policy can be trained independently which contributes to overall convergence, while RL policy is trained jointly with generative models in VI case. 2) Generalization. VAE focuses on learning encoder, generative models (decoder), and RL policy, while VI focuses on learning generative models and RL policy. The learning of encoder shapes the flat RL problem to hierarchical RL problem where the encoder provides high-level generalizable hierarchical policy/prior. However, it is hard to assume or theoretically justify an accurate *combination* of variational distribution $q(\boldsymbol{z}|\boldsymbol{x})$ and prior distribution $p(\boldsymbol{z})$ for global optimum. Recall that $\log p(\boldsymbol{x}) \geq \mathcal{L} = -KL(q(\boldsymbol{z}|\boldsymbol{x}) \parallel p(\boldsymbol{z})) + \mathbb{E}[\log p(\boldsymbol{x}|\boldsymbol{z})]$. In most cases, the maximization of ELBO converges to a local optimum because of inaccurate assumptions in prior distributions that is a standard criticism of Bayesian inference. Moreover, variables covary, instead of being independent. This may result in unstable posterior distribution in optimization. The classical way to handle the criticism of prior selection is sensitivity analysis where the prior specification can be *noninformative (objective)* or *informative (subjective)* [207]. Noninformative prior means no background knowledge before observing current data where current data decides posterior, while informative prior means some background knowledge obtained before observations to specific a better prior where posterior converges faster. Current HMM like HGMM follows the way of informative prior specification and assumes variables are independent. This works well in finding a good local optimum, but the difference between this local optimum and unknow global optimum is hard to quantify. Anyway, the background knowledge cannot guarantee a true prior, like the specification of oracle to fine-tune acquisition function of AL we discussed in the previous section. Bayesian meta learning may be a promising direction for prior specification, but it remains to be a challenging job.

In diffusion model case, agent motion policy can be directly derived by conditioning diffusion model $p_\theta(a)$ on current state $s$ via $p_\theta(a|s)$ where $p_\theta(a)$ is directly modified from $p_\theta(x)$ [208]. Diffusion models may be seen as an extension of estimators [208] where the denoising network can represent inaccurate dynamics because the denoising process is guided unconditionally. To improve dynamics and suit diffusion model to RL problems, the constraints (rewards) must be added to guide the denoising process where the constraints can be described as the optimality $\mathcal{O}$ and $p(\mathcal{O}_t = 1) = \exp(r(s_t, a_t))$. In the settings of diffusion models and RL (diffusion planner) [209], the objective is to generate optimal trajectory $\boldsymbol{\tau}$ and given Bayesian rule $\tilde{p}_\theta(\boldsymbol{\tau}) = p(\boldsymbol{\tau}|\mathcal{O}_{1:T} = 1) \propto p(\boldsymbol{\tau})p(\mathcal{O}_{1:T} = 1|\boldsymbol{\tau})$ where $p(\boldsymbol{\tau})$ is computed by denoising network and classifier guidance $p(\mathcal{O}_{1:T} = 1|\boldsymbol{\tau})$ is computed by independent reward model. To conclude, diffusion planner has many advantages given diffusion and reward models: 1) long-horizon and varying-length planning due to diffusion models (note that just one action is executed) and 2) adaptivity/generalization to unseen scenarios due to a flexible/independent reward model. However, the challenges are obvious: 1) requirements in computation and prior knowledge for the reward model, 2) inherited limitations from diffusion models, and 3) two-dimensional optimization emerges where generated trajectories must be as short as possible and be optimal simultaneously. Therefore, diffusion planner can be seen as a "special" hierarchical RL where the shortest trajectory (outer loop) must be sampled from optimal candidates (inner loop). Diffusion planner can also be used in multi-agent hierarchical RL as high-level controller/planner to coordinate agents' objectives where agents' actual motions are controlled by low-level RL policy [210].

*F. Combining (Bayesian) meta-learning with RL*

Meta-RL trains global and local latent models for each task simultaneously. The benefit is a better generalization, because the trained global latent model (meta model) that contains shared behaviors of tasks can be reused in new tasks after adaptation. Meta-RL avoids learning from scratch, therefore better data efficiency is achieved and recent research finds an analogue of meta-RL in human brain [14] for learning similar tasks. Meta-RL is the extension of meta learning in RL where the labeled dataset $(x, y)$ in supervised learning changes into the transition tuples $< s, a, r, s' >$. General objective of meta-RL [30] is $J(\theta) = \mathbb{E}_{\mathcal{M}^i \sim p(\mathcal{M})}\left[\mathbb{E}_{\mathcal{D}}\left[\sum_{\tau \in \mathcal{D}_{K:H}} G(\tau)|f_\theta, \mathcal{M}^i\right]\right]$ where $G(\tau)$ is discounted return of MDP $\mathcal{M}^i$. $\theta$ is the global latent variable, while the local latent model/variable $\phi = f_\theta(\mathcal{D})$ is generated by learning/adapting dataset $\mathcal{D}$ via $f_\theta$. Episodes $\tau$ are sampled from MDP $\mathcal{M}^i$ to form the training dataset $\mathcal{D}_{K:H}$ where $H$ is the length of total trails, while $K$ is the length of trails for adaptations. When $K = 0$, all dataset $\mathcal{D}$ is for training. Value $K$ is decided by specific tasks. For example, $K = 0$ is common in most challenging autonomous driving tasks where the robot must interact with the environment to collect real-time time-series data for adaptations, resulting in suboptimal local latent variables. When training a human-like robotic arm for cooking with expert experience where $K > 0$, the adapted local latent variable can perform like human to better accomplish cooking tasks. Meta-RL generally consists of optimization-based approaches and black-box (recurrent-based) approaches that is practical for large-scale and high-dimensional tasks. In the black-box approaches, $f_\theta$ can be deep neural network or recurrent neural network (RNN) [211] where local variable $\phi$ is approximated via RNN latent state that varies on every timesteps. The benefit is that decisions made by local latent variable $\phi$ consider the historical experience that is also used for adaptation in every timesteps. The limitation is that adapted $\phi$ works poor in unseen scenarios. Given application scenarios, meta-RL is classified into *few-shot* and *many-shot meta-RL*. Few-shot meta-RL suits small-scale tasks with *similar/narrow* distributions, while many-shot meta-RL suits large-scale tasks with *diverse/broader* distributions. In real-world scenarios, we require meta-RL to fit tasks (e.g., autonomous driving) with diverse distributions and OOD tasks, with less data for adaptations. This challenges few-shot and many-shot meta-RL.

**Few-shot meta-RL.** Few-shot meta-RL includes black-box approaches and optimization-based approaches (also known as the *parameterized policy gradient approaches* (PPG)). Their difference is on how to generate local variables $\phi$. For both approaches, local latent variables are generated via $\phi = f_\theta(\mathcal{D})$. For black-box approaches, $f_\theta$ is either DNN or RNN and can be used as a universal function approximator where the convergence of $\phi$ is not guaranteed, and local $\phi$ and global (meta) $\theta$ have a "tight" relationship, as we discussed in BML. However, PPG attempts to make $\phi$ converge in the *inner loop* first and then make $\theta$ converge in the outer loop, especially in the amortized meta learning [116]. PPG can be extended to model-based MAML where dynamics is meta learned.

In adaptation of PPG, initial policy parameter $p(\phi_0)$ is captured by Gaussian and the posterior is approximated via VI. However, data for adaptation is limited in real-world tasks. It is possible to update/adapt a part of parameters with limited data via updating only the weight and bias of last layer of policy [212][213] or updating only a *context vector* on which the policy is conditioned [213]. When updating $\theta$, the challenge is to compute *second-order derivatives* (meta-gradients) that brings bias and errors, resulting in unstable and data-inefficient training. *Bias* is introduced because of the dependence of collected data for adaptation $\mathcal{D}_{0:K}$ [30]. After each training session, generated $\phi = f_\theta(\mathcal{D}_{0:K})$ is used for collecting new data for another round of training. The quality of $\mathcal{D}_{0:K}$ heavily impacts the data quality or data distribution in the following data collection process, therefore impacting the updates of following $\phi$ and $\theta$. *Variance* is introduced because of multi-step gradient updates of $\phi$. Bias is expected to be mitigated via the bias-variance trade-off [214][215], first-order approximation [216][117], gradient-free optimization like evolution strategies [217][30], and value-based algorithm in updating $\theta$ [218] to avoid the meta-gradients. In adaptation of black-box approaches, lesser data might be required because of "tight" relationship of local and global variable. Here, just a context vector is updated, instead of all parameters. The context vector can be generated via RNN or networks that condition on the history like the transformer [115] and memory-augmented network [121]. For high-dimensional tasks, more data is required for adaptation because of deeper networks. In the update of global latent variable $\theta$, both on-policy algorithms and off-policy algorithms are widely used, but off-policy algorithms [205] are more data efficient. Overall, the black-box approach suit tasks with narrow distributions with lesser data for adaptations, resulting in good specialization, but it works poorly in the generalization of OOD tasks and tasks with broader distributions, despite the optimizations from the convolution and attention mechanisms [115] that prepare good task representations in adaptation. The comparisons of PPG and black-box approach is visualized in Fig. 4. To conclude, PPG suffers from computing the second-order derivatives but has better generalization, while black-box approach is data-efficient but too specialized.

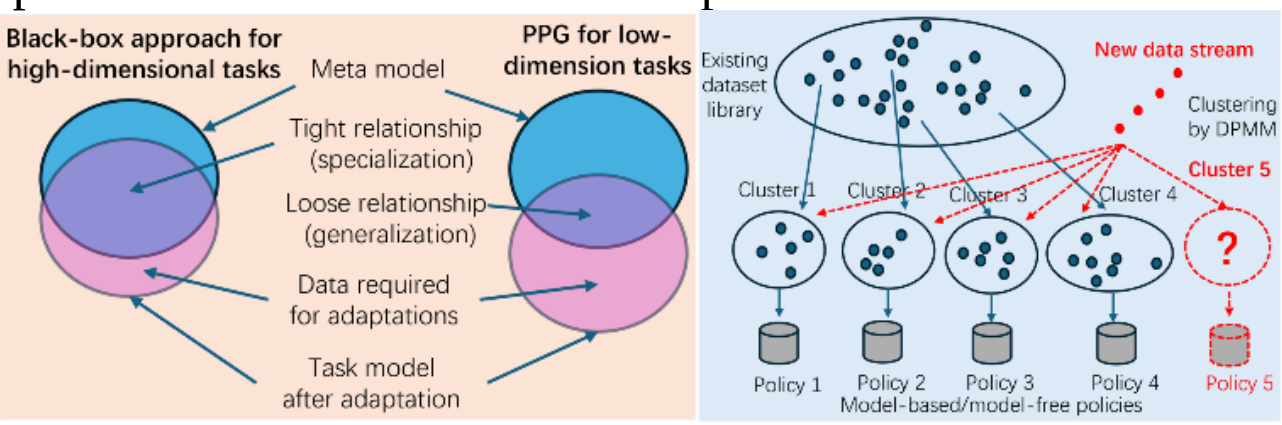

Fig. 4. Comparisons of PPG and black-box approach (left) and the mechanism of lifelong Bayesian learning on RL (right).

**Task inference and exploration methods with uncertainty quantification.** Meta-RL is required to fit tasks with diverse distributions where less samples are required for adaptations. However, task distributions are unknown in practical robotic applications. Task inference methods can quantify the uncertainty of unknown MDP. Therefore, a task distribution conditioned on limited/partial observed data $p(z|\mathcal{D})$ captured by Gaussian can be *mapped* into global $\theta$. Then, adapted local latent variables $\phi$ is computed after adaptations, resulting in local policies $\pi_\theta(a|s,\phi)$ or $\pi_{h_\theta(\phi)}(a|s)$ where $h_\theta$ maps the task into the weight and bias of policy [219]. This might be achieved by adding *additional loss* in the inference process via considering the historical privileged knowledge (e.g., knowledge encoded in RNN) [220], the knowledge like shared presentations and parameters from other agents in multi-tasking [220][221], or the value functions conditioned on latent space of current inferred task to provide better uncertainty [222] for task distribution. Exploration methods explore the policy space via the *exploration policy* to explore high-quality data for local policy, while task inference builds on *prediction models* (task distribution) to decide task and related meta policy. Task inference can contribute to exploration policy. For example, exploration policy can be guided by the objective/intrinsic reward from the task inference [223][224], while exploitation policy learns from scratch by optimizing meta-RL objective [221][225][226]. However, exploration policy is suboptimal when task inference methods work as a part of it, because of intrinsic reward from task inference changes the optimal policy. This problem can be mitigated via the annealed intrinsic reward [227]. Exploration policy can also contribute to task inference. Exploration policy provides high-quality data that grounds heavily impacts tasks inference performance. In special cases, 1) when full exploration is expensive, it is possible to firstly identify the information/condition for the exploitation, and then optimize exploration policy towards recovering this information [221]; 2) In the sparse reward case, it is hard to identify the reward before agents reach the goals. Experience relabeling that considers the rewards of other similar tasks is useful, but how to define the task similarity is under research [228]; 3) In adversarial cases, it is important to identify adversarial tasks and optimize worse-case return for exploration policy [229].

**Many-shot meta-RL.** Few-shot meta-RL takes a few episodes (e.g., ten) in the adaptation, while many-shot meta-RL takes many episodes (e.g., thousand). Complex tasks mean high-dimensional large dataset; therefore black-box approaches with deep neural network is more suitable than optimization-based approaches. Further, black-box approaches can be extended to many-shot meta-RL via introducing specialized RNN where multiple RNNs are rearranged and message passing is allowed among them [217] to make it generalizable to unseen scenarios. Many-shot meta-RL can be used to meta-learn the 1) *intrinsic rewards*, 2) *auxiliary tasks*, and 3) *additional objectives*. Learning from task-defining rewards of different task hierarchies always results in suboptimal exploration policy, because of potential conflicts of interest among the hierarchies, especially in sparse reward case [230]. An *intrinsic reward* [231] can be designed to guide the exploration process. Automatic design of intrinsic reward is possible via many-shot meta-RL. For example, it may be achieved by applying MAML to dueling DQN where meta training and meta testing of reward function are one-shot and many-shot, respectively [232]. Learned intrinsic reward can mix with task-defining reward for better convergence [233], or works alone for better exploration policy [234]. Auxiliary tasks mean subtasks with different intrinsic importance in a complex task. Auxiliary tasks are assumed to be known, and meta-RL is used to learn their importance. After importance weights of auxiliary tasks are meta-learned via unsupervised or self-supervised objectives optimized alongside RL [235], only useful subtasks are used in RL problem [236]. Similar to learning auxiliary tasks, when task hierarchy is known or becomes apparent in HRL, it is possible to derive the options/skills for subtasks and manager policies [237] for choosing these options where manager policies are meta-learned [238][239][240]. Therefore, HRL can generalize to similar tasks fast. Just maximizing task-defining objectives may be not good choices to accelerate the training, and it is possible to meta-learn additional objectives, e.g., the function of future trajectory [241], deep deterministic policy gradient (DDPG)-style objective where the critic is meta-learned [242], optimized task-defining objective via parametrizing coefficient of policy update size [243]. In short, optimizing other task-related objectives or parameterizing some key parts inside task-defining objectives via meta learning contributes to overall convergence.

To conclude, many-shot meta-RL suits tasks with diverse distributions. The key is how to learn a good representation for each hierarchy of task (intrinsic reward, auxiliary task, and additional objective). The challenge is the generalization to OOD tasks because the effectiveness of learned weights (inductive bias) on OOD tasks is unknown. In this challenging case, it is necessary to first identify the task distribution to decide if applying meta model for adaptations. Sometimes, learned meta weights are even harmful for adapting OOD tasks [30], as the negative knowledge transfer in most transfer learning and multi-task learning. It may be helpful to turn to a more complex learning hierarchy like lifelong Bayesian meta learning to handle OOD tasks where multiple meta models are maintained and selected for suitable adaptations.

*G. Combing lifelong Bayesian learning with RL*

Lifelong Bayesian learning casts the learning problem to the lifelong settings where old policy models are maintained, and new policy model is initialized incrementally as Fig. 4. Theoretically, an infinite number of policy models can be maintained. This is achieved by applying DPMM to cluster the dataset as the *top-layer policy* given the data features, while RL policies are initiated and trained for each cluster as *bottom-layer policies*. This method is also known as zero-shot meta-RL when the adaptation is not necessary because it relies on (historical/current) context data for task inference. For top-layer policy, DMPP has two popular variants: CRP and SBP. CRP generates the clusters by the property of MDP, while SBP generates the clusters by the distribution of data. Specifically, in CRP [125][126], the data feature in MDP (e.g., tuple $< s, a, r, s' >$) (cluster assignment variable) might be the *transition from* $s$ *to* $s'$ (e.g., varying size/location of obstacles and randomized physical parameters like mass, friction, damping of joints), *reward* (e.g., varying goal velocity/direction), or the *mixture* of both. These features/indicators decide the similarity of MDP, therefore avoiding negative knowledge transfer if RL policy is trained with data from unsimilar MDP. In SBP [55], the cluster assignment variables may be the *distribution* of latent variable (the embedding of a sequence of data) or the *mixture of distribution and representation* of latent variable. The bottom-layer policies can be model-free policy and model-based policy. CRP can cluster and label the data first with EM for posterior computing, and then feed labeled data to model-free policy for learning [125][126] where CRP and model-free policy can be learned independently. SBP must rely on the encoder of VAE to provide latent variables, and then cluster the latent variables. In this process, SBP, encoder, decoder, and RL policy (captured by Gaussian) can be learned simultaneously [55] where the posterior of SBP is approximated by VI.

Given the above two applications of combining lifelong Bayesian learning with RL, DPMM maintains an infinite number of model-based/model-free policies. Each policy is trained with similar MDP/distribution. DPMM model is expandable and learned policies are reusable for similar tasks, making DPMM and learned policies flexible and generalizable for unseen tasks (e.g., OOD tasks). Applications of seven potential Bayesian methods on RL are summarized in Supplement F4. Seven potential Bayesian methods have great potential to improve RL, but it doesn't mean other Bayesian methods cannot contribute to RL. Bayes' rule [244], Bayesian linear regression [245], Bayesian estimator [246], and probably approximately correct (PAC) Bayesian bound [247][248] can be applied to RL. See Supplement C3.

## V. BAYESIAN SAFE DECISION MAKING

In Bayesian context, safety can be improved indirectly by bounding posterior of Bayesian estimator [249] or applying BNN with MC dropout/BRNN [250][251][93] for better uncertainty quantification. In this section, safety is expected to be further improved via three perspectives: 1) *Safe sets*. 2) *Risk-averse/risk-aware objective*. 3) *Robustness*.

**Safety via safe sets.** Safe sets include the *safe weight sets* and *safe data sets*. Safe weight set focuses on how to derive safe weight sets in BNN and inverse RL (imitation learning) and how to learn a trustable weight (parameter) directly from trustable experience. Safe data set focuses on BO to build safe sets to guarantee safety and convergence simultaneously.

*Safe weight set.* [252] seeks training a safe weight set of BNN from which every action or trajectory is safe where BNN samples its weights from the safe weight sets. Safety certificates are searched in the form of a *safe positive invariant* which proves the safety of the safe weight sets. Finally, BNN policy is re-calibrated by rejecting the unsafe sampled weight. [253] acquires the safe reward function in imitation learning/inverse RL tasks via $R(\xi) = \sum_{s_i \in \xi} \widehat{w}^{\mathrm{T}} \phi(s_i)$ where $\widehat{w}$ is the estimated reward weight and $\phi$ the feature of state. So, the objective is to obtain better $P(\widehat{w}|\mathcal{D})$ where a safe weight set $\widehat{w}$ is required. Safe weight set for reward function is built via *quantifying the uncertainty by normalized Shannon entropy.* It is possible to build a *safe (trustable) parameter set* [254] captured by Beta distribution $t_i \sim Beta(\alpha_i, \beta_i)$ where a trust parameter set is learned from *trustable experience*. It is also promising to train a neural network based on the sum of weighted Gaussian radial basis functions [255] to derive the second-order derivative for safe RL policy with smooth actions.

*Safe data set.* Contextual BO, safe BO (SafeOpt), and SafeOpt with multiple constraint (SafeOpt-MC) are the extensions of BO. Contextual BO [256] grounds the safe BO and it relies on *additional/external variable* (context) $z \in \mathcal{Z}$ that is included in the kernel $k\big((a,i,z),(a',i',z')\big) = k_a\big((a,i),(a',i')\big) \cdot k_z(z,z')$. Safety constraint is a special case of context. SafeOpt [256] aims to find global maximum within the currently known safe set (via exploitation), and expand the safe set (via exploration). SafeOpt-MC [257][258] further extends SafeOpt by incorporating multiple constraints separated from the objective to compute the location of the next sample (See Supplement D1). Other than BO, Lyapunov function [259] and Lipschitz continuity property [260] can also contribute to the construction of safe data sets. To summarize, the safety certificate in BNN, better uncertainty quantification of GP, and second-order derivative based on GP may contribute to the safe weight set. BO with additional safe constraints is useful to build safe datasets.

**Safety via risk-averse/risk-aware objectives.** Here the objective refers to the reward/value function. [261] defines a risk-aware objective to generate a policy with lower risk level (negative reward). This is achieved by defining a safe metrics-based distribution $P(d|\tau)$ where $d$ is *relative position* of cars. Risk level $\tau$ is defined as $\tau \in \Omega \stackrel{\text{def}}{=} \{2,1,0\}$. Therefore, the posterior is computed via Bayesian rules by $P(\tau|d) = \frac{P(d|\tau)P(\tau)}{\sum_{\tau\in\Omega}(P(\tau)P(d|\tau))}$ and a risk-aware objective is defined as $\arg\min_{\pi} \mathbb{E}_{\pi}\{\sum_{i=0}^{\infty} \gamma^i \varepsilon_{t+i} | s_t\}$ where $\gamma$ is a discounted factor, and $\varepsilon$ is the risk coefficient defined by $\varepsilon = \mathbb{E}(\tau) = \sum_{\tau\in\Omega} \tau\, P(\tau|d)$. [262] seeks a high-reward and safe policy by solving a constrained reward objective modeled by constrained MDP (CMDP) where the cost of unsafe behaviors (negative reward) is inferred by GP and it is upper bounded. Unlike CMDP that upper bounds unsafe behaviors cost, [260] constrains the objective of value function to yield safe policy by attaching the penalty to the accumulative reward obtained from the episodes in which the tight state constraints are violated (e.g., the episodes with perturbance). [263] maximizes a *conditional value at risk* (CVaR) objective in the setting of BAMDP for tasks with adversarial perturbance. This contributes to the minimization of *epistemic uncertainty* due to the prior distribution over MDP and *aleatoric uncertainty* due to the inherent stochasticity of MDP. This is achieved by minimizing the expectation of perturbed distribution by the adversary in the inner loop and maximizing the accumulative reward in the outer loop where BO is used to handle the continuous action space for the adversary (see Supplement D2). To conclude, safety is achieved by minimizing negative rewards from high risk level, unsafe behaviors, episodes where tight state constraints are violated, and adversary perturbance.

**Robustness via method synthesis.** Safety is also interpreted as robustness/stability against the approximation error, time-varying disturbances, and mismatch between the simulation and reality. Safety-related methods investigated in this section include *Lyapunov function*, *Barrier function*, *$\mathcal{L}_1$ adaptive controller*, and *domain randomization method*. [259] uses GP to learn the dynamics of a nonlinear system of robots. Safety is guaranteed if the system state stays/explores within a region of attractions (ROA). ROA is built by a mix of control policies (constructed via GP) and Lyapunov function that guarantees a stable control policy within a high-confident region when collecting data. GP with Lyapunov function enlarges ROA and increases the range of safe exploitation. BNN can combine with Lyapunov and Barrier functions [264]. BNN with GP provides an avenue for maintaining appropriate degrees of caution in face of the unknown. Lyapunov function guarantees stable adaptation and convergence of tracking errors, while safety is guaranteed via Barrier function. [265] incorporates GP to $\mathcal{L}_1$ adaptive controller. This contributes to a sample efficient and error-bounded learning via GP. Stability, robustness, and tracking performance under time-varying disturbances are guaranteed via $\mathcal{L}_1$ adaptive controller simultaneously. [266] applies domain randomization via GP with BO to randomize the parameters of simulator, resulting in a new simulator where RL policy trained with this new simulator is robust to errors from sim-to-real transfer [267]. The relationship between simulator parameters $\phi$ and real-world observations $J^{real}$ is modeled as GP written as $J^{real} = GP(\phi)$. BO is used in this process to search for the next samples. Therefore, given real-world observations $J^{real}$ and simulator's posterior, it is possible to select the next suitable simulator's parameter, instead of using the old one. To summarize, GP takes the first step to quantify policy uncertainty. Uncertainty quantification makes us know how uncertain the policy is, but safety of policy is still unknown. Applying BNN/BRNN may be an indirect way to improve robustness/safety [93][251], but GP/BNN/BRNN can further synthesize with Lyapunov function, barrier function, and $\mathcal{L}_1$ adaptive controller to improve robustness. It is also important to fast scale the difference between simulator and real world by GP and BO to enable sim-to-real transfer in robotics. Overall, GP, BO, and BNN play important roles in the learning of safety-related RL policy. Note that recent diffusion models has great potential to derive safety-related policy by adding safety constraints to the independent reward model [109][209].

## VI. OVERALL COMPARISONS

This section first compares agent decision-making algorithms built on the combination of Bayesian methods and RL, regarding four indicators (data efficiency, interpretability, generalization, and safety). Then, a meta perspective is provided to unveil how Bayesian methods work in different processes of RL. We here make general analysis and comparisons given the utility/strength of Bayesian methods.

**Overall comparisons.** With descriptions and references in sections 2-5, we can compare the agent decision-making algorithms built on the combination of Bayesian methods with RL (Table I), given their task scope and four indicators. Here we select model-free RL as a general benchmark.

*Data efficiency.* Bayesian meta-learning and lifelong Bayesian learning will result in excellent data efficiency on RL because they make RL polices reusable in ongoing and future training. VI, BNN and Bayesian generative models has good data efficiency on RL due to VI that makes intractable posterior converge fast [19][20][43][109][180]. BO and Bayesian AL can provide high-quality informative data [64][65][97][98][99] that indirectly improve the convergence of RL potentially. *Interpretability.* Regarding interpretability, we must mention explainability. These two terms are confusing in most cases because they have the same meaning literally. There is no theoretical boundary in these two terms, but in machine learning the interpretability is widely accepted [268] to be that the inherent mechanism of methods is clear/interpretable, while the explainability denotes whether post-training results (trained models) are understandable to human. For simplicity, we use interpretability to describe an overall performance of a method regarding clear inherent mechanism and explainable post-training results to humans. Here we assume Bayesian generative models provide excellent interpretability to RL due to decoder/denoising network [203][204][205][206][209] to provide visualized planning with uncertainty quantification despite the negative impact of VI, while VI just makes acceptable contribution to RL regarding interpretability due to its unreliable optimization process. Other algorithms make good contributions to RL due to uncertainty quantification of policies/some sub-modules. *Generalization.* Bayesian meta-learning and lifelong Bayesian learning [55][125][126] have made an excellent contribution to RL generalization due to effective reuseable policies in unseen cases with a few shots or without adaptation [232][233][236][237][241]. Bayesian generative models make a good contribution to RL generalization because it provides the encoder/denoising network that can be a high-level policy prior/planner in meta-RL/hierarchical RL [203][204][205][206] and denoising network can fast adapt to unseen cases via modifying the conditions in real-time [210]. Other algorithms don't make obvious contributions to RL generalization. *Safety.* BO has made an excellent contribution to RL safety due to the safe set [252][257][258] built on it. BNN may make a good contribution to RL safety if it combines with other safety-related approaches [264]. Other algorithms are built on GP that provides robustness to noise/disturbance; therefore, they make an acceptable contribution to RL safety. Diffusion model has great potential to derive safety-related policy by conditioning denoising process on safety related constraint [209].

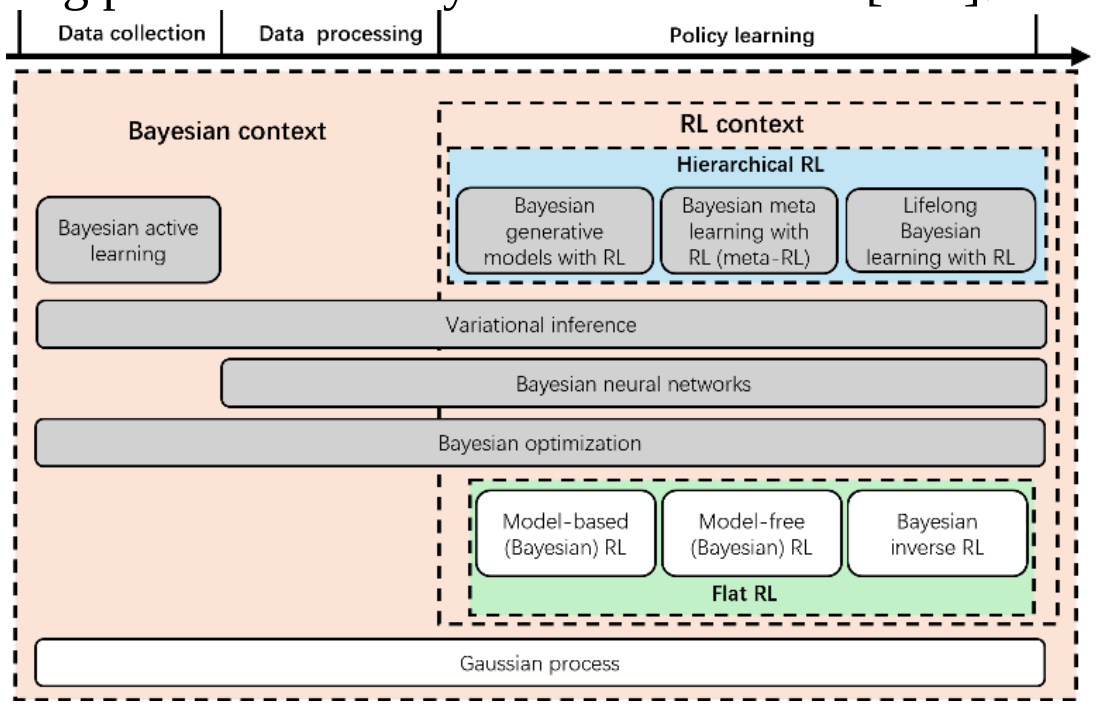

Fig. 5. Bayesian methods on different processes of RL.

Table I. Overall comparisons of agent decision-making algorithms built on the combinations of Bayesian methods and RL

| Algorithms | Task scope | Data efficiency | Interpretability | Generalization | Safety |
|---|---|---|---|---|---|
| Model-free RL (benchmark) | Medium/large-scale high-dimensional | Poor | Poor | --- | Poor |
| GPR-based Bayesian learning | Simple low-dimensional | Acc. | **Good** | --- | Acc. |
| Model-based RL | Small-scale low/high-dimensional | Acc. | **Good** | --- | Acc. |
| Model-based Bayesian RL | Small-scale low/high-dimensional | Acc. | **Good** | --- | Acc. |
| Model-free Bayesian RL | Medium-scale high-dimensional | Acc. | **Good** | --- | Acc. |
| Bayesian inverse RL | Simple low-dimensional | Acc. | **Good** | --- | Acc. |
| Variational inference with RL | Large-scale high-dimensional | **Good** | Acc. | --- | Acc. |
| Bayesian optimization with RL | Large-scale high-dimensional | **Good** | **Good** | --- | **Exc.** |
| Bayesian neural network with RL | Large-scale high-dimensional | **Good** | **Good** | --- | Acc./**Good** |
| Bayesian active learning with RL | Large-scale high-dimensional | **Good** | **Good** | --- | --- |
| Bayesian generative models with RL | Large-scale high-dimensional | **Good** | **Exc.** | **Good** | Acc./**Exc.** (Diffusion) |
| Bayesian meta-learning with RL | Large-scale high-dimensional | **Exc.** | **Good** | **Exc.** | Acc. |
| Lifelong Bayesian learning with RL | Large-scale high-dimensional | **Exc.** | **Good** | **Exc.** | Acc. |

* Level of performance: *Poor, Acc.-Acceptable, Good,* and *Exc.-Excellent.*

**Bayesian methods in different processes of RL.** We are not expected to summarize all statistical methods under Bayesian context but focus on most promising Bayesian methods and analyze their existing and potential contributions to RL for agent decision making. We can notice an important trend: Bayesian methods build on Bayes' rules and GP, and then these methods evolve towards handling more complex tasks by designing complex learning structure/hierarchy (e.g. Bayesian meta learning and lifelong Bayesian learning). Given overall comparisons and detailed demonstrations from Sections 2-5, it is easy to conclude how Bayesian methods work in different processes of RL for decision making potentially (Fig. 5). In *data collection* stage, Bayesian AL can select informative data for training. This improves the data quality, resulting in better data/time efficiency. In *data processing* stage, BNN can embed the data features with uncertainty estimation, resulting in better embedding for RL input. In *policy learning* stage, BNN is an excellent candidate for the function approximation of RL with uncertainty quantification. Bayesian generative models can provide high-level prior (encoder) or planner for hierarchical RL and assist deriving conditioning/hierarchical policy with better interpretability and generalization due to decoder or adaptive planner. Bayesian meta learning and lifelong Bayesian learning maintain a share/meta policy and multiple policies governed by high-level DPMM respectively. This results in better data efficiency and generalization. VI and BO may work at any stage of data collection, data processing, and policy learning where approximations and unknown variable parameterization are required. They contribute to better posterior and finding informative samples/reducing problem dimensions, therefore improving the data efficiency of RL.

## VII. DISCUSSIONS AND FUTURE RESEARCH

This section presents six extensions of standard RL problems: unknown reward, partial observable, multi-agent, multi-task, nonlinear non-Gaussian, and hierarchical RL problems. In real-world tasks, one or more complex cases may appear in one complex real-world task, but we just consider/discuss complex cases separately for simplification.

**Unknown reward problem.** In unknown reward case [165][22], model-based (Bayesian) RL prefers to extend hyper-parameter $\phi$ to $\theta = (\phi, \vartheta)$ where $\vartheta$ is the parameter of reward function. This results in expensive *two-stage* posterior computing/samplings (**Q1**) (See Supplement E1).

**POMDP problem.** In POMDP, the hyper-state space is $S' \in \mathcal{S} \times \Phi$, while hyper-state space is $S' \in \mathcal{S} \times \Phi \times \Psi$ in BA-POMDP [269] (See Supplement E2). Bellman optimality equation $V^*(b_t(s, \phi, \psi))$ requires accessing over all possible hyper-states for every belief. This results in *multi-stage* value function sampling to select the best value function. In this process, conjugate parameter $(\phi, \psi) \in \Phi \times \Psi$ is sampled and fixed therefore one value function can be solved via Bellman equation. Then conjugate parameter $(\phi, \psi)$ is resampled and fixed to solve another value function, and so on. There are $\Phi \times \Psi$ solutions theoretically, therefore multi-stage value function sampling to select the best model is intractable (**Q2**). Bayes risk based method bounds sampling number but provides a myopic solution, so this problem keeps challenging.

**Multi-agent problem.** In multi-agent case, it is hard to assume an accurate prior for dynamics. It is interesting to narrow the choice of prior automatically (**Q3**) using data driving approach, given limited samples at the beginning of tasks. Generally, we assume an inaccurate prior, based on our experience/intuition. In simple tasks, errors from inaccurate prior are not significant, but in complex tasks accumulative errors is not ignorable. Other than prior selection, multi-agent case still faces many challenges including *partial observability*, *non-stationarity*, *scalability (joint policy space)*, *reward assignment/hierarchy* [270][28], and *agent coordination* (See Supplement E3). General solutions of MARL include the centralized and decentralized MARL [271] regarding training schemes: 1) *Centralized MARL* that avoids the partial observability, non-stationarity, and coordination of agents, but suffers from poor scalability when the joint policy space increases exponentially (**Q4**). This problem is the same as Q1-2 where expensive multi-stage posterior sampling is required. Current planning methods may help alleviate this problem [272][273][274][275]. 2) *Decentralized MARL* where new problems arise due to partial observability, non-stationarity, and coordination of agents. It is easy to notice that centralized MARL and decentralized MARL complement each other. So, it is good to decouple the complex environment dynamics of centralized MARL [276] or reduce the dimensions [277][278] like the pairwise interactions of agents [279][280], therefore decoupled dynamics contribute to the update of complex environment dynamics. Similarly, in decentralized MARL, it is good to couple the dynamics of each agent to partly avoid the partial observability, non-stationarity, and agent coordination. Hence multi-stage posterior sampling becomes solvable. The problem is when the dynamics of agents should be coupled (**Q5**).

Generally, we may know suboptimal task-defining reward functions for agents or just know one reward function for the entire task. In (partly) decentralized MARL, there might be conflicts of interests between the agent goals and the task goal where proper reward functions are hard to design. A proper reward function also contributes to agent coordination. It is possible to incorporate expert knowledge (e.g., behavior rules of agents) to coordinate agents. Agent coordination can also consider the historical/shared information [281][282][283][284] or shared dynamics of other agents [285][286][287][288], therefore bounding the space of hyper-state in multi-stage posterior sampling. However, this problem is task-dependent and remains challenging. The challenge of reward assignment and agent coordination can be summarized to how to add deterministic factors to facilitate multi-stage posterior computing by better reward functions/agent coordination (**Q6**).

**Multi-task problem.** The challenges in multi-task case include *partial observability*, *scalability*, *catastrophic forgetting*,

*negative knowledge transfer*, and *distraction dilemma* (See Supplement E4). In multi-task case, it is impossible to acquire a flat RL policy applicable for tasks with broad distributions. Classical RL policy is too specialized, and if there is a large difference in task distributions, RL policy may easily suffer from catastrophic forgetting and negative knowledge transfer. So, the objective of multi-task is to learn a generalizable RL policy that can be used/reused for new task after fast adaptations. This might be achieved by adding hierarchy to RL policy. A typical example is the centroid policy [289][290][291] where a shared policy is extracted from existing policies and reused for new task. However, due to inaccurate task-specific reward design, shared policy is largely impacted by salient tasks, while other less important tasks contribute less to the shared policy, resulting in a conflict of generalization and specialization (**Q7**). This problem can be partly alleviated by constraining the contributions of different task policies to the shared policy [292], but it still keeps challenging.

**Nonlinear non-Gaussian problem.** Real-world problems are mostly nonlinear non-Gaussian that means three components: Nonlinear transformation model, non-Gaussian prior/posterior, and non-Gaussian noise distribution. It's better to hold two components fixed and then investigate the rest one. It is hard to investigate two/three components simultaneously with one component fixed or without fixing any component because it falls into the scope of multi-dimension optimization. Current research searches solutions from three aspects: 1) *Nonlinear transformation model, assuming the prior/posterior, and noise are captured by Gaussian.* In nonlinear transformation model and Gaussian prior case, the posterior is only Gaussian approximately, such that the use of inaccurate Gaussian posterior results in successive approximations with more bias. To handle nonlinear Gaussian case, the nonlinearity assumes to be fixed/stationary and predictable. Fixed nonlinearity is expected to be solved by linearization. Predictable nonlinearity means nonlinearity is possible to be inferred by task-dependent propagation methods like moment-based uncertainty propagation methods [293]. It is possible to derive task-dependent propagation models to approximate nonlinear transformation models but challenging. 2) *Non-Gaussian prior/posterior, assuming a known/linear transformation model and Gaussian noise.* In this case, the prior might be Gaussian [142] or partly Gaussian like truncated Gaussian mixture model (TGMM) [294] or completely non-Gaussian like random vector mixture models [295]. The posterior is still Gaussian in linear TGMM case. Generally, the posterior is not Gaussian because of the nonlinearity of transformation model or/and non-Gaussian prior. However, the distribution of posterior is still predictable by transforming the nonlinear stochastic process to deterministic process, as the moment-based uncertainty propagation methods. The challenge is how to derive task-dependent propagation models to approximate nonlinear transformation models (**Q8**). 3) *Non-Gaussian noise, assuming Gaussian prior/posterior and known nonlinearity/linearity.* Generally, we may assume that noise is simply captured by Gaussian mixture model and approximated by VI [296]. When noise distribution is hard to find, it is possible to improve the robustness against unknown stochastic noise or disturbance by like the tightly coupled distributed Kalman filter (DKF) with covariance intersection fusion that is used to correct cumulative errors from local KFs (LKFs) [297]. It is also possible to use data-driven approaches like stochastic distribution control (SDC) [298] for filter design where posterior/output distribution is modeled as an additional system variable. B-spline/randomized neural network might be used to represent adjustable output distribution, driven by the designed control input. Output distribution is numerically estimated by collected data, and then the distance between estimated output distribution and expected output distribution is minimized via like KL divergence. Therefore, estimated output distribution approximates expected output distribution, and control input is tuned accordingly. The problem is SDC generally works when the difference between output distribution and expected posterior distribution is narrow, therefore control input can compensate the impact of noise/disturbance. So, how to predict/prepare a suboptimal expected posterior distribution in SDC is important (**Q9**).

**Hierarchical RL problem.** HRL works in sparse reward case with hierarchical MDPs (e.g., Montezuma's Revenge) where entire MDP can be decomposed into many MDPs [299][300]. Classical HRL [301] is based on semi-Markov decision process (SMDP) where specific actions compose an action sequence, known as *temporally abstraction/skill*, therefore expensive exploration is reduced. Skills and related *skill policies* are managed by a high-level *control/manager* policy. It is hard to optimize multiple skill policies simultaneously because it falls into the scope of multi-layer/dimensional optimization. Existing HRL focuses on *bottom-up* and *top-down* training. Bottom-up training discovers, and fixes converged skill policies first and then manager policy is updated. Top-down training selects subgoals of skills first and then skill policies are developed. Regarding how to utilize reward functions for training, HRL can be trained in an *end-to-end manner* and *staged manner*. End-to-end manner may result in single-action skill policies or one skill policy to solve entire task, therefore regularizations [302] are required to control the switching of skill policies. Staged manner uses intrinsic rewards/objectives [303][304][305] in the search of skill policies by identifying terminal conditions [239] for avoiding the conflict of interests between skills and overall task. Frameworks of HRL are summarized to 1) *problem-specific*, 2) *option*, and 3) *goal-conditional* frameworks [306]. Problem-specific framework tightly relies on expert hand-craft knowledge. Therefore, skill policies are not generalizable and related manager policy is not applicable to other hierarchical tasks, despite their good interpretability and data-efficiency when learning from scratch. Goal-conditional framework is generic and efficient to find subgoals due to top-down training, but the instability of skill/manager policy, limited skill candidates, and the construction of subgoals keep challenging. Instability of skill/manager policy is expected to be improved by curriculum learning [307]. Limited candidates of skill are the consequence of top-down training. The construction of subgoals is task-dependent and heavily relies on human expertise. These subgoals are suboptimal but improve the overall sample efficiency. Option-based framework is hot due to its advantages of generalization across tasks, optimality of skill (option) policies, and stability of manager policy training. Option-based framework first samples option policy from manager policy. Then, each option policy is updated based on its option reward to provide option policy candidates. Finally, manager policy selects proper candidates from option policies given current conditions to maximize global reward. The option candidate is not always optimal if there is a conflict of interest between option reward and global reward. Therefore, option-based framework suffers from sample inefficiency when searching and managing option policies due to bottom-up training. It is crucial to enable manager policy to better initiate/sample, keep, and drop option candidates, so option policies and the manager policy are learned efficiently (**Q10**).

**Future research.** All the questions mentioned are listed in Supplement F5. We can notice a shared challenge among them:

They require *multi-stage/dimensional optimizations or require addressing complex explicit/implicit hierarchy*, e.g., two-layer problem hierarchy is obvious in Q1, Q2, Q4, Q6, Q7, and Q10. Problem hierarchy is not obvious in Q3, Q5, Q8, and Q9, but it's possible to derive two-layer hierarchy by parameterizing some unknown components as local model while overall objective as global model. Specifically, the prior choice, the possible combinations of agent dynamics, nonlinear transformation model, and expected posterior distribution in Q3, Q5, Q8, and Q9 can be parameterized, therefore reformulating one-dimensional optimization problem to two-dimensional optimization problem. For some two-dimensional optimizations where local and global models are policies, Bayesian meta-RL, lifelong Bayesian learning, and Bayesian generative models can be applied directly to them (e.g. Q10). For most two-dimensional optimization problems where local model is not the policy, we need to fix one parameter and then fast maximize another parameter (e.g. Q2 where local model is the dynamics) and BO is an efficient tool for finding suboptimal/optimal posterior. If the posterior is intractable, VI is useful for approximations. If we can find deterministic factors/connects between local and global modes (e.g., connections of observations/actions governed by RL policy and rewards governed by unknown reward function), two-layer optimization process can be shortened greatly, instead of trying all possible solutions. For general high-dimensional problems, we'd better consider BNN and Bayesian AL for better uncertainty quantification and data efficiency. To conclude, for future complex RL problems, it is necessary to first estimate task dimensions and structures with uncertainty quantification to design hierarchical solutions that suit data, instead of purely increasing the network capacity and size of datasets.

## CONCLUSIONS

This review carefully investigated basic Bayesian methods and potential Bayesian methods, and their combinations with RL for agent decision making. This resulted in model-based (Bayesian) RL, model-free Bayesian RL, Bayesian inverse RL, Bayesian generative models with RL, Bayesian meta-RL, and lifelong Bayesian learning with RL. BO and VI are useful in all decision-making stages of RL where BO can fast find the suboptimal/optimal parameterized components of RL and VI fast approximates the intractable posteriors of parameterized components or policy. BNN is a good choice for function approximation in RL due to its ability to provide uncertainty quantification of predictions, while Bayesian AL provides informative data for better data efficiency. Classical combinations of Bayesian methods with RL and related approximation methods were carefully summarized and compared. The latest novel combinations of Bayesian methods with RL and related applications were carefully presented and compared. Overall comparisons of agent decision-making approaches were presented given four indicators: generalization, interpretability, data efficiency, and safety. Finally, six challenging problem variants of RL were analyzed with ten questions proposed. We concluded a shared challenge of agent decision making: *Multi-layer/dimensional optimizations* where the combinations of Bayesian methods with RL are promising on that. Future research may focus on the solutions for the above two-layer/dimension optimization challenges for *practical applications* and more complex *problems with more than two hierarchies* given the knowledge from neural science. Currently, the research in consciousness/artificial consciousness is heating up, but it is hard to investigate directly. People suggest designing a high-level hierarchy/policy (world model) to govern different sub-policies and data from different brain areas and sensors. It might be an indirect solution for consciousness or provide useful hints to move forward at least. We believe the combination of Bayesian methods with RL would be an effective tool for designing the world model with excellent generalization, data efficiency, interpretability, and safety.

## ACKNOWLEDGEMENTS

This work was supported by the Research Council of Finland project 332177 Sustainable urban development emerging from the merge of cutting-edge Climate, Social, and Computer Science (CouSCOUs) and the Research Council of Finland Flagship program: Finnish Center for Artificial Intelligence (FCAI).

## BIOGRAPHIES

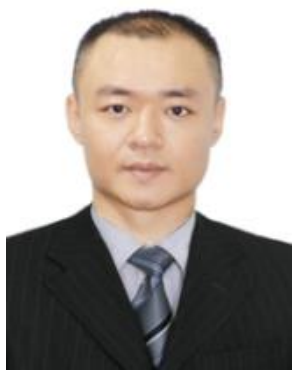

Chengmin Zhou obtained his PhD degree from the University of Eastern Finland in 2023 and works as postdoctoral researcher in Finnish Center for Artificial Intelligence (FCAI) and University of Helsinki from 2024. He currently works as the master thesis and PhD thesis co-supervisor, and published and reviewed many articles in many well-known journals including IEEE transactions on neural networks and learning systems, expert systems with applications, journal of intelligent manufacturing, etc. His research interests include Meta reinforcement learning (RL), Hierarchical RL, Multi-agent RL, Bayesian inference, and lifelong learning.

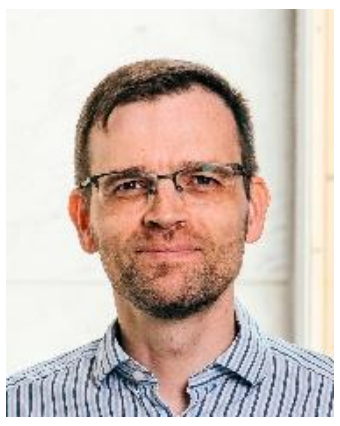

Ville Kyrki is a Full Professor at Aalto University School of Electrical Engineering. He joined Aalto University as an Associate Professor in 2012. He serves as the head of the Intelligent Robotics research group. His research interests lie mainly in intelligent robotic systems with a particular emphasis on developing methods and systems that cope with imperfect knowledge and uncertain senses. His published research covers feature extraction and tracking in computer vision, visual servoing, tactile sensing, robotic grasping and manipulation, sensor fusion, planning under uncertainty, and machine learning related to the previous. His research has been published in numerous forums in the area, including IEEE Transactions on Robotics, International Journal of Robotics Research, IEEE Transactions on Pattern Analysis and Machine Intelligence, IEEE Transactions on Haptics, and IEEE Transactions on Image Processing.

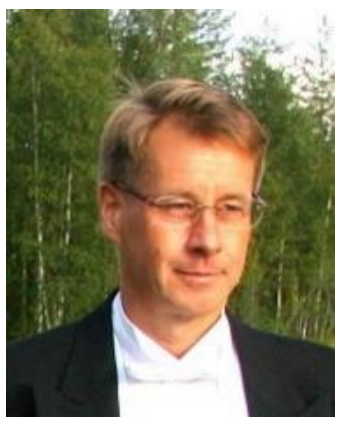

PASI FRÄNTI received his MSc and PhD in 1991 and 1994 from University of Turku, Finland. His research interests are in clustering algorithms, location-based services, analysis and optimization of health care systems among many others. He has been permanent full professor in University of Easter Finland since 2000. During his 30 years in research, he has published 128 peer reviewed journal and 185 conference publications. He has supervised 34 PhD students and is currently guiding ten more. Dr. Fränti has reviewed for 140 journals, and served as associate editor for Pattern Recognition Letters, Journal of Electronic Imaging, Machine Learning with Applications, Applied Sciences, and AI+. He is one of the founding editors of the AIMS Journal of Applied Computing & Intelligence.

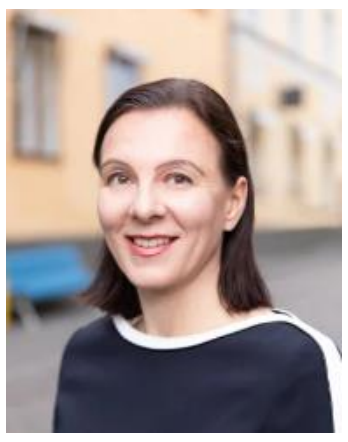

Laura Ruotsalainen is a Professor of Computer Science at the University of Helsinki and leads the Spatiotemporal Data Analysis (SDA) research group. The group conducts research on machine learning methods for forming and analyzing spatiotemporal data to advance sustainability science. She is a member of the steering group for the Finnish Center for Artificial Intelligence (FCAI) and the vice-chair of the ELLIS Institute Finland

# Supplementary materials

Chengmin Zhou, Ville Kyrki, Pasi Fränti, and Laura Ruotsalainen

This is the supplement for the paper: Combining Bayesian inference and reinforcement learning for agent decision making: A review

## Supplement A

### Supplement A1

| Name of the term | Mathematic form | Additional description |
|---|---|---|
| Probability distribution function | $\int_a^b p(x)dx = 1, x \in [a,b]$ | Where *non-negative* probability distribution function $p(x)$ satisfies $\int_a^b p(x)dx = 1$, and $x$ is a *random variable* over the interval $[a,b]$. |
| Two common probability moments (mean and variance) | $\mu = \mathbb{E}[x] = \int xp(x)dx$<br>$\mu = \mathbb{E}[F(x)] = \int F(x)p(x)dx$<br>$\Sigma = \mathbb{E}[(x-\mu)(x-\mu)^T]$<br>$\mu = \frac{1}{N}\sum_{i=1}^N x_i$<br>$\Sigma = \frac{1}{N-1}\sum_{i=1}^N (x_i-\mu)(x_i-\mu)^T$ | The moments here refer to the first and second moments, known as the *mean* and *covariance*. The mean is defined by the expectation of a random variable $x$ via $\mu = \mathbb{E}[x]$. In the matrix case, the mean is defined by the expectation of a general matrix function $F(x)$ via $\mu = \mathbb{E}[F(x)]$. The covariance $\Sigma$ is defined by $\Sigma = \mathbb{E}[(x-\mu)(x-\mu)^T]$. Normally, it is expensive to compute the mean and covariance via computing the integrals. Drawing samples from PDF to approximate the true mean and covariance is widely used. The sample mean is defined by $\mu = \frac{1}{N}\sum_{i=1}^N x_i$ where $N$ denotes the number of samples. The sample covariance is defined by $\Sigma = \frac{1}{N-1}\sum_{i=1}^N (x_i-\mu)(x_i-\mu)^T$ where the denominator is $N-1$, instead of $N$, to acquire *unbiased* estimate of the true covariance given the *Bessel's correction*. |
| Statistically independent, uncorrelated variables | $p(x,y) = p(x)p(y), p(x\|y) = p(x)$<br>$\mathbb{E}[xy^T] = \mathbb{E}[x]\mathbb{E}[y]^T$ | If two random variables $x$ and $y$ are *statistically independent*, the joint density function $p(x,y)$ factorized by $p(x,y) = p(x\|y)p(y) = p(y\|x)p(x) = p(y,x)$ is simplified as $p(x,y) = p(x)p(y), p(x\|y) = p(x)$. If two random variables $x$ and $y$ are *uncorrelated*, their expectation is factorized as $\mathbb{E}[xy^T] = \mathbb{E}[x]\mathbb{E}[y]^T$. |
| Normalized product | $p(x) = \eta p_1(x)p_2(x), \eta = (\int p_1(x)p_2(x)dx)^{-1}$<br>$p(x\|y_1,y_2) = \eta p(x\|y_1)p(x\|y_2), \eta = \frac{p(y_1)p(y_2)}{p(y_1,y_2)p(x)}$ | If $p_1(x)$ and $p_2(x)$ are two PDFs of $x$, the *normalized product (fused probability density function)* $p(x)$ is written as $p(x) = \eta p_1(x)p_2(x), \eta = (\int p_1(x)p_2(x)dx)^{-1}$ where $\eta$ is the *normalization constant* that makes $\int_a^b p(x)dx = 1, x \in [a,b]$. In the Bayesian context, the computing of the normalized product $p(x\|y_1,y_2)$ is written as $p(x\|y_1,y_2) = \eta p(x\|y_1)p(x\|y_2)$. |
| Uncertainty quantification via Shannon information, Mutual information, and Fisher information | $H(x) = -\mathbb{E}[\ln p(x)] = -\int p(x)\ln p(x)\,dx$<br>$I(x,y) = \mathbb{E}\left[\ln\frac{p(x,y)}{p(x)p(y)}\right] = \iint p(x,y)\ln\frac{p(x,y)}{p(x)p(y)}dxdy$<br>$I(x,y) = H(x) + H(y) - H(x,y)$<br>$I(x\|\theta) = \mathbb{E}\left[\left(\frac{\partial \ln p(x\|\theta)}{\partial\theta}\right)^T\left(\frac{\partial \ln p(x\|\theta)}{\partial\theta}\right)\right]$<br>$cov(\hat{\theta}\|x_{meas}) = \mathbb{E}[(\hat{\theta}-\theta)(\hat{\theta}-\theta)^T] \geq I^{-1}(x\|\theta)$ | The uncertainty can be measured by the moments (covariance), Shannon information and Mutual information. Shannon information $H(x)$ represents *how certain we are on our estimate of a PDF for a random variable.* Mutual information $I(x,y)$ represents *how much knowing one of the variables (x) reduces uncertainty about the other (y).* The *relationship* between the Shannon information and mutual information is written as $I(x,y) = H(x) + H(y) - H(x,y)$. Fisher information is defined by $I(x\|\theta)$ where the deterministic parameter $\theta$ decides the random variable $x$. Normally, we can just sample limited variables as the measurement $x_{meas}$ to acquire the *unbiased* estimate of the deterministic parameter $\hat{\theta}$. Fisher information here is to bound the covariance of the unbiased estimate via $cov(\hat{\theta}\|x_{meas})$. This is known as the Cramer-Rao lower bound which sets a limit to how certain we are in the estimation of the unbiased estimate $\hat{\theta}$ under the measurements $x_{meas}$. |

### Supplement A2

In continuous settings, to compute posterior $p(x|y)$, it is necessary to compute likelihood $p(y|x)$, prior $p(x)$, and probability of evidence $p(y)$. Normally, prior $p(x)$ is known (e.g., assuming $p(x)$ is captured by a certain distribution like Gaussian). Given joint density function $p(x,y)$, probability of evidence $p(y)$ is computed by *marginalization* $p(y) =$ $p(y)\int p(x|y)\,dx = \int p(x|y)\,p(y)dx = \int p(y|x)\,p(x)dx$ where $\int p(x|y)\,dx = 1$. Therefore, posterior is rewritten by

$$p(x|y) = p(x)\cdot\frac{p(y|x)}{\int p(y|x)\,p(x)dx}$$

where $p(x)$ is assumed to be known. Integral $\int p(y|x)\,p(x)dx$ is hard to compute in nonlinear case, but we do not need to know its exact value and just treat it as a constant for normalization.

In discrete settings, let $x_1, x_2 \dots, x_n$ be a set of events from sample space $\mathcal{S}$ where $p(x_i)$ is non-zero for any $i, i \in [1, n]$. Given the event $y$ that is independent of $x_i$, posterior is initially written by $p(x_i|y) = p(x_i, y)/p(y)$. Joint probability $p(x_i, y)$ can be factorized into $p(x_i, y) = p(y, x_i) = p(y|x_i)p(x_i)$. Probability of evidence $p(y)$ here denotes the probability of $y$ given the occurrence of *all* events $x_i$, and it is written as $p(y) = \sum_{k=1}^{n} p(y|x_k)p(x_k)$. Therefore, posterior here can be written by

$$p(x_i|y) = p(x_i)\cdot\frac{p(y|x_i)}{\sum_{k=1}^{n} p(y|x_k)p(x_k)}$$

where $\sum_{k=1}^{n} p(y|x_k)p(x_k)$ is unknown but a constant for normalization.

In multivariate Gaussian settings, joint Gaussian of the variable pair $(x, y)$ can be written as $p(x, y) = \mathcal{N}(\begin{bmatrix}\mu_x\\ \mu_y\end{bmatrix}, \begin{bmatrix}\Sigma_{xx} & \Sigma_{xy}\\ \Sigma_{yx} & \Sigma_{yy}\end{bmatrix})$ where kernel $\Sigma_{yx} = \Sigma_{xy}^T$, $p(x) = \mathcal{N}(\mu_x, \Sigma_{xx})$, and $p(y) = \mathcal{N}(\mu_y, \Sigma_{yy})$. Given $p(x, y) = p(x|y)p(y)$ and measurement $y$, posterior here can be written by

$$p(x|y) = \mathcal{N}\left(\mu_x + \Sigma_{xy}\Sigma_{yy}^{-1}(y - \mu_y), \Sigma_{xx} - \Sigma_{xy}\Sigma_{yy}^{-1}\Sigma_{yx}\right)$$

**Supplement A3.**

The variable set $\boldsymbol{X} \in \mathbb{R}^N$ is *Gaussian random variables* if $\boldsymbol{X} \sim \mathcal{N}(\mu, \Sigma)$ where $\mu$ and $\Sigma$ are the mean and covariance, respectively. GP is an *indexed set of Gaussian random variables* that can be represented by function $F(x), x \in \boldsymbol{\chi}$ where $\boldsymbol{\chi} = \{x_1, x_2, \dots, x_T\}$ is the index set. Hence, $\boldsymbol{X} = (F(x_1), F(x_2), \dots, F(x_T))$ that is a *vector-valued Gaussian random variables*. GP Function $F$ is unknown, but it can be fully specified by its mean $\bar{f}(x) = \mathbb{E}[F(x)]$ and its covariance/kernel $k(x, x') = \boldsymbol{Cov}[F(x), F(x')]$. Hence, GP can be simply written as $F(\cdot) \sim \mathcal{N}\left(\bar{f}(\cdot), k(\cdot,\cdot)\right)$ where covariance $k(\cdot,\cdot)$ is also called *kernel function* that encodes the correlations (prior information) between the members of GP.

GPR is a regression approach to infer GP Function $F$. GP Function $F$ and associated indexed random variable $F(x)$ are unknown, and they *cannot be inferred directly*. However, it is possible to collect associated indexed observation $Y(x)$ to predict the moments of $F(x)$. $F(x)$ and $Y(x)$ have some sorts of relationships that might be represented by following process with simple linear statistical model $\boldsymbol{H}$

$$Y(x) = \boldsymbol{H}F(x) + N(x)$$

where $\boldsymbol{H}$ is the *state transformation model* for linear transformation, $N$ the Gaussian noise. Hence, $F$ can be inferred *indirectly*, given $Y(x)$ and $\boldsymbol{H}$, to find *predicted mean and covariance* (posterior moments) of $F$.

**Supplement A4**

In kernel-based/non-parametric case, inference process with linear model $\boldsymbol{H}$ is described as

$$\boldsymbol{Y}_T = \boldsymbol{H}\boldsymbol{F}_T + \boldsymbol{N}_T$$

$\boldsymbol{Y}_T = (y_1, y_2 .. y_T)^T$, $\boldsymbol{F}_T = (F(x_1), F(x_2) .. F(x_T))^T$, and $\boldsymbol{N}_T \sim (0, \boldsymbol{\Sigma})$ where $\boldsymbol{\Sigma}$ is the covariance of measurement noise. Hence, prior is selected via $\boldsymbol{F}_T \sim \mathcal{N}(\bar{\boldsymbol{f}}, \boldsymbol{K})$ where mean $\bar{\boldsymbol{f}} = (\bar{f}(x_1), \bar{f}(x_2) .. \bar{f}(x_T))^T$ and kernel $\boldsymbol{K}_{i,j} = k(x_i, x_j)$. It is easy to see $\boldsymbol{Y}_T \sim \mathcal{N}(\boldsymbol{H}\bar{\boldsymbol{f}}, \boldsymbol{H}\boldsymbol{K}\boldsymbol{H}^T + \boldsymbol{\Sigma})$ where $\boldsymbol{F}_T$ and $\boldsymbol{N}_T$ are captured by Gaussians. Matrix form of $\boldsymbol{Y}_T = \boldsymbol{H}\boldsymbol{F}_T + \boldsymbol{N}_T$ with $F(x)$ can be written as

$$\begin{pmatrix}F(x)\\ \boldsymbol{F}_T\\ \boldsymbol{Y}_T\end{pmatrix} = \begin{bmatrix}1 & 0 & 0\\ 0 & \boldsymbol{I} & 0\\ 0 & \boldsymbol{H} & \boldsymbol{I}\end{bmatrix}\begin{pmatrix}F(x)\\ \boldsymbol{F}_T\\ \boldsymbol{N}_T\end{pmatrix}$$

where $\boldsymbol{I}$ is the identity matrix. *Joint distribution* is therefore written as

$$\begin{pmatrix}F(x)\\ \boldsymbol{F}_T\\ \boldsymbol{Y}_T\end{pmatrix} = \mathcal{N}\left\{\begin{pmatrix}\bar{f}(x)\\ \bar{\boldsymbol{f}}\\ \boldsymbol{H}\bar{\boldsymbol{f}}\end{pmatrix}, \begin{bmatrix}k(x,x) & \boldsymbol{k}(x)^T & \boldsymbol{k}(x)^T\boldsymbol{H}^T\\ \boldsymbol{k}(x) & \boldsymbol{K} & \boldsymbol{K}\boldsymbol{H}^T\\ \boldsymbol{H}\boldsymbol{k}(x) & \boldsymbol{H}\boldsymbol{K} & \boldsymbol{H}\boldsymbol{K}\boldsymbol{H}^T + \boldsymbol{\Sigma}\end{bmatrix}\right\}$$

where $\boldsymbol{k}(x) = \left(k(x_1, x), k(x_2, x) .. k(x_T, x)\right)^T$. According to the Gauss-Markov theorem, posterior distribution $p(F(x)|\boldsymbol{Y}_T)$ is still Gaussian. By factorizing joint distribution $p(F(x), \boldsymbol{Y}_T) = p(\boldsymbol{Y}_T)p(F(x)|\boldsymbol{Y}_T)$ where $p(\boldsymbol{Y}_T)$ is a constant and can be ignored, posterior moments of $F(x)$ conditioned on $D_T$ are

$$\mathbb{E}[F(x)|D_T] = \bar{f}(x) + \boldsymbol{k}(x)^T\boldsymbol{\alpha}$$

$$\boldsymbol{Cov}[F(x), F(x')|D_T] = k(x, x') - \boldsymbol{k}(x)^T\boldsymbol{C}\boldsymbol{k}(x')$$

where $\boldsymbol{\alpha} = \boldsymbol{H}^T(\boldsymbol{H}\boldsymbol{K}\boldsymbol{H}^T + \boldsymbol{\Sigma})^T(\boldsymbol{y_T} - \boldsymbol{H}\bar{\boldsymbol{f}})$ and $\boldsymbol{C} = \boldsymbol{H}^T(\boldsymbol{H}\boldsymbol{K}\boldsymbol{H}^T + \boldsymbol{\Sigma})^T\boldsymbol{H}$. They denote *sufficient statistics of posterior* and $\boldsymbol{y_T} = (y_1, y_2 .. y_T)^T$ is a realization of $\boldsymbol{Y}_T$.

**Supplement A5.**

Dirichlet equals *uniform distribution* once all parameters are equal. It is also a conjugate prior of *categorical distribution* and *multinomial distribution*. Dirichlet distribution describes the probability of $K$ outcomes, instead of two outcomes in Beta distribution. There are $K$ positive parameters $\boldsymbol{\alpha} = \alpha_1, \alpha_2, \dots, \alpha_K$, while there are two parameters ($\alpha$ and $\beta$) in Beta distribution. In contrast, PDF of Beta is defined by $p(\theta; \alpha, \beta) = \frac{\theta^{\alpha-1}(1-\theta)^{\beta-1}}{B(\alpha,\beta)}$ where *Beta function* $B(\alpha, \beta)$ is defined by $B(\alpha, \beta) = \frac{\Gamma(\alpha)\Gamma(\beta)}{\Gamma(\alpha+\beta)} = \int_0^1 t^{\alpha-1}(1-t)^{\beta-1}dt, \Gamma(z) = \int_0^\infty t^{z-1}e^{-t}dt$ . The mean and variance of Beta are $\frac{\alpha}{\alpha+\beta}$ and $\frac{\alpha\beta}{(\alpha+\beta)^2(\alpha+\beta+1)}$.

**Supplement A6.**

The probability of mixing proportions $\pi_i$ in CRP is captured by symmetric (flat) Dirichlet distribution

$$p(\pi_1, \dots, \pi_i | \alpha) \sim Dirichlet\left(\frac{\alpha}{K}, \dots, \frac{\alpha}{K}\right) = \frac{\Gamma(\alpha)}{\Gamma(\alpha/K)^K} \prod_{k=1}^{K} {\pi_k}^{\alpha/K-1}$$

where $\frac{\alpha}{K} = 1$. Assuming $c_i$ is a function of vector $\boldsymbol{\pi}$, above probability distribution can be rewritten as $p(c_1, \dots, c_k | \alpha) = \frac{\Gamma(\alpha)}{\Gamma(n+\alpha)} \prod_{k=1}^{K} \frac{\Gamma(n_k+\alpha/K)}{\Gamma(\alpha/K)}$ where $n$ is the number observations. $n_k$ is the number of observations in $k$-th cluster. The conditional prior of $c_i$ is written as $p(c_{i=k} | \boldsymbol{c}_{-i}, \alpha) = \frac{n_{-i,k}+\alpha/K}{n-1+\alpha}$ where $n_{-i,k}$ is the number of observations in $k$-th cluster, except for observation $y_i$. When $K \to \infty$, the possibility of a new customer indexed by $i$ falling into one ($k$-th) of *existing clusters or tables* is

$$p(c_i | \boldsymbol{c}_{-i}, \alpha) = \frac{n_{-i,k}}{n-1+\alpha}$$

and the possibility of a new customer indexed by $i$ falling into a *new cluster or unoccupied table* is

$$p(c_i \neq c_k \; \forall k \neq i | \boldsymbol{c}_{-i}, \alpha) = \frac{\alpha}{n-1+\alpha}$$

Hence, a higher $\alpha$ encourages building a new cluster. The possibility of a new customer falling into $k$-th existing cluster is proportional to $n_{-i,k}$. Individual data entry has the *exchangeability* that means the order of data entry does not matter in the clustering assignment.

SBP assumes that a unit-length stick can be broken into an infinite number of segments $\pi_k$

$$\pi_k = \beta_k \prod_{i=1}^{k-1} (1 - \beta_i)$$

where $\beta_k \sim \text{Beta}(1, \alpha)$ and $\alpha$ is a positive scalar. An intuitive illustration is shown as the bellow:

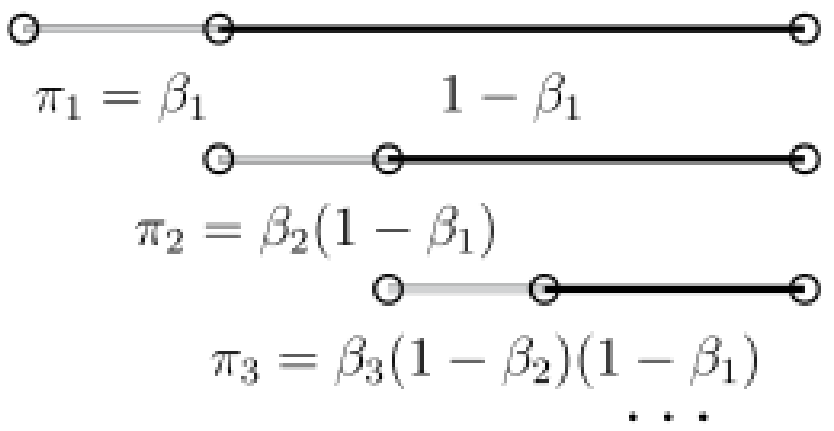

First, $\beta_1$ is sampled from $\beta_1 \sim \text{Beta}(1, \alpha)$ and segment length (mixing proportion) is $\pi_1 = \beta_1$. Then, $\beta_2$ is sampled from $\beta_1 \sim \text{Beta}(1, \alpha)$ but the segment length is $\pi_1 = \beta_2(1 - \beta_1)$. This process continues and we have $\sum_{k=1}^{\infty} \pi_k = 1$. Therefore, the segment length $\pi$ follows a Griffiths-Engen-McCloskey (GEM) distribution

$$\pi \sim GEM(\alpha)$$

For any random discrete distribution $G$ drawn from a DP, it can be written to be a weighted sum of point masses

$$G = \sum_{k=1}^{\infty} \pi_k \delta_{\theta_k^*}$$

where $\delta_{\theta_k^*} = 1$ if the point mass is at cluster parameter $\theta_k^*$, otherwise $\delta_{\theta_k^*} = 0$. The cluster $\theta_k^*$ might be sampled from a base/prior distribution $\theta_k^* \sim H$, therefore $G \sim DP(\alpha, H)$.

A set of observations $\boldsymbol{x} = \{x_1, x_2, x_3 \dots x_N\}$ has latent cluster parameter $\boldsymbol{\theta} = \{\theta_1, \theta_2, \theta_3 \dots \theta_N\}$ where $\theta_i$ is drawn from $G$ independently. In extreme case, each $x_i$ may have unique $\theta_i$ and also belong to unique cluster. In reality, each $\theta_i$ may have same/similar value with other $\theta_j$, therefore they can be clustered given a cluster assignment variable $v$ and here $\theta_v^*$ is used to represent true cluster that $x_i$ falls into. There are infinite number of true cluster $\theta_v^*$ theoretically, and the concentration/temperature $\alpha$ decide the probability of instantiating a new stick-breaking action.

**Supplement A7.**

How to derive ELBO. $KL(q(\boldsymbol{z})||p(\boldsymbol{z}|\boldsymbol{x}))$ is not computable, because given the conditional density $p(\boldsymbol{z}|\boldsymbol{x}) = \frac{p(\boldsymbol{z},\boldsymbol{x})}{p(\boldsymbol{x})}$, $KL(q(\boldsymbol{z})||p(\boldsymbol{z}|\boldsymbol{x})) = \mathbb{E}[\log q(\boldsymbol{z})] - \mathbb{E}[\log p(\boldsymbol{z}|\boldsymbol{x})] = \mathbb{E}[\log q(\boldsymbol{z})] - \mathbb{E}[\log p(\boldsymbol{z},\boldsymbol{x})] + \log p(\boldsymbol{x})$ where expectations are taken with respect to $q(\boldsymbol{z})$. The computing of $\log p(\boldsymbol{x})$ is intractable because the marginal density (evidence) $p(\boldsymbol{x}) = \int p(\boldsymbol{z},\boldsymbol{x})d\boldsymbol{z}$ cannot be computed in closed form. Instead of minimizing $KL(q(\boldsymbol{z})||p(\boldsymbol{z}|\boldsymbol{x}))$, it is plausible to maximize $-KL(q(\boldsymbol{z})||p(\boldsymbol{z}|\boldsymbol{x})) = \mathbb{E}[\log p(\boldsymbol{z},\boldsymbol{x})] - \mathbb{E}[\log q(\boldsymbol{z})] - \log p(\boldsymbol{x})$ that is still not computable because of $\log p(\boldsymbol{x})$. However, in the optimization, $\log p(\boldsymbol{x})$ *is equivalent to a constant* with respect to the $q(\boldsymbol{z})$. Hence, if it is possible find the maximum of $\mathbb{E}[\log p(\boldsymbol{z},\boldsymbol{x})] - \mathbb{E}[\log q(\boldsymbol{z})]$, we can claim we find the maximum of $\mathbb{E}[\log p(\boldsymbol{z},\boldsymbol{x})] - \mathbb{E}[\log q(\boldsymbol{z})] - \log p(\boldsymbol{x})$ where $\mathbb{E}[\log p(\boldsymbol{z},\boldsymbol{x})] - \mathbb{E}[\log q(\boldsymbol{z})]$ is known as ELBO defined by

$$ELBO(q) = \mathbb{E}[\log p(\boldsymbol{z},\boldsymbol{x})] - \mathbb{E}[\log q(\boldsymbol{z})] = \mathbb{E}[\log p(\boldsymbol{z})] + \mathbb{E}[\log p(\boldsymbol{x}|\boldsymbol{z})] - \mathbb{E}[\log q(\boldsymbol{z})] = \mathbb{E}[\log p(\boldsymbol{x}|\boldsymbol{z})] - KL(q(\boldsymbol{z}) \parallel p(\boldsymbol{z}))$$

The first part $\mathbb{E}[\log p(\boldsymbol{x}|\boldsymbol{z})]$ is an expected likelihood, while the second part $-KL(q(\boldsymbol{z})||p(\boldsymbol{z}))$ is the negative divergence of variational density $q(\boldsymbol{z})$ and prior density $p(\boldsymbol{z})$. Finally, the minimization of $KL(q(\boldsymbol{z})||p(\boldsymbol{z}|\boldsymbol{x}))$ turns to *find the balance* of $\mathbb{E}[\log p(\boldsymbol{x}|\boldsymbol{z})]$ and $KL(q(\boldsymbol{z})||p(\boldsymbol{z}))$ when maximizing $ELBO(q)$.

High computation of CAVI. The *full conditional* of $z_j$ is defined by $p(z_j|\boldsymbol{z_{-j}},\boldsymbol{x}), j = 1, \dots, n$ where $\boldsymbol{z_{-j}} = (z_1, \dots, z_{j-1}, z_{j+1}, \dots, z_n)$ is the supplement of $z_j$ given $\boldsymbol{z}$. Other variational factors are written as $q_\ell(z_\ell), \ell \neq j$. Given the mean-field property (independent latent variables) and fixed $q_\ell(z_\ell)$, the optimal $q_j(z_j)$ is proportional to the exponentiated expected log of full conditional

$$q_j(z_j) \propto \exp\{\mathbb{E}_{-j}[\log p(z_j|\boldsymbol{z_{-j}},\boldsymbol{x})]\} \propto \exp\{\mathbb{E}_{-j}[\log p(z_j,\boldsymbol{z_{-j}},\boldsymbol{x})]\}$$

During optimization, CAVI sequentially optimizes each $q_j(z_j)$ and holds others $q_\ell(z_\ell), \ell \neq j$ fixed. The computing of $\mathbb{E}_{-j}[\log p(z_j,\boldsymbol{z_{-j}},\boldsymbol{x})]$ requires traversing entire dataset in each iteration, resulting in expensive computation.

**Supplement A8.**

Improvement-based acquisition function. The acquisition function can be written as

$$a_{\mathrm{PI}}(x;\mathcal{D}_n) \coloneqq \mathbb{P}[v > \tau] = \phi(\frac{\mu_n(x) - \tau}{\sigma_n(x)})$$

where $\phi(\cdot)$ is a cumulative distribution function and $v$ latent variable. The improvement in PI is measured via the indicator (utility function) $\mathbb{I}[v > \tau]$, while the improvement can be also measured via another indicator $(v - \tau)\mathbb{I}[v > \tau]$, resulting in the *expected improvement* (EI) where related acquisition function is written as

$$a_{\mathrm{EI}}(x;\mathcal{D}_n) := \mathbb{E}\left[(v - \tau)\mathbb{I}[v > \tau]\right] = (\mu_n(x) - \tau)\phi\left(\frac{\mu_n(x) - \tau}{\sigma_n(x)}\right) + \sigma_n(x)\phi\left(\frac{\mu_n(x) - \tau}{\sigma_n(x)}\right)$$

Optimistic acquisition function. GP-UCB uses mean $\mu$ and variance $\sigma$ to compute the location of the next sample $a_n$ via

$$a_n = \arg\max_{a\in\mathcal{A}} \mu_{n-1}(a) + \beta_t^{\frac{1}{2}}\sigma_{n-1}(a)$$

where $\beta_t^{\frac{1}{2}} \geq 0$ is an iteration-dependent weight that depends upon the number of evaluations $t$ and reflects the confidence interval of GP. When $\beta_t^{\frac{1}{2}} = 0$, optimization becomes a pure exploitation scheme and converges to a local optimum prematurely. When $\beta_t^{\frac{1}{2}}$ is large, optimization is a pure exploration, resulting in global optimum but slow convergence. Therefore, the trade-off of $\beta_t^{\frac{1}{2}}$ matters. Repeatedly evaluating the system at locations $a$ improves the mean estimate of the underlying function and decreases the uncertainty at candidate locations for the maximum, such that the global maximum is provably found eventually.

Information-based acquisition function. Unlike Thompson sampling and spectral sampling where the next location point $x_{n+1}$ is sampled from the posterior of the objective function directly, ES selects the point that may cause the largest reduction of entropy in posterior. This is achieved by measuring the *expected information gain* $H(x^*|\mathcal{D}_n) - H(x^*|\mathcal{D}_n \cup \{(x,y)\})$ and selecting the point $x^*$ that offers the most information via

$$a_{\mathrm{ES}}(x;\mathcal{D}_n) := H(x^*|\mathcal{D}_n) - \mathbb{E}_{y|\mathcal{D}_n,x} H(x^*|\mathcal{D}_n \cup \{(x,y)\})$$

where $H(x^*|\mathcal{D}_n)$ is the differential entropy of posterior, and the expensive expectation is computed over the distribution of random variable $y$ (location value).

Portfolios of acquisition function. GP-Hedge portfolio is based on a fixed objective function as the performance evaluator that does not consider the information of exploration, resulting in slow convergence of posterior. ESP is based on the information gain from the exploration, and this adaptive acquisition function contributes to the convergence of posterior where the meta-criterion is written as

$$a_{\mathrm{ESP}}(\boldsymbol{x};\mathcal{D}_n) := -\mathbb{E}_{y|\mathcal{D}_n,x} H(\boldsymbol{x}^*|\mathcal{D}_n \cup \{(\boldsymbol{x},y)\})$$

where entry point is selected from the candidates by $x = \arg\max_{x_{1:K,n}} a_{\mathrm{ESP}}(\boldsymbol{x}; \mathcal{D}_n)$. $K$ here is the number of acquisition functions.

Additionally, BO can be extended to complex search spaces with categorical and conditional inputs. When objective function $f$ is nonconvex and multimodal, BO can make the best of historical full information to make the search process efficient.

**Supplement A9.**

VI assumes that a suitable posterior $q_\theta(\boldsymbol{w})$ is restricted to a certain family of distributions parameterized by $\theta$, and this posterior can be obtained by minimizing its dissimilarity to true posterior $\pi(\boldsymbol{w}|\mathcal{D})$ via KL-divergence $KL(q_\theta(\boldsymbol{w})||\,\pi(\boldsymbol{w}|\mathcal{D}))$. However, $\pi(\boldsymbol{w}|\mathcal{D})$ is still unknown, but KL-divergence can be rewritten as

$$KL\big(q_\theta(\boldsymbol{w})||\pi(\boldsymbol{w}|\mathcal{D})\big) = KL(q_\theta(\boldsymbol{w})||p(\boldsymbol{w})) - \mathbb{E}_q(\log p(\mathcal{D}|\boldsymbol{w})) + \log p(\mathcal{D}) = -ELBO + \log p(\mathcal{D})$$

where $\log p(\mathcal{D})$ can be ignored in the minimizing process because it does not depend on variational parameter $\theta$. Hence, the minimization process becomes the maximization of $-KL(q_\theta(\boldsymbol{w})||p(\boldsymbol{w})) + \mathbb{E}_q(\log p(\mathcal{D}|\boldsymbol{w}))$ where prior and likelihood are specified and they can be seen to be computable. Representatives of VI include: 1) Mean-Field Variational Bayes like CAVI that reduces the computation, but sacrifices the strong *correlation* between network weights, resulting in unreliable and inaccurate approximations. 2) SVI. In deep BNN, stochastic gradient descent with VI is commonly used to train BNN. Under this framework, MC integration/method that numerically computes a high-dimensional integral can be used in this process to approximate the gradient of posterior, resulting in large variance of gradient. This problem can be partly alleviated via the *score function estimato*r that relies on log-derivative property or the *path-wise derivative estimator* that builds on the reparameterization tricks to better approximate the posterior gradient.

An increase in the stochasticity of network weights contributes to expressivity, but a full stochasticity is unnecessary. The *reparameterization* trick is applied (e.g., in path-wise derivative estimator) to add stochasticity to the network weights. Another popular way to add stochasticity is the *regularization* techniques like Dropout.

Gaussian distributions can be applied to deep BNN, resulting in cascade deep Gaussian process where previous Gaussian serves as the input of the next Gaussian. The limitation of deep Gaussian process is an exponential computation with the increase of network layers. Problems in computations are expected to be alleviated by VI, but it yields unexplainable convergence of approximations.

Practical methods in the case of deep BNN with SVI include BBB and MC dropout where their priors are captured by a mixture of Gaussians. Their posteriors are fully factorized Gaussian and Bernoulli respectively. In MC dropout, noise can be efficiently transferred to network weights. This is achieved by introducing Bernoulli distributed independent random variable $\rho_\mu$ to the input space $x_i$ via

$$\phi_j = \theta\left(\sum_{i=1}^{N}(x_i\rho_\mu)\, w_{ij}\right), \rho_\mu \sim Bernoulli(p)$$

In BBB, the reparameterization trick is used for efficient sampling of expectations where expectations are approximated via MC estimates. The weight $w \sim \mathcal{N}(\mu, \sigma^2)$ is reparametrized by

$$w = g(\theta = \{\mu, \rho\}, \epsilon) = \mu + \sigma \odot \epsilon, \sigma = \mathrm{softplus}(\rho)$$

where $\sigma = \mathrm{softplus}(\rho)$ is to ensure the standard deviation parameter is positive. $\epsilon \sim \mathcal{N}(0,1)$ and $\odot$ is the Hadamard element-wise product.

**Supplement A10**

Lifelong learning with infinite mixture models and multi-task learning share the same objective as meta learning in model generalization. Meta learning is more general and lifelong learning (where dataset is clustered into subsets and parameter is captured by Dirichlet process mixture model) and multi-task learning (where multiple actors work in parallel to collect subsets and update shared parameter) can be seen as special cases of meta learning.

In amortized VI, the likelihood is expanded by

$$\begin{aligned}
\log\prod_{i=1}^{M} p(\mathcal{D}_i) &= \log\left[\int p(\theta)\left[\prod_{i=1}^{M}\int p(\mathcal{D}_i|\phi_i)p(\phi_i|\theta)d\phi_i\right]d\theta\right] \\
&\geq \mathbb{E}_{q(\theta;\psi)}\left[\log\left(\prod_{i=1}^{M}\int p(\mathcal{D}_i|\phi_i)p(\phi_i|\theta)d\phi_i\right)\right] - KL\big(q(\theta;\psi)\parallel p(\theta)\big) \\
&= \mathbb{E}_{q(\theta;\psi)}\left[\sum_{i=1}^{M}\log\left(\int p(\mathcal{D}_i|\phi_i)p(\phi_i|\theta)d\phi_i\right)\right] - KL\big(q(\theta;\psi)\parallel p(\theta)\big) \\
&\geq \mathbb{E}_{q(\theta;\psi)}\left[\sum_{i=1}^{M}\mathbb{E}_{q(\phi_i;\lambda_i)}[\log p(\mathcal{D}_i|\phi_i)] - KL\big(q(\phi_i;\lambda_i)\parallel p(\phi_i|\theta)\big)\right] - KL\big(q(\theta;\psi)\parallel p(\theta)\big) \\
&= \mathcal{L}(\psi, \lambda_1, \ldots, \lambda_M)
\end{aligned}$$

To solve above two-layer optimization problem, amortized VI first holds $\theta$ constant and compute $\lambda_i$ on the fly with amortized VI, and then $\theta$ is used to compute each $\lambda_i$ from $\mathcal{D}_i$. This is achieved by the inner loop and outer loop to compute local variables and global variable, respectively.

*Inner loop to compute local variables.* Each local latent variable $\lambda$ is computed via $K$-step stochastic gradient descent $SGD_K(\mathcal{D}, \lambda^{init}, \theta)$ where fixed $\theta$ is firstly used to initialize $\lambda^{init}$ and then each optimal local latent variable is computed via

$$\lambda^{(k+1)} = \lambda^{(k)} - \alpha \nabla_{\lambda^{(k)}} \mathcal{L}_{\mathcal{D}}\big(\lambda^{(k)}, \theta\big), k = 0, \dots, K-1$$

where $\alpha$ is a learning rate. $\mathcal{D}$ is each task dataset. $\mathcal{L}_{\mathcal{D}_i}(\lambda, \theta) = -\mathbb{E}_{q(\phi_i;\lambda)}[\log p(\mathcal{D}_i|\phi_i)] + KL\big(q(\phi_i; \lambda) \parallel p(\phi_i|\theta)\big)$ is a part of objective corresponding to $\mathcal{D}_i$ where likelihood $p(\mathcal{D}_i|\phi_i)$ and prior $p(\phi_i|\theta)$ are assumed to be Gaussians. Now, variational posterior approximation of local latent variable can be simply rewritten as $q(\phi_i|\mathcal{D}_i) = q\big(\phi_i; SGD_K(\mathcal{D}_i, \lambda^{init} = \theta, \theta)\big) = q(\phi_i; \lambda_i)$.

*Outer loop to compute global variable.* Maximizing $\mathcal{L}(\psi, \lambda_1, \dots, \lambda_M)$ equals to minimizing $-\mathcal{L}(\psi, \lambda_1, \dots, \lambda_M)$. After the inner loop, the objective turns to

$$\arg\min_{\psi} \mathbb{E}_{q(\theta;\psi)}\left[\sum_{i=1}^{M} -\mathbb{E}_{q(\phi_i|\mathcal{D}_i)}[\log p(\mathcal{D}_i|\phi_i)] + KL\big(q(\phi_i|\mathcal{D}_i) \parallel p(\phi_i|\theta)\big)\right] + KL\big(q(\theta;\psi) \parallel p(\theta)\big)$$

where $q(\phi_i|\mathcal{D}_i)$ is solved and fixed. Variational posterior approximation of $\psi$ can be easily computed via the gradient descent process. When adapting to new challenging task with small dataset $\mathcal{D}'$, *new* local latent parameter with good performance is computed via applying $K$-step gradient descent $SGD_K(\mathcal{D}', \lambda^{init}, \theta)$ again.

## Supplement B

### Supplement B1.

We assume that the robot follows the following *motion* and *observation models* respectively

$$\begin{cases} x_k = A_{k-1}x_{k-1} + v_k + w_k, k = 1..K \\ y_k = C_k x_k + n_k, k = 0..K \end{cases}$$

where $x_k$ denotes the system states. $A_{k-1}$ denotes the state transition matrix. $v_k$ denotes the input. $w_k$ denotes the process noise. $y_k$ denotes the observation or measurement. $C_k$ denotes the observation matrix. $n_k$ denotes the measurement noise. We also assume that the initial state $x_0 \sim \mathcal{N}(\check{x}_0, \check{P}_0)$ where $\check{x}_0$ and $\check{P}_0$ denotes the prior's moment estimations. The process noise $w_k \sim \mathcal{N}(0, Q_k)$. The measurement noise $n_k \sim \mathcal{N}(0, R_k)$. There are two situations to access dataset. First, we can access a batch of data with or without the time-sequential dependency (*batch discrete-time case*). Second, we can recursively access time-sequential data (*recursive discrete-time case*).

GPR for batch discrete-time case. From the perspective of Bayesian Gaussian, the prior can be written as $p(x|v) = \mathcal{N}\big(\check{x}, \check{P}\big) = \mathcal{N}(Av, AQA^T)$. Estimated prior mean is $\check{x} = \mathbb{E}[x] = \mathbb{E}[A(v+w)] = Av$ where $A$ is lower-triangular, and $x = A(v+w)$ is the *lifted matrix form* of motion model that considers the batch data in the entire trajectory. The lifted covariance is $\check{P} = \mathbb{E}[(x - \mathbb{E}[x])(x - \mathbb{E}[x])^{\mathsf{T}}] = AQA^T$where $Q = \mathbb{E}[ww^T]$. The joint density of lifted state and observation is written as $p(x, y|v) = \mathcal{N}\left(\begin{bmatrix} \check{x} \\ C\check{x} \end{bmatrix}, \begin{bmatrix} \check{P} & \check{P}C^T \\ C\check{P} & C\check{P}C^T + R \end{bmatrix}\right)$ where $R = \mathbb{E}[nn^T]$. Joint density is factored by $p(x, y|v) = p(x|y, v)p(y|v)$ where $p(y|v)$ is the known Gaussian distribution. Hence, full Bayesian posterior $p(x|y, v)$ is computed via the factoring of joint Gaussian distribution

$$p(x|y, v) = \mathcal{N}\left(\big(\check{P}^{-1} + C^T R^{-1} C\big)^{-1}\big(\check{P}^{-1}\check{x} + C^T R^{-1} y\big), \big(\check{P}^{-1} + C^T R^{-1} C\big)^{-1}\right)$$

The above process repeats until the convergence of the posterior moments.

Kalman filter via GPR for recursive discrete-time case. Gaussian prior estimation at $k-1$ is written as $p(x_{k-1}|\check{x}_0, v_{1:k-1}, y_{0:k-1}) = \mathcal{N}\big(\hat{x}_{k-1}, \hat{P}_{k-1}\big)$ where $\check{x}$ denotes prior estimation and $\hat{x}$ denotes the posterior estimation. According to the linearity of the motion model, prior estimation at $k$ (*prediction step*) is written as $p(x_k|\check{x}_0, v_{1:k}, y_{0:k-1}) = \mathcal{N}\big(\check{x}_k, \check{P}_k\big)$where the prior moments are acquired by $\check{x}_k = A_{k-1}\hat{x}_{k-1} + v_k$ and $\check{P}_k = A_{k-1}\hat{P}_{k-1}{A_{k-1}}^T + Q_k$. The joint density of state and observation at time $k$ is written as $p(x_k, y_k|\check{x}_0, v_{1:k}, y_{0:k-1}) = \mathcal{N}\left(\begin{bmatrix} \mu_x \\ \mu_y \end{bmatrix}, \begin{bmatrix} \Sigma_{xx} & \Sigma_{xy} \\ \Sigma_{yx} & \Sigma_{yy} \end{bmatrix}\right) = \mathcal{N}\left(\begin{bmatrix} \check{x}_k \\ C_k\check{x}_k \end{bmatrix}, \begin{bmatrix} \check{P}_k & \check{P}_k{C_k}^T \\ C_k\check{P}_k & C_k\check{P}_k{C_k}^T + R_k \end{bmatrix}\right)$. By factorizing the joint density, the posterior $p(x_k|\check{x}_0, v_{1:k}, y_{0:k})$ at time $k$ (*correction step*) is written as

$$p(x_k|\check{x}_0, v_{1:k}, y_{0:k}) = \mathcal{N}\big(\mu_x + \Sigma_{xy}{\Sigma_{yy}}^{-1}(y_k - \mu_y), \Sigma_{xx} - \Sigma_{xy}{\Sigma_{yy}}^{-1}\Sigma_{yx}\big)$$

where mean is $\hat{x}_k = \mu_x + \Sigma_{xy}{\Sigma_{yy}}^{-1}(y_k - \mu_y)$, and covariance is $\hat{P}_k = \Sigma_{xx} - \Sigma_{xy}{\Sigma_{yy}}^{-1}\Sigma_{yx}$. After the prediction-correction steps, we acquire the predictor, Kalman gain, and corrector written as

$$\text{Predictor:} \begin{cases} \check{x}_k = A_{k-1}\hat{x}_{k-1} + v_k \\ \check{P}_k = A_{k-1}\hat{P}_{k-1}{A_{k-1}}^T + Q_k \end{cases}$$

$$\text{Kalman gain: } K_k = \check{P}_k{C_k}^T(C_k\check{P}_k{C_k}^T + R_k)^{-1}$$

$$\text{Corrector:} \begin{cases} \hat{x}_k = \check{x}_k + K_k(y_k - C_k\check{x}_k) \\ \hat{P}_k = (1 - K_kC_k)\check{P}_k \end{cases}$$

where $y_k - C_k\check{x}_k$ is the *innovation* (difference between actual and expected observations). *Kalman gain* $K_k$ weights innovation's contribution to the moment estimations.

**Supplement B2.**

**Bayes filter for recursive discrete-time case.** In the case where the motion and observation models are nonlinear, models are rewritten as

$$\begin{cases} x_k = f(x_{k-1}, v_k, w_k), k = 1..K \\ y_k = g(x_k, n_k), k = 0..K \end{cases}$$

Posterior belief is factored by Bayes rule and Markov property via

$$p(x_k|\check{x}_0, v_{1:k}, y_{0:k}) = \eta p(y_k|x_k)p(x_k|\check{x}_0, v_{1:k}, y_{0:k-1}) = \eta p(y_k|x_k)\int p(x_k|x_{k-1}, v_k)p(x_{k-1}|\check{x}_0, v_{1:k-1}, y_{0:k-1})\,dx_{k-1}$$

because $p(x_k|\check{x}_0, v_{1:k}, y_{0:k-1}) = \int p(x_k, x_{k-1}|\check{x}_0, v_{1:k}, y_{0:k-1})\,dx_{k-1} = \int p(x_k|x_{k-1}, \check{x}_0, v_{1:k}, y_{0:k-1})p(x_{k-1}|\check{x}_0, v_{1:k}, y_{0:k-1})\,dx_{k-1}$ where $p(x_k|x_{k-1}, \check{x}_0, v_{1:k}, y_{0:k-1}) = p(x_k|x_{k-1}, v_k)$ and $p(x_{k-1}|\check{x}_0, v_{1:k}, y_{0:k-1}) = p(x_{k-1}|\check{x}_0, v_{1:k-1}, y_{0:k-1})$. The $\eta$ preserves the axiom of total probability. $p(y_k|x_k)$ is the observation correction based on observation model $g(\cdot)$. $p(x_k|x_{k-1}, v_k)$ is the motion prediction based on motion model $f(\cdot)$. $p(x_{k-1}|\check{x}_0, v_{1:k-1}, y_{0:k-1})$ is the prior belief.

In the prediction step, prior belief $p(x_{k-1}|\check{x}_0, v_{1:k-1}, y_{0:k-1})$ propagates forward to acquire the *predicted belief* $p(x_k|\check{x}_0, v_{1:k}, y_{0:k-1})$, given the $v_k$ and motion model $f(\cdot)$. In the correction step, the predicted belief is updated, given the observation $y_k$ and the observation model $g(\cdot)$. This prediction-correction steps results in posterior belief $p(x_k|\check{x}_0, v_{1:k}, y_{0:k})$.

**MAP, maximum likelihood and SWFs.** The basic idea of MAP in linear-Gaussian case is to maximize the posterior distribution $p(x|v,y)$ to find the best state estimate (posterior mean) $\hat{x}$ by applying the Bayesian rules $\hat{x} = \arg\max_x p(x|v,y) = \arg\max_x \frac{p(y|x,v)p(x|v)}{p(y|v)} = \arg\max_x p(y|x)p(x|v)$ where $p(y|x) = \prod_{k=0}^{K} p(y_k|x_k)$ and $p(x|v) = p(x_0|\check{x}_0)\prod_{k=1}^{K} p(x_k|x_{k-1}, v_k)$. The maximum likelihood only use the observations to find the maximum posterior without the prior. We assume the observation model is $y_k = g_k(x) + n_k$ where the observation noise $n_k \sim \mathcal{N}(0, R_k)$ and $k$ here is an arbitrary data index, instead of the time index. The objective function is defined as $J(x) = \frac{1}{2}\sum_k \left(y_k - g_k(x)\right)^T {R_k}^{-1}\left(y_k - g_k(x)\right) = -\log p(y|x) + C$ where $C$ is a constant. The posterior mean is acquired by minimizing the objective function (maximizing the likelihood) via $\hat{x} = \arg\min_x J(x) = \arg\max_x \log p(y|x)$. MAP and maximum likelihood can be further solved by Gauss-Newton. Gauss-Newton iterates in the entire trajectory, while IEKF iterates in one timestep. The SWFs find the trade-off between Gauss-Newton and IEFK in terms of the iteration by iterating over many time-steps within a sliding window. In nonlinear cases with high-dimensional large dataset where the solving of MAP or maximum likelihood is expensive, MCMC is preferable choice to combine to achieve fast convergence with reasonable computing resources.

**Supplement B3.**

**Linearization and EKF.** During the linearization of EKF, posterior belief, and noise are constrained to be Gaussian via $p(x_k|\check{x}_0, v_{1:k}, y_{0:k}) = \mathcal{N}(\hat{x}_k, \hat{P}_k)$, $w_k \sim \mathcal{N}(0, Q_k)$, and $n_k \sim \mathcal{N}(0, R_k)$ where $\hat{x}_k$ and $\hat{P}_k$ denote posterior mean and covariance. $w_k$ is the *process noise* and the $n_k$ is the *measurement noise*. To linearize the current estimated state mean, nonlinear models are rewritten as

$$\begin{cases} x_k = f(x_{k-1}, v_k, w_k) \approx \check{x}_k + F_{k-1}(x_{k-1} - \hat{x}_{k-1}) + w'_k \\ y_k = g(x_k, n_k) \approx \check{y}_k + G_k(x_k - \check{x}_k) + n'_k \end{cases}$$

where current estimated state mean $\check{x}_k = f(\hat{x}_{k-1}, v_k, 0)$, $F_{k-1} = \frac{\partial f(x_{k-1}, v_k, w_k)}{\partial x_{k-1}}|_{\hat{x}_{k-1}, v_k, 0}$, and $w'_k = w_k \frac{\partial f(x_{k-1}, v_k, w_k)}{\partial w_k}|_{\hat{x}_{k-1}, v_k, 0}$. Current estimated observation $\check{y}_k = g(\check{x}_k, 0)$, $G_k = \frac{\partial g(x_k, n_k)}{\partial x_k}|_{\check{x}_k, 0}$, and $n'_k = n_k \frac{\partial g(x_k, n_k)}{\partial n_k}|_{\check{x}_k, 0}$. Given the above linearization, the statistic properties of current state $x_k$ and observation $y_k$ are obtained. Given the statistic properties and Bayes filter, the integral $\int p(x_k|x_{k-1}, v_k)p(x_{k-1}|\check{x}_0, v_{1:k-1}, y_{0:k-1})dx_{k-1} = \mathcal{N}(\check{x}_k, F_{k-1}\hat{P}_{k-1}{F_{k-1}}^T + Q'_k)$ because the integral is still Gaussian after passing a Gaussian through a (stochastic) nonlinearity. The posterior $p(x_k|\check{x}_0, v_{1:k}, y_{0:k})$ is also Gaussian and

$$p(x_k|\check{x}_0, v_{1:k}, y_{0:k}) = \mathcal{N}(\check{x}_k + K_k(y_k - \check{y}_k), (1 - K_k G_k)(F_{k-1}\hat{P}_{k-1}{F_{k-1}}^T + Q'_k))$$

according to the normalized product of Gaussian PDF. Finally, the predictor, Kalman gain, and corrector of EKF are obtained as that of KF. The performance of EKF can be further improved by IEKF. This is achieved by replacing the estimated mean $\check{x}_k$ with the *arbitrary* operating point $x_{op,k}$. However, the estimated mean $\check{x}_k$ or *arbitrary* operating point $x_{op,k}$ is just assumed to be the real prior mean, but they are not, in reality. Moreover, new PDF may not be Gaussian after passing the PDF through the non-linearity, but we assume it is. These factors bring inaccuracy to the posterior approximation.

**Supplement B4.**

**MC sampling on sigmapoint KF and particle filter.** The *prediction steps* of SPKF are as follows: 1) Convert the Gaussian representation $\{\mu_z, \Sigma_{zz}\}$ stacked with the prior and motion noise $\mu_z = \begin{bmatrix} \hat{x}_{k-1} \\ 0 \end{bmatrix}, \Sigma_{zz} = \begin{bmatrix} \hat{P}_{k-1} & 0 \\ 0 & Q_k \end{bmatrix}$ to a Sigmapoint representation $z_i$ via $z_0 = \mu_z, z_i = \mu_z + \sqrt{L+\kappa}\mathrm{col_i}\mathbf{L}, z_{i+L} = \mu_z - \sqrt{L+\kappa}\mathrm{col_i}\mathbf{L}, i = 1 \dots L$ where $\mathbf{L}$ is the lower-triangular and $\mathbf{L}\mathbf{L}^T = \Sigma_{zz}$. The $\kappa$ is a user-definable parameter which scales how far away the Sigmapoints are from the mean and affects the accuracy of conversion. The dimension $L = \dim \mu_z$. 2) Unstack the Sigmapoints representation and pass them through the nonlinear motion model $\check{x}_{k,i} = f(\hat{x}_{k-1,i}, v_k, w_{k,i}), z_i = \begin{bmatrix} \hat{x}_{k-1,i} \\ w_{k,i} \end{bmatrix}$; 3) Recombine the transformed Sigmapoints to acquire the *predicted posterior* via $\check{x}_k = \sum_{i=0}^{2L} \alpha_i \check{x}_{k,i}, \check{P}_k = \sum_{i=0}^{2L} \alpha_i (\check{x}_{k,i} - \check{x}_k)(\check{x}_{k,i} - \check{x}_k)^T$ where $\alpha_i = \frac{\kappa}{L+\kappa} (if\ i = 0)\ or\ \frac{1}{2(L+\kappa)}$ ; 4) The predicted posterior, observation noise and observation model replace the prior, motion noise, and motion model. Then steps 1-2 repeat to compute the *desired moments*

$$\begin{cases} \mu_{y,k} = \sum_{i=0}^{2L} \alpha_i \check{y}_{k,i} \\ \Sigma_{yy,k} = \sum_{i=0}^{2L} \alpha_i (\check{y}_{k,i} - \mu_{y,k})(\check{y}_{k,i} - \mu_{y,k})^T \\ \Sigma_{xy,k} = \sum_{i=0}^{2L} \alpha_i (\check{x}_{k,i} - \check{x}_k)(\check{y}_{k,i} - \mu_{y,k})^T \end{cases}$$

where $\check{y}_{k,i}$ is the outcome of nonlinear observation model $g(\cdot)$. In the *correction step*, the procedures are the same as that in KF where the predicted posterior is corrected by incorporating the observation $y_k$, and the posterior are

$$\begin{cases} K_k = \Sigma_{xy,k} \Sigma_{y,k}^{-1} \\ \hat{x}_k = \check{x}_k + K_k (y_k - \mu_{y,k}) \\ \hat{P}_k = \check{P}_k - K_k \Sigma_{xy,k}^T \end{cases}$$

SPKF is improved by iterated SPKF (ISPKF) by computing input Sigmapoints around an operating point $x_{op,k}$ where $x_{op,k} = \begin{cases} \check{x}_k, \text{first iteration} \\ \hat{x}_k, \text{rest iteration} \end{cases}$.

Particle filter is based on the MC sampling, and its steps are as follows: 1) Generate $M$ samples from the joint density $\begin{bmatrix} \hat{x}_{k-1,m} \\ w_{k,m} \end{bmatrix} \leftarrow p(x_{k-1} | \check{x}_0, v_{1:k-1}, y_{0:k-1}) p(w_k)$ where $m$ is the particle index. The $p(x_{k-1} | \check{x}_0, v_{1:k-1}, y_{0:k-1})$ is the prior density. The $p(w_k)$ is the motion noise density; 2) Generate the predicted posterior (*prediction step*) by passing samples and input $v_k$ through nonlinear motion model $f(\cdot)$ via $\check{x}_{k,m} = f(\hat{x}_{k-1,m}, v_k, w_{k,m})$ ; 3) Correct the predicted posterior (*correction step*) by incorporating observation $y_k$ in the *resampling* process (sampling the importance sampling) via $\hat{x}_{k,m} \xleftarrow{\text{resample}} \{\check{x}_{k,m}, W_{k,m}\}$ where $W_{k,m}$ is the *importance weight* of predicted posterior (particle) based on the *divergence* between the desired posterior and predicted posterior

$$W_{k,m} = \frac{p(x_{k-1} | \check{x}_0, v_{1:k-1}, y_{0:k})}{p(x_{k-1} | \check{x}_0, v_{1:k-1}, y_{0:k-1})} = \eta p(y_k | \check{x}_{k,m})$$

where $\eta$ is a normalization constant. We assume $p(y_k | \check{x}_{k,m})$ is Gaussian $p(y_k | \check{y}_{k,m})$, and $\check{y}_{k,m}$ is acquired via simulating an expected sensor reading by passing $\check{x}_{k,m}$ through nonlinear observation model $g(\cdot)$ by $\check{y}_{k,m} = g(\check{x}_{k,m}, 0)$.

**Supplement B5.**

**Rejection algorithm and Markov chain.** Instead of sampling the target PDF $p(x)$ directly, we prefer to sample $q(x)$ which is an easy-to-sample proposal distribution and satisfies $p(x) = Mq(x), M < \infty$. Then, $x^{(i)}$ sampled from the $q(x)$ are accepted if the acceptance condition is satisfied via the rejection sampling algorithm as the bellow:

Set $i = 1$
Repeat until $i = N$
1. Sample $x^{(i)} \sim q(x)$ and $u \sim \mathcal{U}_{(0,1)}$.
2. If $u < \frac{p(x^{(i)})}{Mq(x^{(i)})}$ then accept $x^{(i)}$ and increment the counter $i$ by 1. Otherwise, reject.

where $u \sim \mathcal{U}_{(0,1)}$ denotes sampling a uniform random variable on the interval $(0,1)$. The stochastic process $x^{(i)} \in \mathcal{X} = \{x_1, x_1, \dots, x_s\}$ is called a Markov chain if $p(x^{(i)} | x^{(i-1)}, \dots, x^{(1)}) = T(x^{(i)} | x^{(i-1)})$. This means the evolution of the chain only depends on the *current state* and a *fixed transition matrix* $T$ in the space $\mathcal{X}$, and after $t$ iterations $\mu(x^{(1)})T^t$ converges to an invariant distribution $p(x)$ where $\mu(x^{(1)})$ denotes the probability of the initial state $x^{(1)}$. $T$ should have the *irreducibility* and *aperiodicity*, and invariant $p(x)$ should have the *reversibility*. The irreducibility denotes that $T$ cannot be reduced to separate smaller matrixes, while the aperiodicity denotes the chain cannot be trapped in cycles. The reversibility denotes $p\left(x^{(i)}\right)T(x^{(i-1)}|x^{(i)}) = p\left(x^{(i-1)}\right)T(x^{(i)}|x^{(i-1)})$.

**MH.** The steps of MH are shown as the bellow:

1. Initialise $x^{(0)}$.
2. For $i = 0$ to $N-1$
   - Sample $u \sim \mathcal{U}_{[0,1]}$.
   - Sample $x^{\star} \sim q(x^{\star}|x^{(i)})$.
   - If $u < \mathcal{A}(x^{(i)}, x^{\star}) = \min\left\{1, \frac{p(x^{\star})q(x^{(i)}|x^{\star})}{p(x^{(i)})q(x^{\star}|x^{(i)})}\right\}$

     $x^{(i+1)} = x^{\star}$

     else

     $x^{(i+1)} = x^{(i)}$

First, the candidate sample $x^*$ is sampled from proposal distribution $q(x^*|x^{(i)})$ given the current sample $x^{(i)}$. Then, Markov chain moves to the candidate sample which works as the current sample if acceptance condition $u < \mathcal{A}(x^{(i)}, x^*)$ is satisfied. The proposal distribution $q(x^*|x^{(i)})$ is self-designed, and it can be Gaussian $\mathcal{N}(x^{(i)}, \sigma^2)$ for example. The choice of covariance $\sigma^2$ results in different width of proposal distribution, and therefore it should be *balanced*. Narrow proposal distribution means only one mode of $p(x)$ is visited, while wide proposal results in high rejection rate and subsequent high correlation of samples. Given the rejection sampling algorithm and Markov chain, the transition of MH can be constructed as

$$K_{MH}\left(x^{(i+1)}\middle|x^{(i)}\right) = q\left(x^{(i+1)}\middle|x^{(i)}\right)\mathcal{A}\left(x^{(i)}, x^{(i+1)}\right) + \delta_{x^{(i)}}(x^{(i+1)})r(x^{(i)})$$

where $r\left(x^{(i)}\right) = \int q\left(x^*\middle|x^{(i)}\right)(1 - \mathcal{A}(x^{(i)}, x^*))dx^*$ is a rejection-related term. The $\delta_{x^{(i)}}(x^{(i+1)})$ denotes the delta-Dirac mass located at $x^{(i)}$ and the proposal distribution $q\left(x^{(i+1)}\middle|x^{(i)}\right) = \frac{1}{N}\sum_{i=1}^{N}\delta_{x^{(i)}}(x^{(i+1)})$. If the proposal distribution is independent of the current state $q\left(x^*\middle|x^{(i)}\right) = q(x^*)$ and the proposal is assumed to be a symmetric random walk proposal $q\left(x^*\middle|x^{(i)}\right) = q(x^{(i)}|x^*)$ where the acceptance ratio $\mathcal{A}$ can be simplified further, this will result in *independent sampler* and *Metropolis algorithm* respectively.

**Gibbs.** The step of Gibbs is shown as the bellow:

1. Initialise $x_{0,1:n}$.
2. For $i = 0$ to $N-1$
   - Sample $x_1^{(i+1)} \sim p(x_1|x_2^{(i)}, x_3^{(i)}, \ldots, x_n^{(i)})$.
   - Sample $x_2^{(i+1)} \sim p(x_2|x_1^{(i+1)}, x_3^{(i)}, \ldots, x_n^{(i)})$.

     ⋮
   - Sample $x_j^{(i+1)} \sim p(x_j|x_1^{(i+1)}, \ldots, x_{j-1}^{(i+1)}, x_{j+1}^{(i)}, \ldots, x_n^{(i)})$.

     ⋮
   - Sample $x_n^{(i+1)} \sim p(x_n|x_1^{(i+1)}, x_2^{(i+1)}, \ldots x_{n-1}^{(i+1)})$.

Gibbs sampler can be seen as a special case of MH where the proposal distribution of Gibbs sampler is

$$q\left(x^*\middle|x^{(i)}\right) = \begin{cases} p\left(x_j^*\middle|x_{-j}^{(i)}\right), \text{If } x_{-j}^* = x_{-j}^{(i)} \\ 0, \text{Otherwise} \end{cases}$$

Given that $p(x)$ here refers to the full conditionals $p(x_j|x_{-j})$, the acceptance ratio $\mathcal{A}(x^{(i)}, x^*)$ is written as $\mathcal{A}\left(x^{(i)}, x^*\right) = \min\{1, \frac{p(x^*)q(x^{(i)}|x^*)}{p(x^{(i)})q(x^*|x^{(i)})}\} = 1$. This is the reason why new samples can be drawn directly without satisfying the acceptance conditions.

In the case of directed acyclic graphs, PDF $p(x)$ is written by $p(x) = \prod_j p(x_j|x_{pa(j)})$ and full conditions $p\left(x_j\middle|x_{-j}\right)$ is written by $p\left(x_j\middle|x_{-j}\right) = p(x_j|x_{pa(j)})\prod_{k\in ch(j)}p(x_k|x_{pa(k)})$ where $pa(j)$ and $ch(j)$ denote the parent and children's nodes of $x_j$ respectively. The parent $pa(j)$, children $ch(j)$, and children's parents $pa(k), k \in ch(j)$ are known as the *Markov basket*.

**Supplement B6.**

| Name of the term | Mathematic form | Additional description |
|---|---|---|
| Expected value of the reward $R(s,a)$ | $\bar{r}(s,a) = \int r(s,a)q(dr|s,a)$ | - $r(s,a)$ is the realization of $R(s,a)$<br>- $R(s,a) \sim q(\cdot\,|s,a)$ is a random variable representing the reward obtained when action $a$ is taken in state $s$ |

| | | |
|---|---|---|
| Transition density | $P^{\mu}\left(z'\middle\|z\right) = P\left(s'\middle\|s,a\right)\mu\left(a'\middle\|s'\right)$ | - $P_0^{\mu}(z_0) = P_0(s_0)\mu(s_0\|a_0)$<br>- $z = (s,a)$<br>- Policy $\mu: S \rightarrow A$ |
| Transition density on a trajectory $\xi$ | $Pr^{\mu}(\xi\|\mu) = P_0^{\mu}(z_0)\prod_{t=1}^{T} P^{\mu}(z_t\|z_{t-1})$ | --- |
| Reward discount (return) on a trajectory | $\bar{\rho}(\xi) = \sum_{t=0}^{T}\gamma^t\bar{r}(z_t)$ | - The discount $\gamma \in [0,1]$ |
| Expected return | $\mathbb{E}[\bar{\rho}(\xi)] = \int \bar{\rho}(\xi)Pr^{\mu}(\xi\|\mu)d\,\xi$ | --- |
| Discounted return | $D^{\mu}(s) = \sum_{t=0}^{T}\gamma^t\bar{r}(z_t)\|\,z_0 = (s,\mu(\cdot\|s))$ | - $s_{t+1} = P^{\mu}(\cdot\|s_t)$ |
| Value function | $V^{\mu}(s) = \mathbb{E}[D^{\mu}(s)]$ | --- |
| Action-value function | $Q^{\mu}(z) = \mathbb{E}[D^{\mu}(z)] = \sum_{t=0}^{T}\gamma^t\bar{r}(z_t)\|\,z_0 = z$ | --- |
| Bellman equation for $V^{\mu}(s)$ | $V^{\mu}(s) = R^{\mu}(s) + \gamma\int P^{\mu}\left(s'\middle\|s\right)V^{\mu}(s')ds',\ s \in S$ | - $R^{\mu}(s)$ is the immediate reward |
| Bellman optimal equation | $V^*(s) = \max_{a\in A}\left[R^{\mu}(s) + \gamma\int P^{\mu}\left(s'\middle\|s\right)V^*(s')ds'\right], s \in S$ | - $V^*(s) = V^{\mu^*}(s)$ |
| Value iteration to find the optimal value function | $V_i^*(s) = \max_{a\in A}\left[R^{\mu}(s) + \gamma\int P^{\mu}\left(s'\middle\|s\right)V_{i-1}^*(s')ds'\right], s \in S$ | - Near all methods to find the optimal solution of an MDP are based on the *value iteration* and *policy iteration* of dynamic programming (DP) algorithms<br>- Policy iteration: The policy iteration consists of *policy evaluation* and *policy improvement* where value function is evaluated and improved until the convergence of value function |

MDP is defined as a tuple $< S, A, P, P_0, q(\cdot|s,a) >$ where $S$ is the states, $A$ the actions, $P(\cdot|s,a)$ probability distribution over next states or transition density, $P_0$ the initial transitional probability distribution. $q(\cdot|s,a)$ a random variable which represents reward $R(s,a)\sim q(\cdot|s,a)$. POMDP is defined as a tuple $< S, A, O, P, \Omega, P_0, q(\cdot|s,a) >$. The difference between POMDP and MDP is $\Omega(\cdot|s,a) = P(O)$ where $\Omega$ is the transition density of the observation $O$ which is observed after executing action $a$. Bellman optimal equation of POMDP is written as

$$V^*(b_t) = \max_{a\in A}\left[\iint R(s,a)b_t(s)ds + \gamma\int Pr(o|b_t,a)V^*(\tau(b_t,a,o))do\right]$$

where $b_t(s)$ is the *information state* (*belief*) which represents the state under partial observation. The immediate reward cannot be obtained directly, but it can be computed indirectly by $\int R(s,a)b_t(s)ds$. Transition density also cannot be computed directly, therefore it is computed indirectly by $\int Pr(o|b_t,a)do$ conditioned on the historical observations and information state. $\tau(b_t,a,o)$ represents the next information state, and it is defined and computed recursively by

$$\tau(b_t,a,o) = b_{t+1}(s') = \frac{\Omega(o_{t+1}|s',a_t)\int P(s'|s,a_t)b_t(s)ds}{\int \Omega(o_{t+1}|s'',a_t)\int P(s''|s,a_t)\,b_t(s)dsds''}$$

which describes the relationship of the information state, observation, and state. Bellman optimality equation of POMDP is proved to be piecewise-linear and convex, therefore it can be computed by

$$V^*(b_t) = \max_{a\in\Gamma_t}\int \alpha(s)b_t(s)ds$$

where the linear segment $\Gamma_t = \{\alpha_0,\alpha_1..,\alpha_m\}$. $\alpha$ is the $\alpha$-function, and $\alpha_i(b_t) = \int \alpha_i(s)b_t(s)ds$.

In POMDP, the belief (information state) after $t$ step trajectory is written as $b_{t+1}(s') = \frac{\Omega(o_{t+1}|s',a_t)\int P(s'|s,a_t)b_t(s)ds}{\int \Omega(o_{t+1}|s'',a_t)\int P(s''|s,a_t)b_t(s)dsds''}$. Recall that if the unknown variables $x$ is captured by two independent distributions $p_1(x)$ and $p_2(x)$, its probability density $p(x)$ is written as $p(x) = \eta p_1(x)p_2(x), \eta = (\int p_1(x)p_2(x)dx)^{-1}$ or $p(x|y_1,y_2) =$

$\eta p(x|y_1)p(x|y_2), \eta = \frac{p(y_1)p(y_2)}{p(y_1,y_2)p(x)}$ in Bayesian context. Similarly, in the case of *belief-MDP* (e.g., POMDP), we assume the unknown transition $\theta$ is captured by *two* independent distributions $\theta_{s,a,s'}$ and $\theta_{s,a,r}$, and the belief is computed by $b_t(\theta) = \prod_{s,a} b_t(\theta_{s,a,r}) b_t(\theta_{s,a,s'})$ where $\theta_{s,a,s'}$ is the unknown probability transition from state $s$ to state $s'$ after the action $a$, and $\theta_{s,a,r}$ is the unknown probability to obtain the reward $r$. Hence, the belief $b_{t+1}(\theta')$ in the next time step after learning from the samples can be written as $b_{t+1}(\theta') = \eta \Pr(s', r|s, a, \theta') \int \Pr(s', \theta'|s, a, \theta)\, b_t(\theta) ds d\theta$ where $\eta$ is the normalization factor, $\theta$ the unknown transition probability. $\Pr(s', r|s, a, \theta')$ corresponds to the *updated reward probability*, while $\int \Pr(s', \theta'|s, a, \theta)\, b_t(\theta) ds d\theta$ corresponds to the *updated state transition probability*. The reward probability $\Pr(s', r|s, a, \theta')$ is computable, but the state transition probability $\int \Pr(s', \theta'|s, a, \theta)\, b_t(\theta) ds d\theta$ is intractable because it requires the computation over all observable states and possible belief states. This process is computationally expensive. The computation of the intractable integral in the belief-MDP can be simplified by BAMDP where the probability densities are captured by the Dirichlet distribution.

**Supplement B7.**

The *hyper-state transition density* $\Pr(s', \phi'|s, a, \phi)$ of BAMDP is factorized by

$$\Pr(s', \phi'|s, a, \phi) = \Pr(s'|s, a, \phi) \Pr(\phi'|s, a, s', \phi)$$

where the first part $\Pr(s'|s, a, \phi) = \frac{\phi'_{s,a,s'}}{\sum_{s'' \in S} \phi'_{s,a,s''}}$ by taking the expectation over all possible transition functions, and the second part $\Pr(\phi'|s, a, s', \phi) = \begin{cases} 1 \text{ if } \phi'_{s,a,s'} = \phi_{s,a,s'} + 1 \\ 0, otherwise \end{cases}$ since the update of $\phi$ to $\phi'$ is deterministic. Hence, the hyper-state transition density is written as

$$P'(s', \phi'|s, a, \phi) = \frac{\phi'_{s,a,s'}}{\sum_{s'' \in S} \phi'_{s,a,s''}} \Pr(\phi'|s, a, s', \phi) = \frac{\phi'_{s,a,s'}}{\sum_{s'' \in S} \phi'_{s,a,s''}} \mathbb{I}(\phi'_{s,a,s'} = \phi_{s,a,s'} + 1)$$

Then, the hyper-state transition density is used to compute the policy where the action is selected via the value function. Given the Bellman optimal equation, the optimal value function at a time step is written as

$$V^*(s, \phi) = \max_{a \in A} \left[ R'(s, \phi, a) + \gamma \sum_{(s', \phi') \in S'} P'(s', \phi'|s, a, \phi) V^*(s', \phi') \right]$$
$$= \max_{a \in A} \left[ R(s, a) + \gamma \sum_{s' \in S'} P'(s', \phi'|s, a, \phi) V^*(s', \phi') \right]$$

To ensure the convergence of the policy, the value iteration is applied to compute the optimal policy of BAMDP via

$$V_t{}^*(s, \phi) = \max_{a \in A} \left[ R^\mu(s) + \gamma \int \mathrm{P}^\mu(s'|s) V^*_{t-1}(s') ds' \right] = \max_{a \in A} \left[ R(s, a) + \gamma \int \mathrm{P}'(s', \phi'|s, a, \phi) V^*_{t-1}(s', \phi') ds' d\phi' \right]$$

**Supplement B8.**

**Value approximation methods.** Offline value approximation. The finite-state controllers use the graph to define BAMDP where the nodes denote the memory states and the edges denote the observations. Then, the expected value is computed expensively in closed form by recursively applying Bellman equations. The difference between BEETLE and POMDP is that the hyper-states in BEETLE are randomly *sampled* from interactions, but this method is still computationally expensive.

online near-myopic value approximation. Bayesian dynamic programming samples a model from the posterior distribution over parameters, based on dynamic programming via the simulation. This method ignores the posterior uncertainty and causes slow convergence of the posterior. The convergence can be improved by keeping maximum likelihood estimations of the value function. The value of information heuristic considers the expected return and expected value. It uses Dirichlet distribution to capture the distribution over the action value $Q^*(s, a)$ which is used to estimate the improvement of policy (one-step estimation). However, one-step estimation is myopia and results in suboptimal convergence.

Online tree search approximation. The method forward search tree builds a fixed depth forward search tree which includes the hyper-states with limited steps (depth $d$) in a trajectory. The leaf nodes of the tree uses the immediate reward as the default value function which is necessary but naive. BAMCP incorporates two policies to traverse and grow the forward search tree. Two policies are the upper confidence bounds (UCT) applied to trees and the rollout, respectively. The action is selected by UCT and value function via $a^* = \arg\max_a Q(s, h, a) + c\sqrt{\frac{\log n(s,h)}{n(s,h,a)}}$ where $Q(s, h, a)$ is the value function which can be the immediate reward, $c$ the constant, $n(s, h)$ the number of times where the node corresponding to state $s$ and history $h$ has been visited in the tree, $n(s, h, a)$ the number of times where action $a$ was chosen in this node. An untried action is selected to start the rollout until the terminal node. The node visited will not be added to the tree, and $n(s, h)$ and $n(s, h, a)$ are saved in rollout process. Just one model is sampled from the posterior at the tree root to reduce the computation. The rollout is navigated to accelerate the search by a model-free value estimation $\hat{Q}(s_t, a_t) \leftarrow \hat{Q}(s_t, a_t) +$ $\alpha(r_t + \gamma \max_a \hat{Q}(s_{t+1}, a) - \hat{Q}(s_t, a_t))$ where $\alpha$ is a learning rate.

**Exploration Bonus Methods.** Value function of BEB is written by

$$\tilde{V}_t^{\,*}(s,\phi) = \max_{a\in A}\left[R(s,a) + \frac{\beta}{1+\sum_{s'\in S}\phi_{s,a,s'}} + \sum_{s\in S} P(s',\phi'|s,a,\phi)\tilde{V}_{t-1}^{*}(s',\phi')\right]$$

where $\frac{\beta}{1+\sum_{s'\in S}\phi_{s,a,s'}}$ is the exploration bonus. $\beta$ is the constant that decides the magnitude of the bonus. The exploration bonus decays with $\frac{1}{n}$ where $n\sim \sum_{s'\in S}\phi_{s,a,s'}$.

In VBRB, the modified value function is

$$\tilde{V}_t^{\,*}(s,\phi) = \max_{a\in A}\left[R(s,\phi,a) + \hat{R}_{s,\phi,a} + \sum_{s\in S} P(s',\phi'|s,a,\phi)\tilde{V}_{t-1}^{*}(s',\phi')\right]$$

where the bonus $\hat{R}_{s,\phi,a}$ is defined as

$$\begin{cases} \hat{R}_{s,\phi,a} = \beta_R \sigma_{R(s,\phi,a)} + \beta_P \sqrt{\sum_{s'\in S}\sigma^2_{P(s',\phi'|s,a,\phi)}} \\ \sigma^2_{R(s,\phi,a)} = \int R(s,\theta,a)^2 b(\theta)d\theta - R(s,\phi,a)^2 \\ \sigma^2_{P(s',\phi'|s,a,\phi)} = \int P(s'|s,\theta,a)^2 b(\theta)d\theta - P(s',\phi'|s,a,\phi)^2 \end{cases}$$

where $\beta_R$ and $\beta_P$ are constants to decide the magnitude of reward variance $\sigma^2_{R(s,\phi,a)}$ and transition variance $\sigma^2_{P(s',\phi'|s,a,\phi)}$, respectively.

**Supplement B9.**

**Value function Bayesian RL.** Recall that *value function* is defined to be expected value of discounted return $V^\mu(s) = \mathbb{E}[D^\mu(s)]$, therefore discounted return is decomposed to $D(s) = \mathbb{E}[D(s)] + D(s) - \mathbb{E}[D(s)] = V(s) + \Delta V(s)$ where $\Delta V(s) = D(s) - V(s)$. Also recall that discounted return is defined by $D^\mu(s) = \sum_{t=0}^{T}\gamma^t \bar{r}(z_t)|_{z_0=(s,\mu(\cdot|S)),\, s_{t+1}=P^\mu(\cdot|s_t)}$. Hence, discounted return of current state $s$ and next state $s'$ is defined by $D(s) = R(s) + \gamma D(s')$ where $R(s)$ is the reword received in current state. Hence, it is easy to see

$$\begin{aligned} R(s) = D(s) - \gamma D(s') = V(s) + \Delta V(s) - \gamma\big(V(s') + \Delta V(s')\big) = V(s) - \gamma V(s') + \Delta V(s) - \gamma \Delta V(s') \\ = V(s) - \gamma V(s') + N(s,s'), N(s,s') = \Delta V(s) - \gamma \Delta V(s') \end{aligned}$$

Above equation can be rewritten as $\boldsymbol{R_{T-1}} = \boldsymbol{HV_T} + \boldsymbol{N_{T-1}}$ where $\boldsymbol{R_T} = (R(s_0), \dots R(s_{T-1}))^{\boldsymbol{T}}$, $\boldsymbol{V_{T+1}} = (V(s_0), \dots V(s_T))^T$, $\boldsymbol{N_T} = (N(s_0,s_1), \dots N(s_{T-1},s_T))^T$, and $\boldsymbol{H}$ is the $T\times T$ matrix.

**Bayesian quadrature.** Recall that the performance of RL policy in a trajectory is measured by the expected return $\eta(\mu) = \mathbb{E}[\bar{\rho}(\xi)] = \int \bar{\rho}(\xi) Pr^\mu(\xi|\mu) d\xi$ where the reward discount (return) is defined by $\bar{\rho}(\xi) = \sum_{t=0}^{T}\gamma^t \bar{r}(z_t), \gamma \in [0,1]$. The transition density on a trajectory $\xi$ is defined by $Pr^\mu(\xi|\mu) = P_0^\mu(z_0)\prod_{t=1}^{T} P^\mu(z_t|z_{t-1})$. The policy is updated by the gradient of expected return via $\nabla\eta(\theta) = \int \bar{\rho}(\xi)\frac{\nabla \Pr(\xi;\theta)}{\Pr(\xi;\theta)}\Pr(\xi;\theta)\, d\xi = \int \bar{\rho}(\xi)\boldsymbol{u}(\boldsymbol{\xi};\boldsymbol{\theta})\Pr(\xi;\theta)\, d\xi$ where $\theta$ is the approximation of policy. $\frac{\nabla \Pr(\xi;\theta)}{\Pr(\xi;\theta)}$ is the likelihood ratio of trajectory (Fisher score function), and it can be rewritten as $\boldsymbol{u}(\boldsymbol{\xi};\boldsymbol{\theta}) = \frac{\nabla \Pr(\xi;\theta)}{\Pr(\xi;\theta)} = \nabla \log \Pr(\xi;\theta) = \sum_{t=0}^{T-1}\nabla \log \mu(a_t|s_t;\theta)$. It is hard to compute the $\nabla\eta(\theta)$ directly when $\nabla\eta(\theta)$ is in the form of integral $\int \bar{\rho}(\xi)\boldsymbol{u}(\boldsymbol{\xi};\boldsymbol{\theta})\Pr(\xi;\theta)\, d\xi$.

BQ solves the computing of integral by simplifying the integral to the form

$$\zeta = \nabla\eta(\theta) = \int f(x)g(x)\, dx$$

where $f(x)$ is *unknown* and it can be solved via Bayesian inference to infer its posterior. $g(x)$ is a *known* function. It is possible to model $f(x)$ as GP by specifying Normal prior to $f(x)$. Therefore, $f(\cdot)\sim\mathcal{N}(\bar{f}(\cdot), k(\cdot,\cdot))$, and $\mathbb{E}[f(x)] = \bar{f}(\cdot), Cov[f(x), f(x')] = k(x,x')$. Let samples $D_M = \{(x_i, y(x_i))\}_{i=1}^{M}$ where $y(x)$ is the noise-corrupted samples of $f(x)$. To infer the posterior of $f(x)$, it is possible to apply Gaussian prior to $f(x)$ where $f(x) \sim \mathcal{N}\big(\bar{f}, \mathbf{K}\big), \boldsymbol{K}_{i,j} = k(x_i, x_j)$, therefore the mean $\mathbb{E}[f(x)|D_M]$ and covariance $\boldsymbol{Cov}[f(x), f(x')|D_M]$ are easy to compute. Assuming that the integral $\zeta$ is *linear*, posterior moments of $\zeta$ are given by

$$\begin{cases} \mathbb{E}[\zeta|D_M] = \int \mathbb{E}[f(x)|D_M]g(x)dx = \zeta_0 + \boldsymbol{b}^T\boldsymbol{C}(\boldsymbol{y} - \bar{\boldsymbol{f}}) \\ \boldsymbol{Var}[\zeta|D_M] = \iint \boldsymbol{Cov}[f(x), f(x')|D_M]g(x)g(x')dxdx' \\ \qquad = b_0 - \boldsymbol{b}^T\boldsymbol{C}\boldsymbol{b} \end{cases}$$

where $\boldsymbol{Var}[\zeta|D_M]$ is the gradient variance, $\zeta_0 = \int \bar{f}(x)g(x)\,dx$, $\boldsymbol{b} = \int \boldsymbol{k}(x)g(x)\,dx$, and $b_0 = \iint k(x,x')g(x)g(x')dxdx'$.

**Bayesian policy gradient.** Given BQ, the gradient of BPG $\nabla\eta(\theta) = \int \bar{\rho}(\xi)\nabla\log\Pr(\xi;\theta)\Pr(\xi;\theta)\,d\,\xi$ can be simplified to the form $\int f(x)g(x)\,dx$ where

$$\begin{cases} f(x) = \bar{\rho}(\xi)\nabla\log\Pr(\xi;\theta) \\ g(x) = \Pr(\xi;\theta) \end{cases}$$

or $f(x) = \bar{\rho}(\xi)$ and $g(x) = \nabla\log\Pr(\xi;\theta)\Pr(\xi;\theta)$. $f(x)$ is an unknow function modeled as GP. $g(x)$ should be a known function, but in BPG, the trajectory transition function $\Pr(\xi;\theta)$ is unknown. However, it is sufficient to assign $\Pr(\xi;\theta)$ to $g(x)$ via a Fisher kernel, instead of using $\Pr(\xi;\theta)$ directly. When $f(x) = \bar{\rho}(\xi)\nabla\log\Pr(\xi;\theta)$ and $g(x) = \Pr(\xi;\theta)$, Fisher kernel written as $k(\xi,\xi') = (1+\boldsymbol{u}(\xi)^T\boldsymbol{G}^{-1}\boldsymbol{u}(\xi'))^2, \boldsymbol{G}(\theta) = \mathbb{E}[\boldsymbol{u}(\xi)\boldsymbol{u}(\xi)^T]$ is selected to solve the integral where $\boldsymbol{G}(\theta)$ is Fisher information matrix. Finally, the moments of gradient are obtained via

$$\begin{cases} \mathbb{E}[\nabla\eta(\theta)|D_M] = \boldsymbol{YCb} \\ \boldsymbol{Var}[\nabla\eta(\theta)|D_M] = (b_0 - \boldsymbol{b}^T\boldsymbol{Cb})\boldsymbol{I} \\ (\boldsymbol{b})_i = 1 + \boldsymbol{u}(\xi_i)^T\boldsymbol{G}^{-1}\boldsymbol{u}(\xi_i) \\ b_0 = 1 + n \end{cases}$$

The computing of gradient moments when $f(x) = \bar{\rho}(\xi)$ and $g(x) = \nabla\log\Pr(\xi;\theta)\Pr(\xi;\theta)$ is the same as the method above.

**Bayesian actor-critic.** The observation unit of classical policy gradient is a trajectory, while in classical actor-critic the observation is one-step transition. Hence, the expected return of the policy $\mu$ in actor-critic may be written as $\eta(\mu) = \int w^{\mu}(z)\bar{r}(z)dz$ where $z = (s,a)$, and $w^{\mu}(z) = \sum_{t=0}^{\infty}\gamma^t P_t^{\mu}(z)$ denotes the *discounting weight* of $z$. The discounting weight can be rewritten as $w^{\mu}(s) = \int w^{\mu}(s,a)da$ over the state $s$. $P_t^{\mu}$ is the $t$-step state-action occupancy density defined by $P_t^{\mu}(z_t) = \int \prod_{i=1}^{t} P^{\mu}(z_i|z_{i-1})\,dz_0 \ldots dz_{t-1}$ . Therefore, the gradient of expected return in actor-critic is $\nabla\eta(\theta) = \int w^{\mu}(s,a;\theta)\nabla\mu(a|s;\theta)Q(s,a;\theta)\,dsda$.

In actor-critic, the actor is updated according to $\nabla\eta(\theta)$ to obtain new actor for generating new action and action value, while the critic provides action value $Q(s,a;\theta)$ for the update of the actor. The actor and critic contribute to the convergence of each other.

In Bayesian AC, the computing of integral $\int w(s;\theta)\nabla\mu(a|s;\theta)Q(s,a;\theta)\,dsda$ is simplified to the form $\int f(x)g(x)\,dx$ as that of BPG. Let

$$\begin{cases} f(x) = Q(z;\theta), z = (s,a) \\ g(x) = g(z;\theta) = w^{\mu}(z;\theta)\nabla\mu(a|s;\theta) \end{cases}$$

According to GPTD, it is easy to obtain the posterior moments of $Q(z;\theta)$ after $t$ time-steps, that are $\mathbb{E}[Q(z)|D_t]$ and $\boldsymbol{Cov}[Q(z),Q(z')|D_t]$. According to the *linearization* of the integral $\int w(s;\theta)\nabla\mu(a|s;\theta)Q(s,a;\theta)dsda$, it is easy to obtain the moments of policy gradient $\nabla\eta(\theta)$

$$\begin{cases} \mathbb{E}[\nabla\eta(\theta)|D_t] = \int g(z;\theta)\mathbb{E}[Q(z)|D_t]dz \\ \boldsymbol{Cov}[\nabla\eta(\theta)|D_t] = \\ \iint g(z;\theta)\boldsymbol{Cov}[Q(z),Q(z')|D_t]g(z';\theta)^T dzdz' \end{cases}$$

This is the general form of $\nabla\eta(\theta)$ posterior, and the following computing of gradient posterior is almost the same as that of BPG.

**Supplement B10.**

**Bayesian IRL.** Due to the assumption that $f(x)$ and $e_i$ are the Gaussians, the noisy reward function is written as $\boldsymbol{y}\sim\mathcal{N}(\boldsymbol{\mu},\boldsymbol{K_y})$ where $\boldsymbol{\mu} = \mu(x_i)$, and $(K_y)_{ij} \coloneqq k(x_i,x_j) + \sigma^2\delta_{ij}$. Hence, the posterior of the reward function can be written as

$$f(x)|\mathcal{D}\sim\mathcal{N}(\boldsymbol{\mu_p},\boldsymbol{\sigma_p^2})$$

The posterior moments here are written by $\boldsymbol{\mu_p} = \boldsymbol{\mu} + \boldsymbol{k}(x)^T\boldsymbol{K_y^{-1}}(\boldsymbol{y}-\mu)$ and $\boldsymbol{\sigma_p^2} = k(x,x) - \boldsymbol{k}(x)^T\boldsymbol{K_y^{-1}}\boldsymbol{k}(x)$ where $\boldsymbol{k}(x) = \left(k(x_1,x),k(x_2,x)..k(x_T,x)\right)^T$.

The kernel $k(x,x')$ can be parameterized as $k(x,x';\theta)$ where $\theta$ is the hyperparameter that can be acquired by maximizing the log likelihood

$$\theta = \arg\max_{\theta}\mathcal{L}(\theta|\mathcal{D})$$

and the hyperparameter is learnt by the gradient ascent $\theta \leftarrow \theta + B\nabla_{\theta}\mathcal{L}(\theta|\mathcal{D})$ where $B$ is a recursive approximation of the inverse Hessian. The convergence of learning the hyperparameter can be further stabilized and improved by $l_1$-regularizer, therefore the learning objective is written as

$$\text{minimize} - \mathcal{L}(\theta|\mathcal{D}) + \lambda\|\boldsymbol{w}\|_1, \text{subject to} - \theta_i \leq 0, i = 1,\ldots,P+2$$

where $\boldsymbol{w} = [w_1, w_1, \dots, w_P]^T$ is the regression coefficient vector and $\lambda$ is a discounted factor.

## Supplement C

### Supplement C1.

**VI to assist deriving RL policy.** The system can be modeled as probabilistic graphical model (PGM) and computes the optimal parameters of PGM by maximizing the likelihood $\log \prod_{(\vec{x},\vec{m},\vec{a},\vec{r})\in\mathcal{D}} p(\vec{x},\vec{m},\vec{O}^p|\vec{a})$ (sensor inputs $\vec{x}$, mask $\vec{m}$, action $\vec{a}$, optimality $\vec{O}^p$ which is a reward-related binary random parameter, $p$ "post", and $\mathcal{D}$ dataset). It is impossible to compute the likelihood directly, but ELBO can be derived. That is

$$ELBO = \mathbb{E}_{q(\vec{z}^w, \vec{a}^p|\vec{x},\vec{a})}\left[\log p(\vec{x},\vec{m},\vec{O}^p,\vec{z}^w,\vec{a}^p|\vec{a}) - \log q(\vec{z}^w,\vec{a}^p|\vec{x},\vec{a})\right] \le \log p(\vec{x},\vec{m},\vec{O}^p|\vec{a})$$

where likelihood is expanded with latent state $\log \iint p(\vec{x},\vec{m},\vec{O}^p,\vec{z}^w,\vec{a}^p|\vec{a}) \frac{q(\vec{z}^w,\vec{a}^p|\vec{x},\vec{a})}{q(\vec{z}^w,\vec{a}^p|\vec{x},\vec{a})} d\vec{z}^w d\vec{a}^p$ (latent state $\vec{z}^w$ and $w$ "whole trajectory $\tau$ with length $H$"), and posterior is approximated by VI with proper variational distribution defined by

$$q(\vec{z}^w,\vec{a}^p|\vec{x},\vec{a}) = q(\vec{z}|\vec{x},\vec{a})\pi(a_{\tau+H}|z_{\tau+H}) \prod_{t=\tau+1}^{\tau+H-1} p(z_{t+1}|z_t,a_t)\pi(a_t|z_t)$$

where $\pi(a_{\tau+H}|z_{\tau+H}) \prod_{t=\tau+1}^{\tau+H-1} p(z_{t+1}|z_t,a_t)\pi(a_t|z_t)$ is the distribution of trajectory $\tau$ and it is acquired by executing policy $\pi$ with latent state transition $p$. $q(\vec{z}|\vec{x},\vec{a})$ denotes the posterior of latent states. Finally, the problem is simplified to maximizing ELBO where $\mathbb{E}_{q(\vec{z}^w, \vec{a}^p|\vec{x},\vec{a})}\left[\log p(\vec{x},\vec{m},\vec{O}^p,\vec{z}^w,\vec{a}^p|\vec{a})\right]$ corresponds to the learning of generative models. The second $ELBO$ $\mathbb{E}_{q(\vec{z}^w, \vec{a}^p|\vec{x},\vec{a})}[-\log q(\vec{z}^w,\vec{a}^p|\vec{x},\vec{a})]$ corresponds to the learning of RL model with latent state $\vec{z}^w$. These two parts of $ELBO$ can be maximized jointly to constrain the learning of generative models and RL policy, therefore RL policy and generative models are acquired simultaneously.

**VI to assist approximating agent's reward function (extrinsic and intrinsic).** The maximizing ELBO on *log-likelihood of future states* results in the objective

$$J = \mathbb{E}_{z_{t+1\sim q(z)}}[\log p_\theta(s_{t+1}|z_{t+1})] - \beta D_{KL}[q_\theta(z_{t+1}|s_t,a_t,s_{t+1})||p_\theta(z_{t+1}|s_t,a_t)]$$

where $\log p_\theta(s_{t+1}|z_{t+1})$ is the reconstruction model which reconstructs the future state $s_{t+1}$ from the latent future state $z_{t+1}$. $\beta$ is to control disentanglement in the latent representation. Variational distribution $q_\theta(z_{t+1}|s_t,a_t,s_{t+1})$ approximates the true posterior of the latent variable. $p_\theta(z_{t+1}|s_t,a_t)$ is latent prior. The information gain $\mathcal{I}$ approximated by Bayesian surprise $D_{KL}[q_\theta(z_{t+1}|s_t,a_t,s_{t+1})||p_\theta(z_{t+1}|s_t,a_t)$ can be taken as the intrinsic reward via

$$\begin{aligned} r_t^i = \mathcal{I}(z_{t+1}; s_{t+1}|s_t,a_t) &\approx D_{KL}[q_\theta(z_{t+1}|s_t,a_t,s_{t+1})||p_\theta(z_{t+1}|s_t,a_t) = \mathbb{E}_{q_\theta(z_{t+1}|\cdot)}[\log q_\theta(z_{t+1}|\cdot) - \log p_\theta(z_{t+1}|\cdot)] \\ &= -H[q_\theta(z_{t+1}|\cdot)] + H[q_\theta(z_{t+1}|\cdot), p_\theta(z_{t+1}|\cdot)] \end{aligned}$$

This means the maximum parameters can be acquired by searching for states with minimal entropy of the posterior $-H[q_\theta(z_{t+1}|\cdot)]$ and a high cross-entropy value between the posterior and the prior $H[q_\theta(z_{t+1}|\cdot), p_\theta(z_{t+1}|\cdot)$.

### Supplement C2.

Training objective $p(\theta|X,Y) = p(Y|X,\theta)p(\theta)/p(Y|X)$ is computed by BNN (with dropout VI) where Q function is parameterized by $\theta$. $X$ capture all the state-action pairs in the training set, while $Y$ capture true Q value of the states. Dropout VI trains RL with dropout before each weighted layer, and also performs dropout at test time, resulting in a variance defined by

$$Var[Q(s,a)] \approx \sigma^2 + \frac{1}{N}\hat{Q}_t(s,a)^T\hat{Q}_t(s,a) - \mathbb{E}[\hat{Q}(s,a)]^T[\hat{Q}(s,a)]$$

where $\sigma$ is the data noise and $N$ is the number of stochastic forward passes. $\frac{1}{N}\hat{Q}_t(s,a)^T\hat{Q}_t(s,a)$ represents *how much the model is uncertain about its predictions* $\hat{Q}_t(s,a)$. The OOD samples are detected and decided by $\frac{1}{N}\hat{Q}_t(s,a)^T\hat{Q}_t(s,a) - \mathbb{E}[\hat{Q}(s,a)]^T[\hat{Q}(s,a)]$. $\mathbb{E}[\hat{Q}(s,a)]$ is the mean of prediction. The uncertainty-weighted policy distribution $\pi'(a|s)$ is defined by inserting the variance $Var[Q(s,a)]$ to old policy distribution $\pi(a|s)$ via

$$\pi'(a|s) = \frac{\frac{\beta}{Var\left[Q_\theta^{\pi'}(s,a)\right]}\pi(a|s)}{Z(s)}$$

where $Z(s) = \int \frac{\beta}{Var\left[Q_\theta^{\pi'}(s,a)\right]}\pi(a|s)da$. Penalty of OOD data is achieved by inserting variance $Var[Q(s,a)]$ to the actor loss $\mathcal{L}(\pi)$ and critic loss $\mathcal{L}(Q_\theta)$ via

$$\mathcal{L}(Q_\theta) = \mathbb{E}_{(s'|s,a)\sim\mathcal{D}}\mathbb{E}_{a'\sim\pi(\cdot|s')}\left[\frac{\beta}{Var\left[Q_{\theta'}\left(s',a'\right)\right]}Err(s,a,s',a')^2\right]$$

$$\mathcal{L}(\pi) = -\mathbb{E}_{a\sim\pi(\cdot|s)}\left[\frac{\beta}{Var[Q_\theta(s,a)]}Q_\theta(s,a)\right]$$

where $Err = Q_\theta(s,a) - \left[R(s,a) + \gamma Q_{\theta'}\left(s',a'\right)\right]$. $\beta$ is a factor for normalization.

**Supplement C3**

Seven potential Bayesian methods have great potential to improve RL, but it doesn't mean other Bayesian methods cannot contribute to RL. Bayes' rule, Bayesian linear regression, Bayesian estimator, and probably approximately correct (PAC) Bayesian bound can be applied to RL. Bayes' rules can be used to fuses RL policy with a known handcrafted (risk-averse but suboptimal) control prior action distribution for better safety and convergence. Bayesian linear regression with neural networks is useful to derive better intrinsic reward. Iterative Bayesian estimators can assist the prediction of the next state for better policy iteration. PAC Bayesian bound can be used to upper bound the value approximation error (soft Bellman error) to correct a potential overestimation of current state value.

Hybrid RL policy via Bayes' rule. RL policy $\pi(\cdot\,|s_t)\sim\mathcal{N}(\mu_\pi,\sigma_\pi^2)$ can be fused with a known handcrafted (risk-averse but suboptimal) control prior action distribution $\psi(\cdot\,|s_t)\sim\mathcal{N}(\mu_\psi,\sigma_\psi^2)$ to form a *hybrid distribution* $\phi(\cdot\,|s_t)\sim\mathcal{N}(\mu_\phi,\sigma_\phi^2)$ to improve the convergence and safety. Control prior is a deterministic mapping function $a_t = \psi(s_t)$. The fusion of policies is achieved by factoring fused distribution via Bayes' rule

$$p\left(a\middle|\theta_\pi,\theta_\psi\right) = \frac{p(\theta_\pi,\theta_\psi|a)p(a)}{p(\theta_\pi,\theta_\psi)} = \eta p(a|\theta_\pi)p\left(a\middle|\theta_\psi\right)$$

where $\eta = \frac{p(\theta_\pi)p(\theta_\psi)}{p(\theta_\pi,\theta_\psi)p(a)}$. $p\left(a\middle|\theta_\pi,\theta_\psi\right)$ forms hybrid policy $\phi$ where its mean and variance are computed by $\mu_\phi = \frac{\mu_\pi\sigma_\psi^2+\mu_\psi\sigma_\pi^2}{\sigma_\pi^2+\sigma_\psi^2}, \sigma_\phi^2 = \frac{\sigma_\pi^2\sigma_\psi^2}{\sigma_\psi^2+\sigma_\pi^2}$.

Curiosity from Bayesian linear regression with neural networks to prepare intrinsic reward. Bayesian curiosity is taken to form combined reward $r_t$ for RL via

$$r_t = e_t + \eta\cdot c_t, c_t = \log(\sigma^2(o_t))$$

where $e_t$ is the immediate reward from the environment, and $c_t$ is the curiosity reward from Bayesian curiosity. $\eta$ is a hyperparameter that controls the weight of the curiosity reward. $\sigma^2(o_t)$ is the variance of *lower-bounded predictive posterior* (written as $\mu_N(\boldsymbol{o_i})$ and $\sigma_N^2(\boldsymbol{o_i})$) for intrinsic reward where predictive posterior is based on Bayesian linear regression (BLR) with neural network computed by

$$\mu_N(\boldsymbol{o_i}) = \boldsymbol{m_N^T}\phi_\psi(\boldsymbol{o_i}), \sigma_N^2(\boldsymbol{o_i}) = \beta^{-1} + \phi_\psi(\boldsymbol{o_i})^T\boldsymbol{S_N}\phi_\psi(\boldsymbol{o_i})$$

where network $\phi_\psi(\boldsymbol{o_i})$ with weight $\psi$ is used to capture nonlinearity and reduce dimensionality of the input, because BLR works poorly in high-dimensional dataset. $\beta$ is a hyperparameter representing noise precision in data. The mean $\boldsymbol{m_N}$ and variance $\boldsymbol{S_N}$ are the *initial posterior moments of BLR* after training with $N$ samples where the prior is captured by Gaussian. For any arbitrary demonstration $(x_i,a_i)\in\mathcal{D}$, network $\phi_\psi(x_i)$ is trained by minimizing the negative log-likelihood loss via SGD

$$\mathcal{L}(x_i,a_i) = \frac{\log 2\pi}{2} + \frac{\log\sigma_N^2(x_i)}{2} + \frac{(a_i-\mu_N(x_i))^2}{\sigma_N^2(x_i)}$$

Iterative Bayesian estimator to assist the prediction of future state. Future (next) state of agents can be predicted by combining designer's knowledge and explored data to improve the convergence of RL. This is achieved by a mixed dynamics

$$x' = f(x,a) + \xi^M$$

where $f$ is the deterministic part of dynamics. $\xi^M$ is the additive stochastic uncertainty captured by Gaussian with unknown mean and covariance. $\xi^M$ exists in collected data and it can be written as $\xi_j^\mathcal{D} = x_{j+1}^\mathcal{D} - f(x_j^\mathcal{D},u_j^\mathcal{D})$ where $\xi_j^\mathcal{D}$ is the uncertainty in dataset $\mathcal{D} = \{(x_j^\mathcal{D},u_j^\mathcal{D},x_{j+1}^\mathcal{D})\}$. It is possible to use $\xi_j^\mathcal{D}$ to approximate $\xi^M$ in an iterative Bayesian way via

$$\begin{bmatrix}\mu_k\\ \mathcal{K}_k\end{bmatrix} = IBE(\mu_{k-1},\mathcal{K}_{k-1},\xi_k^\mathcal{D})$$

where $\mu_k$ and $\mathcal{K}_k$ are the mean and covariance of dynamics. $IBE$ denotes the iterative Bayesian estimator. This results in mixed stochastic model $x' = f(x,u) + \hat{\xi}$ where $\hat{\xi}$ is the approximation of $\xi^M$. Finally, estimated next state $x'$ is used in policy iteration of RL to improve the convergence.

PAC Bayesian bound as the objective for RL. VI can be used to approximate the posterior of value function in RL, but the approximation error is unknown. The value approximation error (soft Bellman error) can be upper bounded by PAC via

$$\left\|Q_\theta^{\pi_\psi}(s,a) - Q^{\pi_\psi}(s,a)\right\|_{p_\infty^\psi,\pi_\psi}^2 \le \frac{1}{(1-\gamma)^2}\left\|\mathbb{E}_{a\sim\pi_\psi(\cdot|a)}\left[\mathcal{T}_\psi Q_\theta^{\pi_\psi}(s,a) - Q_\theta^{\pi_\psi}(s,a)\right]\right\|_{p_\infty^\psi,\pi_\psi}^2$$

where $Q_\theta^{\pi_\psi}(s,a)$ parameterized by $\theta$ is the approximation of true value $Q^{\pi_\psi}(s,a)$. $p_\infty^\psi$ is the stationary distribution of the state transition $p(\cdot\,|s,a)$ under policy $\pi_\psi$. $\gamma$ is a discounted factor. $\mathcal{T}_\psi$ is the Bellman backup operator. The upper bound of approximation error is used as a part of *PAC Bayesian bound* (loss of SAC) simplified as

$$L_{PAC}(\phi) \coloneqq R_N(q;\phi) + \sqrt{\frac{D_{KL}(q(\theta;\phi)||p_0(\theta))}{N}} - \mathbb{E}_{\theta,s,a}\left[r + \gamma\mathbb{V}_{s'}\left[\mathbb{E}_{a'}[Q_\theta^{\pi_\psi}(s',a')]\right]\right]$$

where the first part $R_N(q;\phi)$ is the empirical risk defined as the soft Bellman error. The second part $\sqrt{\frac{D_{KL}(q(\theta;\phi)||p_0(\theta))}{N}}$ pushes the predictor (variational posterior) $q(\theta;\phi)$ close to prior $p_0(\theta)$ where $N$ is the number of observations. The third part $\mathbb{E}_{\theta,s,a}\left[r + \gamma\mathbb{V}_{s'}\left[\mathbb{E}_{a'}[Q_\theta^{\pi_\psi}(s',a')]\right]\right]$ corrects a potential overestimation of current state value.

## Supplement D

### Supplement D1.

SafeOpt-MC computes the next sample via

$$a_n \leftarrow \arg\max_{a\in G_n\cup M_n}\max_{i\in\mathcal{I}} w_n(a,i)$$

A safe sets $S_n$ should be constructed to guarantee safety and convergence simultaneously. $G_n$ and $M_n$ are related to the construction of safe sets. $G_n$ is an optimistic set of parameters that could potentially enlarge the safe set $S_n$ which is defined using Lipschitz continuity property, while $M_n$ is the subset of $S_n$ that could either improve the estimate of the maximum or expand the safe set. $w$ is defined by $w_n(a,i) = u_n^i(a) - l_n^i(a)$. The upper bound of the contained set $C_n(a,i)$ at iteration $n$ is defined by $u_n^i(a) \coloneqq \max C_n(a,i)$. The lower bound of $C_n(a,i)$ is defined by $l_n^i(a) \coloneqq \min C_n(a,i)$ where the contained set is defined by $C_n(a,i) = C_{n-1}(a,i) \cap Q_n(a,i)$. The $Q_n(a,i)$ is the GP's confidence intervals for the surrogate function and $Q_n(a,i) \coloneqq [\mu_{n-1}(a,i) \pm \beta_n^{\frac{1}{2}}\sigma_{n-1}(a,i)]$ where $\beta$ is a scalar that determines desired confidence level. The convergence to global optimum can be further improved by exploring outside the initial safe area to create new safe area while still guaranteeing safety with high probability.

### Supplement D2

The constrained reward objective is written by

$$\max_{\pi\in\Pi}\max_{p_\theta\in\mathcal{P}} J(\pi,p_\theta)\,,\text{s.t.}\max_{p_{\theta^i}\in\mathcal{P}} J^i\left(\pi,p_{\theta^i}\right) \le d^i, \forall i \in \{1,\ldots,C\}$$

where reward objective $J(\pi,p_\theta) = \mathbb{E}[\sum_{t=0}^T r_t|s_0]$. Actions $a_t \sim \pi(\cdot\,|s_t)$. Next state $s_{t+1}$, reward $r_t$, and cost of unsafe behavior $c_t^i$ are inferred by unknown transition $p_\theta$ and $s_{t+1}, r_t, c_t^i \sim p_\theta(\cdot\,|s_t,a_t)$ where its posterior $p(\theta|\mathcal{D})$ is computed by GP. The *constraint of unsafe behaviors* $J^i\left(\pi,p_{\theta^i}\right) = \mathbb{E}\left[\sum_{t=0}^T c_t^i|s_0\right] \le d^i$. Cost $c_t^i \sim p(\cdot\,|s_t,a_t)$ and $c_t^i = 1$ if $s_t$ is harmful (e.g., the robot hits obstacles) where $i$ is a distinct unsafe behavior that should be avoided by agents. $d^i$ is a human-defined threshold. $\mathcal{P}$ is plausible transition distributions acquired via the methods *optimism in the face of uncertainty* and *upper confidence RL* (UCRL). Finally, CMDP is solved via the augmented *Lagrangian with proximal relaxation method*.

The objective of value function is written by

$$V_{\tilde\theta,t}^\theta \coloneqq \mathbb{E}_E\left[\sum_{j=t}^{T-1} r(j,x_j,u_j;\theta_R) - \left(I(\rho_j)\middle|\begin{array}{c} s_t = s \\ a_j = \bar a^*(j,s_j;\tilde\theta) \\ s_{j+1} = f(s_j,a_j;\theta_F) + \epsilon_{F,j} \\ \rho_j = \min_{\rho\ge 0}\rho\ \text{s.t.}\ s_j \in \mathcal{S}_\delta(\rho)\end{array}\right)\right]$$

where $\tilde\theta = (\tilde\theta_R,\tilde\theta_F)$ is the imperfect estimate of true parameter $\theta = (\theta_R,\theta_F)$. $\epsilon_{F,j}$ is the external perturbance. $I(\rho_j)$ is the penalty where $\rho \ge 0$ is the slack variable. $\mathcal{S}_\delta(\rho)$ is a cautious soft-constraint formulation with parameter $\delta$ which defines the degree of cautiousness. *Penalty* $I(\rho_j)$ is acquired by optimizing state trajectories subject to a *tightened state constraint set*. This bounds the number of unsafe episodes based on the Lipschitz continuity property, like the number of episodes in which the tight state constraints are violated, resulting in $\mathcal{S}_\delta(\rho)$. This method combines model-based RL using GP and posterior sampling with a safe policy approximated by $V_{\tilde\theta,t}^\theta$, and yields a new class of model-based RL policies with safety guarantees.

The conditional value at risk (CVaR) objective is written by

$$\max_{\pi\in\Pi^{M^+}} CVaR_\alpha\left(G_\pi^{M^+}\right) = \max_{\pi\in\Pi^{\mathcal{G}^+}}\min_{\sigma\in\Sigma^{\mathcal{G}^+}}\mathbb{E}[G_{(\pi,\sigma)}^{\mathcal{G}^+}]$$

where $\alpha$ is the confidence level. $G_\pi^{M^+}$ and $G_{(\pi,\sigma)}^{\mathcal{G}^+}$ are the distributions over total returns where $G_{(\pi,\sigma)}^{\mathcal{G}^+}$ is the perturbed distribution by the adversary, and they are defined by $G_\pi^M = \sum_{t=0}^H R(s_t,a_t)$. $M^+$ denotes a BAMDP. $\mathcal{G}^+$ is the Bayes-Adaptive CVaR stochastic game where "+" denotes the augmented parameter in BAMDP. $\Pi^{\mathcal{G}^+}$ denotes a set of Markovian agent policies mapping agent states to agent actions. $\Sigma^{\mathcal{G}^+}$ denotes a set of *Markovian adversary policies*. Maximizing CVaR objective leads to a perturbed agent-adversary policy $(\pi, \sigma)$ that is robust to epistemic uncertainty and aleatoric uncertainty. Computing of CVaR is intractable. Therefore, MCTS approximation is used to handle the large state space by focusing the search in promising areas, while the progressive widening paired with BO is used to handle the continuous action space for the adversary.

## Supplement E

### Supplement E1.

Model-based Bayesian RL with unknown reward is realized by extending the hyper-states $\phi$ to unknown parameters $\theta$ via $\theta = (\phi, \vartheta), S' \in \mathcal{S} \times \Theta$ where $\vartheta$ is the prior over unknown reward function. Hence, modeling the reward function by a suitable distribution is the key to moving forward. In simple cases, the reward function can be captured by Beta distribution or Dirichlet distribution. In complex reward case, it can be captured by Gaussian distribution $\vartheta = \{\mu, \sigma\}$. The prior can be defined on the variance by

$$f(\psi) \propto \psi \exp(-\psi^2 \frac{\sigma_0^2}{2})$$

where $f(\psi)$ is the prior density, and the precision $\psi = \frac{1}{\sigma}$. The prior can be also defined on the mean by

$$f(\mu) \propto \mathcal{N}(\mu_0, \sigma^2)$$

where $f(\mu)$ is the prior density. With $n$ sampled rewards (observations), $f(\psi)$ and $f(\mu)$ switch into $f(\psi) \propto \psi^{n-1} \exp(-\psi^2 \frac{(n\hat{\sigma}+\sigma_0^2)}{2})$ and $f(\mu) \propto \mathcal{N}(\hat{\mu}, \sigma^2/n)$ where $\hat{\mu}$ and $\hat{\sigma}$ are the mean and variance of $n$ samples. Hence, when sampling the posterior of unknown parameters to compute the policy like the value function of model-based Bayesian RL, the posterior of the reward function should be sampled simultaneously. This process includes two-stage posterior samplings, resulting in a big challenge in the computation.

### Supplement E2.

The difference of BA-POMDP and POMDP is that BA-POMDP's hyper-states $S' \in \mathcal{S} \times \Phi \times \Psi$ where $\Phi$ and $\Psi$ capture the spaces of Dirichlet distribution for the conjugate priors over *unknown transition distribution* $P(s'|s,a)$ and *observation distribution* $P(o|s,a)$, respectively. In BA-POMDP, the hyper-state transition is factorized by

$$P\left(s', \phi', \psi', o \middle| s, \phi, \psi, a\right) = P\left(s' \middle| s, \phi, a\right) P(o|s', \psi, a) P\left(\phi' \middle| \phi, s, a, s'\right) P\left(\psi' \middle| \psi, o, a, s'\right)$$

where

$$\begin{cases} P\left(\phi' \middle| s, a, s', \phi\right) = \begin{cases} 1 \text{ if } \phi'_{s,a,s'} = \phi_{s,a,s'} + 1 \\ 0, \text{otherwise} \end{cases} \\ P\left(\psi' \middle| \psi, o, a, s'\right) = \begin{cases} 1 \text{ if } \phi'_{s,a,s'} = \phi_{s,a,s'} + 1 \\ 0, \text{otherwise} \end{cases} \\ P\left(s' \middle| s, \phi, a\right) = \dfrac{\phi'_{s,a,s'}}{\sum_{s'' \in S} \phi'_{s,a,s''}} \\ P(o|s', \psi, a) = \dfrac{\phi'_{o,a,s'}}{\sum_{o \in O} \phi'_{o,a,s'}} \end{cases}$$

The optimal value function of BA-POMDP is defined by

$$V^*(b_t(s, \phi, \psi)) = \max_{a \in A} \left[ R'(s, \phi, \psi, a) b_t(s, \phi, \psi) + \gamma \sum_{o \in O} P(o|b_{t-1}, a) V^*(\tau(b_t, a, o)) \right]$$

which includes many beliefs because the hyper-state count is $\Phi \times \Psi$ in BA-POMDP, instead of $\Phi$ in POMDP. The value function estimation requires estimating the Bellman equation over all possible hyper states for every belief. This means all models should be sampled from Dirichlet posterior. Then, models are solved and action is sampled from solved models. However, this estimation method is intractable. One possible way to solve this problem is selecting actions via Bayes risk which is the smallest expected loss defined by

$$BR(a) = \sum_{s,\phi,\psi} Q(b_t(s, \phi, \psi, a)) b_t(s, \phi, \psi) - \sum_{s,\phi,\psi} Q(b_t(s, \phi, \psi, a^*)) b_t(s, \phi, \psi)$$

This method generates a bound of the sample number, but provides a myopic view of uncertainty.

### Supplement E3.

The multi-agent RL (MARL) focuses on finding the *global* maximum reward. In this process, each agent tries to maximize their own *interest* for a higher reward. However, the interest of one agent may contradict the interests of other agents. MARL faces the following challenges: 1) *Partial observability.* Each agent just knowns its own state and the states of its neighbors; 2) *Unknown system dynamics.* It's hard to search and find the best prior distribution for the system model, and the observations always mix with noise, bias, and error; 3) *non-stationarity.* The environment becomes non-stationary when the agents interact with each other. The optimal behavior of each agent depends not only on the environmental information received but also on the behavior of other agents. Reward functions and agent's behavior policies change overtime. This makes the overall dynamics or decoupled dynamics hard to converge; 4) *Scalability with exponential joint policy space.* The joint policy space expands exponentially with the increase of the agent number. This means exponential computations; 5) *Reward assignment.* The agents can share the same (global) reward where their joint behaviors can be measured. However, the behavior of individual agents cannot be measured in the case with shared reward, but this problem may be solvable by assigning different (local) rewards to the agents with a proper reward structure. How to design the reward structure/hierarchy still remains to be a challenging job; 6) *The coordination of the agent's policies.* It matters to coordinate the interest of each agent to find better balanced policies of agents for maximum global reward.

MARL (Model-based) can be solved by formulating it to *centralized MARL* and *decentralized MARL*, given its training scheme.

Centralized MARL. Global environment dynamics is maintained in the centralized MARL where the current state and joint action are used to predict the next joint observation, next state, and current reward via $P\left(s', o', r \middle| s, a\right)$. Theoretically, the obvious *advantage* of centralized MARL is in addressing the problems of partial observability, non-stationarity, and coordination of agents' policies. Centralized MARL assumes the full observability where partial observability, non-stationarity, and coordination of agents will not be considered or can be ignored, because of just maintaining a global environment dynamics as the behavior polies of all agents. The obvious *disadvantage* of centralized MARL is poor scalability that results from exponential joint observation and action spaces, with the increase of agent number.

In practice, full observability is impossible, and the *centralized training and decentralized execution* (CTDE) paradigm improves partial observability and non-stationarity to alleviate this problem. Exponential observation and action spaces that results in poor scalability can be alleviated by d*isentangling complex global (joint) dynamics* into dynamics of individual agents where each agent has to consider the interactions with other agents and infer the future state and action of other agents. Problems of exponential observation and action spaces can be also alleviated by directly applying *dimensionality reduction*. The interactions among agents can be simplified as the *pairwise interactions* via model-based multi-agent mean-field RL.

Decentralized MARL. The dynamics of all agents are maintained in the decentralized MARL, instead of one global environment dynamics in centralized MARL. Decentralized MARL emphasizes the observation space, action space, and dynamics of each agent, as well as the interactions of agents. Therefore, obvious *disadvantages* of decentralized MARL are the partial observability, non-stationarity, and coordination of agents. However, the scalability problem (obvious *advantage*) in centralized MARL is not so important in decentralized MARL, because exponential joint dynamics may not be maintained.

The problem of partial observability is expected to be solved via communication to *share observations*. The problem of non-stationary can be alleviated via *learning a joint value function* or global environment dynamics, under the sacrifice of scalability. The problems of non-stationary and agent coordination can be alleviated via *learning the dynamics or behavior models of other agents*, or via communication to *share the behavior policies or intentions of agents*.

To further reduce the computation for better scalability, in the implementation, planning methods like MCTS are popular to be applied to the POMDP for online planning, resulting in *partially observable MC planning* (POMCP). The joint policy space can be reduced by the *factored-value POMCP* (FV-POMCP) which decomposes the value function and global look-ahead tree into overlapping factors and multiple local look-ahead trees respectively. Moreover, the above tricks can be combined together, resulting in the *transition-decoupled factored value based MC online planning* (TD-FV-MCP) to further reduce the computation of multi-agent motion planning. However, the searching and expanding the local look-ahead tree for each agent lacks efficiency, resulting in suboptimal convergence in training.

Overall, the research of MARL is still in the early stage and few practical methods are found to better solve the challenges of MARL. For real-world problems, MARL still has a long way to go to address more challenges like high dimensions, unexpected noise and disturbance, and OOD data. Bayesian inference gives a good direction for MARL to alleviate these challenges. There are some early-stage attempts to combine the BAMDP and MARL by formulating MARL problems as the BA-TD-POMDP to address the partial observability and policy uncertainty, but more Bayesian methods on MARL are still under-researched.

**Supplement E4.**

Complex problems in the real world can be divided into multiple *interdependent* tasks. If the knowledge in these tasks can be shared in the learning process, the learning speed of each task is expected to be improved. Here, knowledge refers to the shared instances, representations, and parameters that are robust and transferable abstractions of the environment. Generally, multi-tasking can be applied from two levels: *Single-agent multiple tasks* and *multi-agent multiple tasks*. Single-agent multiple tasks refers to learning sequential/continuous tasks by one agent, therefore knowledge learned from former tasks can be used to learn the later tasks (e.g., lifelong learning). Multi-agent multiple tasks refers to learning many tasks by multiple agents *in parallel* (parallel multi-task RL), *in a sequential manner*, or *in a mixed manner*. Parallel multi-task RL (MTRL) is a widely accepted multi-task learning methodology where *one* learner (critic) combines with *many* actors or agents to learn different tasks in the same environment. Asynchronous advantage actor-critic (A3C) or advantage actor-critic (A2C) can be seen as a kind of parallel MTRL where multiple agents work in parallel in different tasks of the same environment to update the global shared critic. MTRL faces many challenges: 1) *Scalability*. With the increase of task number, collected data and related processing time increase exponentially. MTRL should make the best of knowledge from each task or agent to accelerate the learning of other tasks for better data and computation efficiencies; 2) *Distraction dilemma*. There is a conflict of interest between the ultimate goal for generalization and the goals of individual tasks (specialization), due to the rewards associated with tasks (in-task rewards). This makes MTRL focus on wrong or salient tasks and ignore less important tasks especially in hierarchical RL (HRL), resulting in poor generalization. Hence, a balanced reward design in MTRL matters; 3) *Partial observability.* The same as MARL, agents in MTRL don't know the full observation of the environment in real-world problems; 4) *Catastrophic forgetting.* Neural networks in MTRL may lose the knowledge learned from previous tasks when learning a new task, due to the conflict objectives/interests of tasks; 5) *Negative knowledge transfer.* If big differences exist among tasks, knowledge learned from one task may negatively impact the learning of other tasks.

To address these challenges, a *centroid policy* that can transfer common behaviors to agents is distilled from policies of agents. This results in robustness against the distraction dilemma, catastrophic forgetting and negative knowledge transfer via regularizing agent policies using centroid policy. A3C with one learner and multiple actors is extended to distributed case with *multiple learners and multiple actors*, resulting in the improvement in scalability. Distraction dilemma is alleviated by *constraining the proportional impact of each agent to overall dynamics objective*.

## Supplement F.

### Supplement F1.

Comparisons of Bayesian surrogate models

| Models | Advantages | Limitations |
|---|---|---|
| Gaussian | Suits time continuous stochastic settings | Hard to cope with skewed or multi-modal distributions |
| Dirichlet | Applicable for categorical/discrete settings | --- |
| Finite mixture and hierarchical | Computationally convenient estimation via hierarchical clustering | Hard to select prior/size of prior |
| Infinite mixture and hierarchical | Applicable for data consisting of infinite clusters | Hard to select prior/size of prior;<br>Trade-off of temperature (concentration) |

### Supplement F2.

Comparisons of Bayesian methods.

| Methods | Utility | Advantage | Limitation |
|---|---|---|---|
| GPR-based Bayesian learning | Computing posterior (transition model) | --- | The difficulty in selection of kernel/weight size<br>Computational inefficiency with large dataset<br>Errors from linear Gaussian assumptions<br>Difficulty in handling nonlinear transformation model<br>Sample inefficiency in high-dimensional cases |
| Variational inference | Posterior approximation | CAVI (---)<br>MC-CAVI (low computation)<br>SVI (fast convergence, and high accuracy) | CAVI (high computation)<br>MC-CAVI (trade-off of computation and accuracy)<br>SVI (poor interpretability) |
| Bayesian optimization | Design of data-efficient acquisition function;<br>Computing posterior | *Improvement-based:* PI & EI (---)<br>*Optimistic*: GP-UCB (---)<br>*Information-based:*<br>ES (avoid aggressive exploration)<br>PES (reduce the computation)<br>MES (reduce computation and higher sample efficiency)<br>*Portfolios:*<br>GP-Hedge (---)<br>ESP (consider the exploration information) | *Improvement-based*: PI & EI (aggressive exploitation)<br>*Optimistic*: GP-UCB (trade-off of uncertainty weight)<br>*Information-based:*<br>ES (expensive computation in posterior approximation)<br>PES (---)<br>MES (---)<br>*Portfolios:*<br>GP-Hedge (not consider the exploration information)<br>ESP (---) |
| Bayesian neural networks | Computing posterior in high-dimensional case | BNN with Laplace approximation (---)<br>BNN with MCMC (reliable and explainable approximations)<br>BNN with CAVI (---)<br>BBB (suits smooth data, sufficient data quantity, and sufficient time for inference)<br>BNN with MC dropout (suits large networks, sparse data, and limited time requirement for inference; less sensitive to prior choice; sampling-efficient) | BNN with Laplace approximation (overfits with the increase of model complexity<br>BNN with MCMC (hard to balance accuracy and sample efficiency)<br>BNN with CAVI (unreliable and inaccurate approximations)<br>BBB (under-estimated variance and unpredictable convergence)<br>BNN with MC dropout (under-estimated variance and unpredictable convergence) |
| Bayesian active learning | Selecting informative samples with acquisition functions | BMDAL (considers sample uncertainty and correlations)<br>DBAL (better sample uncertainty)<br>DBAL with RL (tunable acquisition function)<br>DBAL with meta learning (balance of sample importance between historical and new samples) | BMDAL (---)<br>DBAL (pattern collapse phenomenon)<br>DBAL with RL (---)<br>DBAL with meta learning (---) |
| Bayesian generative models | VAE (encoder and decoder)<br>Diffusion model (denoising network) | VAE (high-quality representations and better interpretability)<br>Diffusion model (flexibility to incorporate guidance for high-quality prediction for unseen scenarios) | VAE (unexplainable optimization process)<br>Diffusion model (challenges in data representation, noise schedule, selection of terminal distribution, parameterization of denoising network, and guidance of denoising/sampling process.) |
| Bayesian meta learning | Deriving local models and meta model | Black-box approach (good specialization of meta model for high-dimensional problems)<br>Optimization-based approach (good generalization among low-dimensional tasks with different distributions) | Black-box approach (data-inefficient due to tight relationship of local and meta models)<br>Optimization-based approach (convergence is sensitive to meta model initialization) |
| Lifelong Bayesian learning | Task/data clustering via DPMM | DPMM with EM (---)<br>DPMM with MCMC (---)<br>DPMM with VI (fast approximation) | DPMM with EM (slow approximation)<br>DPMM with MCMC (suffer from the trade-off of computation and accuracy)<br>DPMM with VI (unreliable optimization process) |

## Supplement F3.

Comparisons of Bayesian methods on classical RL

| **Algorithms** | **Bayesian methods & RL (if exist)** | **Utility** | **Advantage** | **limitation** |
|---|---|---|---|---|
| Model-based RL | GPR;<br>Bayesian filter with appro;<br>Optimal value RL; | GPR (posterior computing in linear-Gaussian case)<br>Bayesian filter with appro.( posterior computing in nonlinear case)<br>Dynamics with function approximation | Suit high-dimensional case due to function approximations with (Bayesian) neural networks | GPR (---)<br>Bayesian filter with appro.(trade-off of accuracy, computation, and interpretability due to different approximations)<br>Two-layer optimization problem of local dynamics and RL policy |
| Model-baesd Bayesian RL | BAMDP | Learning of large transition and RL policy | Large transition suits complex small-scale problems<br>Posterior is assumed to be Dirichlet to simplify expensive computation of transition | Expensive computation and slow convergence of value function that require value approximation and exploration bonus<br>Two-layer optimization problem of local dynamics and RL policy |
| Model-free Bayesian RL | GPTD;<br>BPG;<br>BAC | GPTD (value function is modeled by GP)<br>BPG (gradient is rewritten to BQ and solved by GP)<br>BAC (Q value and gradient rely on GP and BQ) | GPTD (---)<br>BPG (lesser computation due to BQ)<br>BAC (faster convergence and lesser computation due to actor-critic architecture and BQ) | GPTD (posterior relies on approximations)<br>BPG (large variance due to trajectory-based policy gradient)<br>BAC (large bias due to step-based return) |
| Bayesian inverse RL | GPR | Computing reward function modeled by GP | --- | Unstable training that requires regularization<br>Posterior relies on approximations |

## Supplement F4.

Applications of potential Bayesian methods on RL

| **Bayesian methods** | **Utility** | **Applications when combining with RL** |
|---|---|---|
| Variational inference | Provide ELBO for optimization | Assist deriving RL policy<br>Assist approximating agent's reward function (extrinsic and intrinsic) |
| Bayesian optimization | Provide acquisition function to reduce samples | Hyper-parameter tunning<br>Parameter space reduction for policy search |
| Bayesian neural network | Provide uncertainty quantification of network output | Uncertainty for exploration policy<br>Uncertainty for intrinsic reward<br>Uncertainty for loss/objective function<br>BNN as environment dynamics/transition (model-based) |
| Bayesian active learning | Provide acquisition function to select informative samples | Aid BO, inverse RL, imitation learning/learning from demonstrations, and preference-based RL for better sample/time efficiency |
| Bayesian generative models | Provide encoder and decoder (VAE)<br>Provide denoising network (Diffusion model) | *VAE:*<br>Provide encoder as high-level generalizable hierarchical policy/prior and decoder for interpretability<br>Enable independent training of encoder, decoder, and RL policy<br>*Diffusion models:*<br>Provide planner with the help of an independent reward model<br>Provide high-level planner/controller for multi-agent RL problems |
| Bayesian meta-learning | Provide optimization-based and black-box approaches | Few-shot meta-RL for simple tasks with narrow distribution<br>Many-shot meta-RL for complex hierarchical tasks with diverse distributions |
| Lifelong Bayesian learning | Provide DPMM like CRP and SBP for infinite clustering | Enable infinite number of model-free policies<br>Enable infinite number of model-based policies |

## Supplement F5.

The proposed questions after analysis.

Q1. In unknown reward cases, expensive *two-stage posterior computing (local and global models)* in model-based (Bayesian) RL.

Q2. In POMDP, high-dimensional hyper-state space and corresponding *multi-stage value function computing (local and global models)*.

Q3. In MARL, how to *narrow the choice of prior distribution* automatically to get a sub-optimal assumption of prior distribution (local model), for facilitating the posterior computing (global model).

Q4. In MARL, expensive *multi-stage/dimensional posterior computing (local and global models)* for better scalability in centralized MARL.

Q5. In MARL, when the *dynamics of agents should be coupled (local model)* in decentralized MARL for better scalability and reward (global model).

Q6. In MARL, how to add more deterministic factors to facilitate the *multi-stage/dimension posterior computing (local and global models)* by better reward functions and agent coordination.

| |
|---|
| Q7. In multi-task problem, how to better scale the *contribution of each subtask (local model)* for the learning overall task (global model), therefore facilitating the reward design. |
| Q8. In nonlinear non-Gaussian case, how to derive *task-dependent propagation models to approximate nonlinear transformation models (local model)*, therefore nonlinear stochastic process becomes deterministic process to facilitate posterior computing (global model). |
| Q9. In nonlinear non-Gaussian case, how to *predict/estimate a suboptimal expected posterior distribution* (local model) to facilitate SDC (global model). |
| Q10. In HRL. how to *enable manager policy (global model) to better initiate/sample, keep, and drop option candidates (local model)* for faster convergence of option policies. |